\ifx\pdfoutput\undefined\else\pdfoutput=1\fi
\documentclass[10pt,a4paper]{article}
\usepackage[margin=2.0cm]{geometry}
\usepackage{graphicx}
\usepackage{booktabs}
\usepackage{xcolor}
\usepackage{hyperref}
\usepackage{amsmath,amssymb}
\usepackage{enumitem}
\usepackage{float}
\usepackage{multicol}
\usepackage[breakable]{tcolorbox}
\usepackage[framemethod=TikZ]{mdframed}
\usepackage{caption}
\usepackage{subcaption}
\usepackage{listings}
\usepackage{pdfpages}

\definecolor{cblue}{HTML}{4C72B0}
\definecolor{cgold}{HTML}{DD8452}
\definecolor{cred}{HTML}{C44E52}
\definecolor{cgreen}{HTML}{55A868}
\definecolor{codebg}{HTML}{F7F7F7}
\definecolor{bgskill}{HTML}{F0F7FF}
\definecolor{abstractbg}{RGB}{232, 240, 253}
\definecolor{abstractlink}{RGB}{20, 90, 170}  
\definecolor{msblue}{HTML}{0078D4}
\definecolor{msred}{HTML}{F25022}
\definecolor{msgreen}{HTML}{7FBA00}
\definecolor{msyellow}{HTML}{FFB900}

\hypersetup{colorlinks=true, linkcolor=cblue, citecolor=cblue, urlcolor=cblue, breaklinks=true}

\newcommand{\ideaword}{\textcolor{msred}{I}\textcolor{msgreen}{d}\textcolor{msblue}{e}\textcolor{msyellow}{a}}
\newcommand{\corrauthmark}{\textsuperscript{\ddag}}
\newcommand{\authorunit}[3]{\mbox{#1\textsuperscript{#2}#3}}
\newcommand{\reporttitle}{ResearchStudio-\ideaword: An Evidence-Grounded Research-Ideation Skill Suite\\from ML Conference Outcomes}
\newcommand{\reportauthorslineone}{\authorunit{Qihao~Zhao}{1}{}, \authorunit{Yangyu~Huang}{2}{\corrauthmark}, \authorunit{Yalun~Dai}{1,2}{}, \authorunit{Lingao~Xiao}{2,3}{}, \authorunit{Jianjun~Gao}{1}{}, \authorunit{Xin~Zhang}{2}{}, \authorunit{Wenshan~Wu}{2}{}}
\newcommand{\reportauthorslinetwo}{\authorunit{Scarlett~Li}{2}{}, \authorunit{Yang~He}{3,4}{}, \authorunit{Yan~Lu}{2}{}, \authorunit{Yap~Kim~Hui}{1}{\corrauthmark}}
\newcommand{\reportauthors}{\reportauthorslineone\\[-1pt]\reportauthorslinetwo}
\newcommand{\reportaffiliations}{\color{black!55}\textsuperscript{1}Nanyang Technological University \quad \color{black!55}\textsuperscript{2}Microsoft Research \quad \color{black!55}\textsuperscript{3}National University of Singapore\\[-1pt]\color{black!55}\textsuperscript{4}CFAR, A*STAR}

\title{\textbf{\reporttitle}}
\author{\reportauthors}
\date{\reportaffiliations}

\begin{document}
\begin{center}
{\large\bfseries \reporttitle\par}
\vspace{8pt}
{\normalsize \reportauthorslineone\par}
{\normalsize \reportauthorslinetwo\par}
\vspace{4pt}
{\small \reportaffiliations\par}
\end{center}

\begin{mdframed}[backgroundcolor=abstractbg,
                 linewidth=0pt,
                 roundcorner=12pt,
                 leftmargin=12pt,
                 rightmargin=12pt,
                 innerleftmargin=16pt,
                 innerrightmargin=16pt,
                 innertopmargin=12pt,
                 innerbottommargin=12pt,
                 skipabove=6pt,
                 skipbelow=6pt]
\hypersetup{urlcolor=abstractlink}
Large language models have made research ideation increasingly accessible, yet effective idea development requires more than generating candidate directions. Researchers must ground a problem in current literature, identify meaningful bottlenecks, differentiate from existing solutions, and evaluate risks before committing to implementation. We present ResearchStudio-Idea as a reusable skill suite for this first mile of research ideation. The suite includes Paper-Search, a standalone multi-source literature search skill; Scoop-Check, a standalone prior-art collision checker for novelty claims; and \emph{IdeaSpark}, the end-to-end skill that composes evidence grounding, pattern-guided generation, collision retrieval, audit, and idea-card rendering into one workflow. IdeaSpark is constructed from a corpus of 1{,}947 machine learning conference papers collected from ICLR, ICML, and NeurIPS between 2021 and 2025, including Oral papers, a separately tracked high-citation subset, and rejected submissions. Analysis of these outcomes reveals 31 recurring ideation sub-patterns, consolidated into 15 reusable ideation patterns. Each pattern is operationalized as a structured card containing research contexts, bottleneck types, differentiation strategies, supporting precedents, and common failure modes. Given a research problem and an evidence bundle, IdeaSpark evaluates evidence readiness, reconstructs the surrounding research context, identifies unresolved bottlenecks, selects relevant patterns, instantiates one candidate direction, retrieves potentially conflicting prior work, and performs outcome-informed auditing. This workflow transforms reusable ideation patterns into traceable research proposals. Blind automated-judge evaluations show that IdeaSpark consistently produces stronger research proposals than no-skill and generic-skill baselines while maintaining competitive novelty. These results suggest that large-scale conference outcomes contain reusable signals about how impactful research directions are formulated, differentiated, and evaluated, and that such signals can be operationalized as practical skills for evidence-grounded research ideation.

\vspace{6pt}
\noindent
\begin{minipage}[c]{0.50\linewidth}
{\small\textbf{Project:}~\url{https://aka.ms/ResearchStudio}}
\end{minipage}\hfill
\begin{minipage}[c]{0.44\linewidth}
\raggedleft
\raisebox{-0.48\height}{\includegraphics[height=1.10cm]{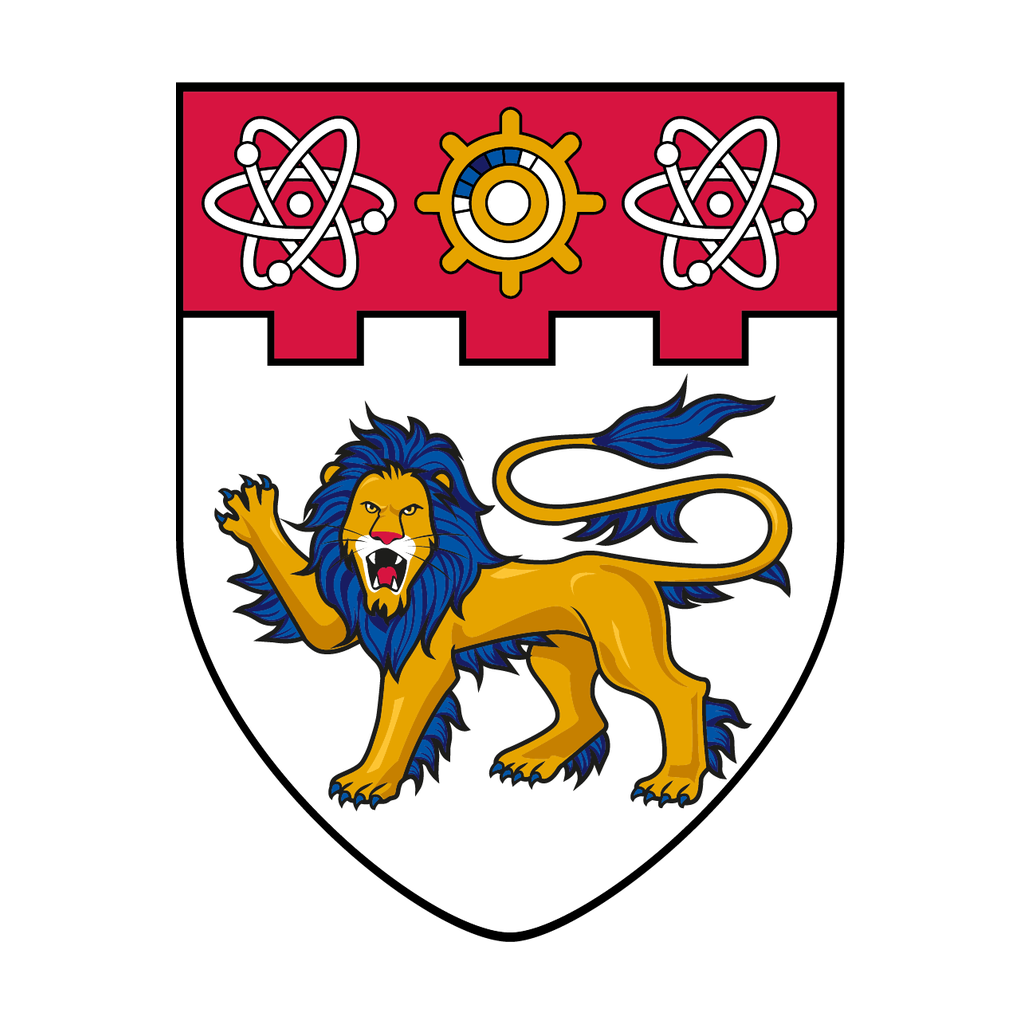}}\hspace{5pt}\raisebox{-0.48\height}{\includegraphics[height=1.10cm]{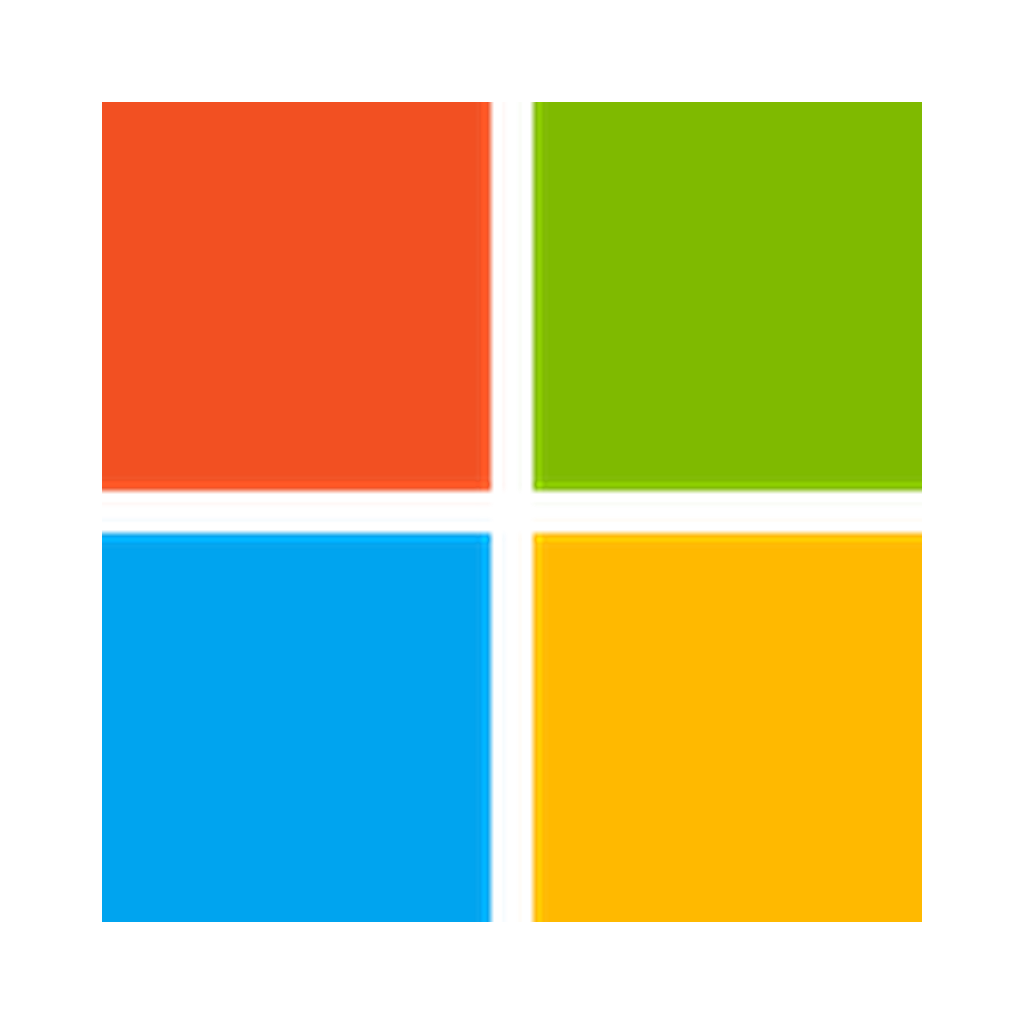}}\hspace{5pt}\raisebox{-0.48\height}{\includegraphics[height=1.10cm]{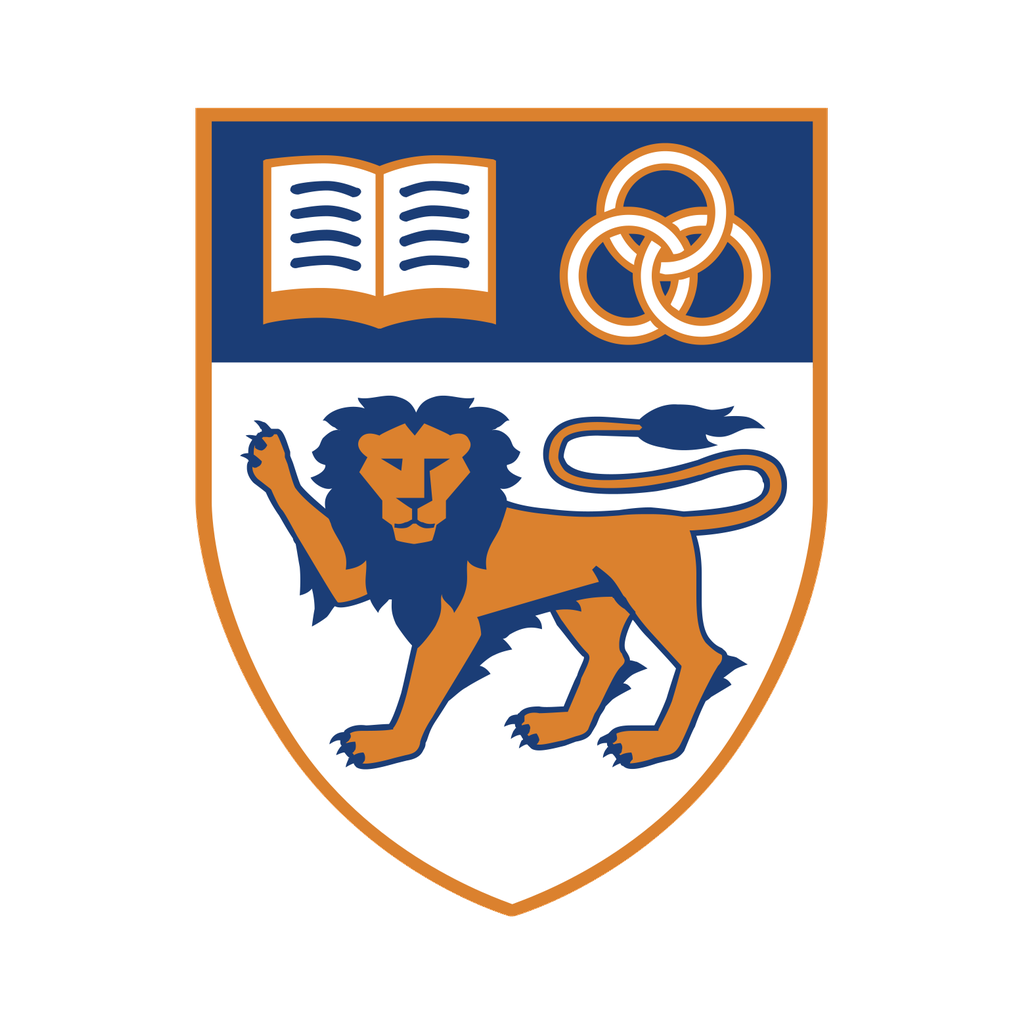}}\hspace{5pt}\raisebox{-0.48\height}{\includegraphics[height=1.10cm]{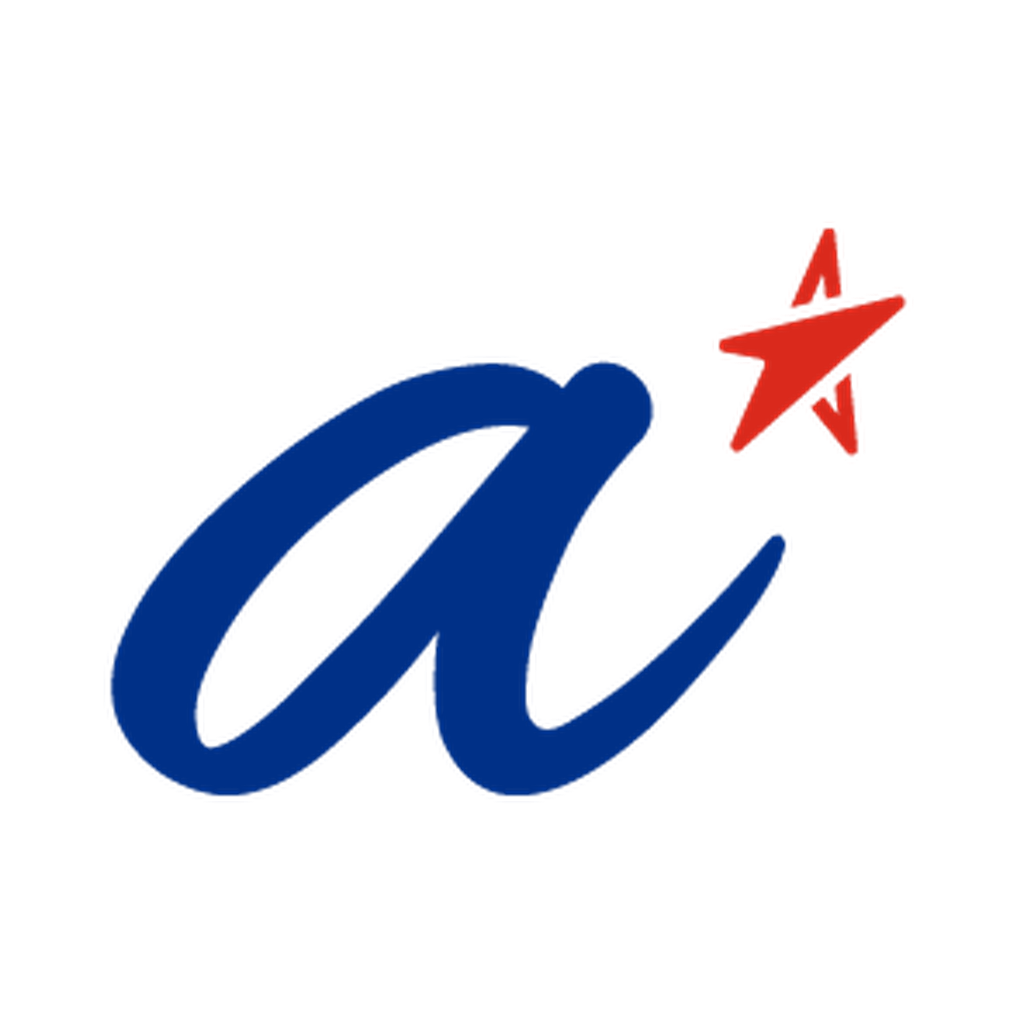}}
\end{minipage}
\end{mdframed}

\begin{figure}[H]
\centering
\captionsetup{font=small}
\includegraphics[height=5.40cm]{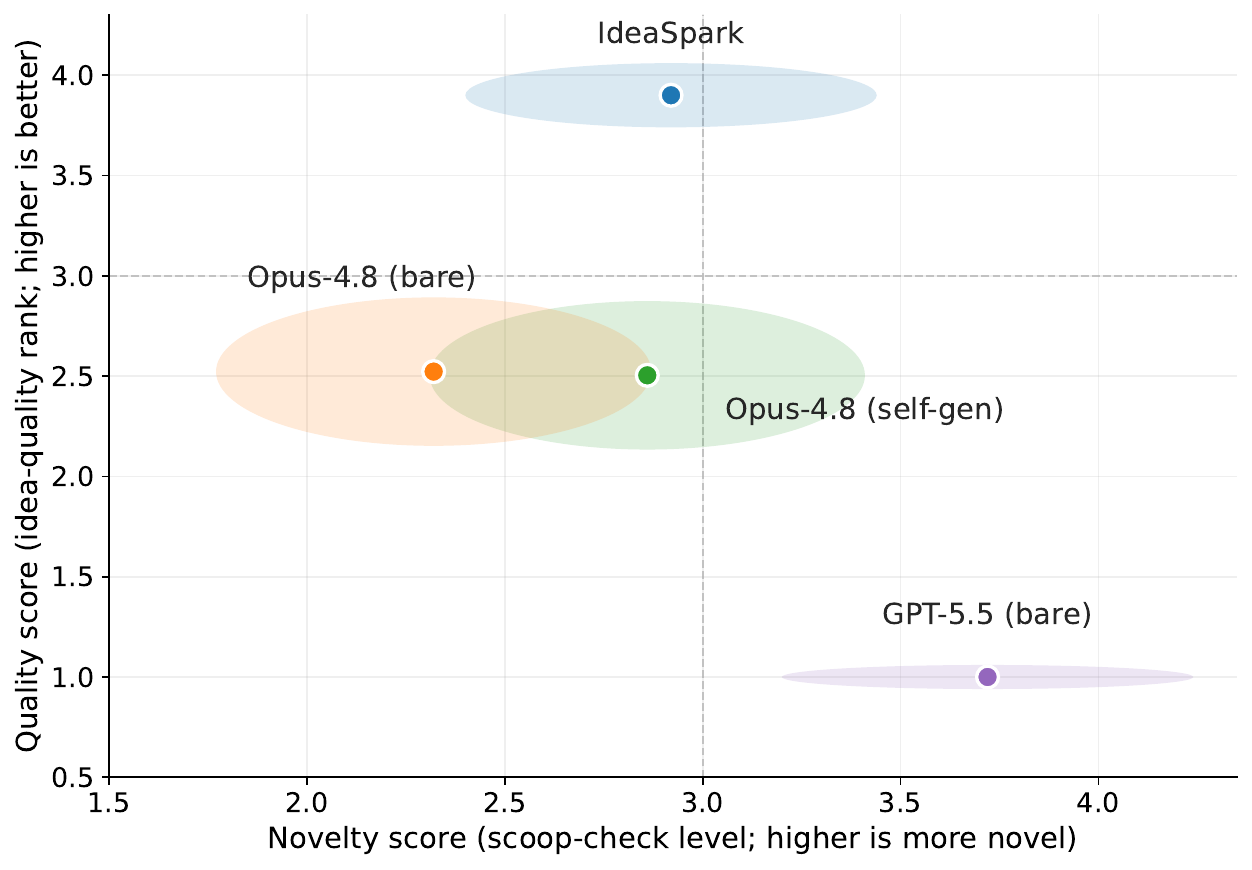}\hspace{10pt}\includegraphics[height=5.40cm]{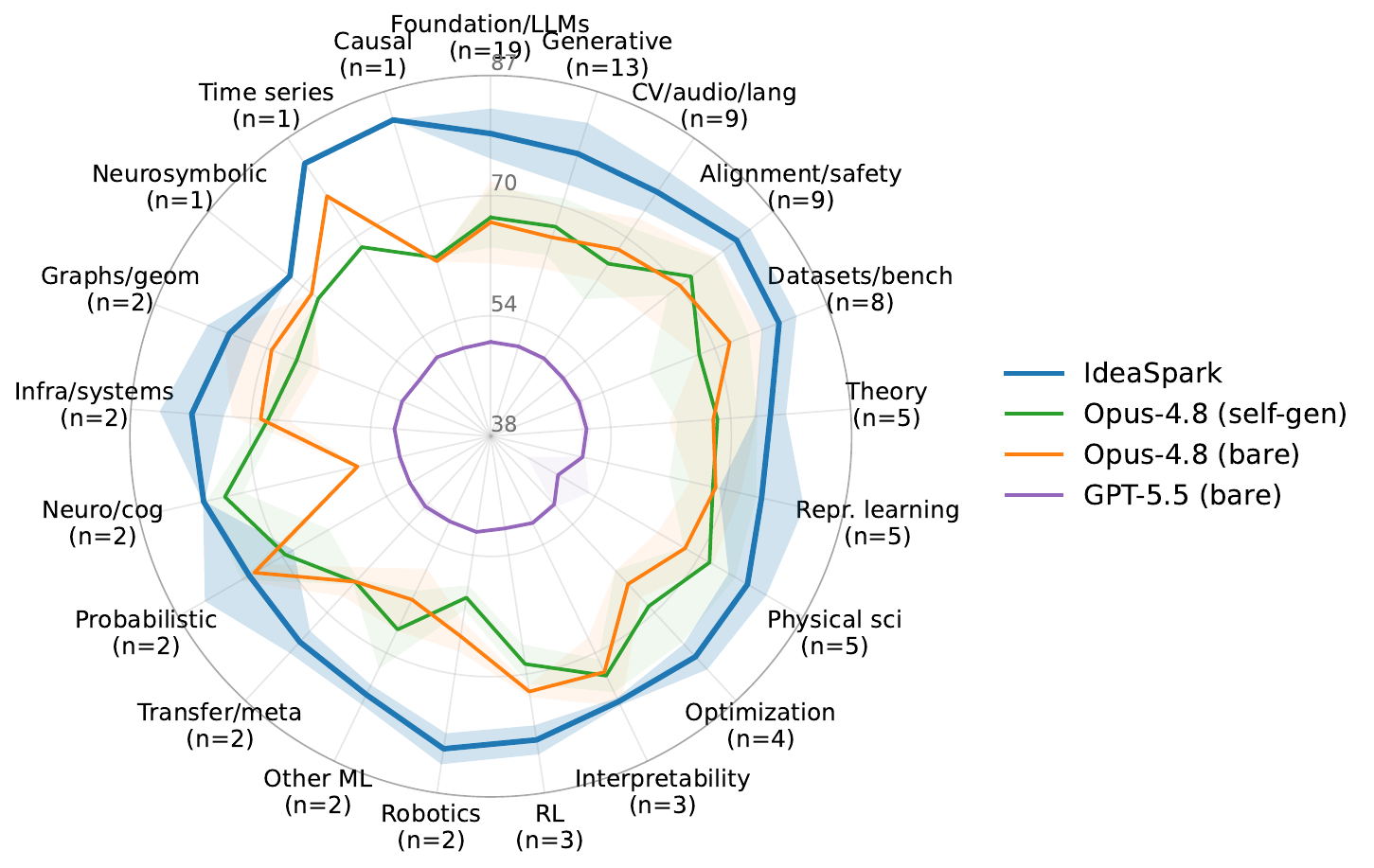}
\caption{IdeaSpark improves idea quality while maintaining competitive novelty in blind automated-judge evaluations. Left: quality--novelty trade-off over 100 ICLR-2026-Oral seeds, each judged in 3 blind rounds against three baselines: Opus-4.8 bare, Opus-4.8 self-generated, and GPT-5.5 bare. IdeaSpark occupies the high-quality, competitively novel region, whereas GPT-5.5 illustrates a novel-but-empty failure mode: high apparent novelty but substantially lower quality. Right: mean idea-quality across 21 ICLR primary-area domains; IdeaSpark is highest in every domain, suggesting the gain is broad rather than domain-specific.}
\label{fig:quality_novelty}
\end{figure}

\renewcommand{\thefootnote}{}
\makeatletter
\long\def\@makefntext#1{\noindent #1}
\makeatother
\hypersetup{hidelinks}
\footnotetext{ \textsuperscript{\ddag}Corresponding author: \href{mailto:yanghuan@microsoft.com}{yanghuan@microsoft.com}, \href{mailto:EKHYap@ntu.edu.sg}{EKHYap@ntu.edu.sg}}
\addtocounter{footnote}{-1}

\newpage
{\small
\tableofcontents
}
\newpage

\section{Introduction}
\label{sec:intro}

\subsection{Problem and scope}

LLM-based research agents have made scientific ideation easier to scale. Recent systems can retrieve papers, propose hypotheses, coordinate specialist agents, plan experiments, write code, and draft reports \cite{lu2024aiscientist,agentlab2025,airesearcher2025,idea2plan2025,ai_coscientist2025,autoresearch2025}. Search-based methods expand the candidate space. Novelty tools make prior-art checking more explicit. This progress shifts the bottleneck. The hard question is no longer whether an LLM can produce a plausible proposal. It is whether early-stage research work can be organized into reusable skills that ground evidence, generate one defensible direction, and audit its distance from prior art before experiments begin.

ResearchStudio-Idea addresses this first-mile problem as a suite of three open skills. \emph{Paper-Search} provides reusable literature grounding across arXiv, DBLP, OpenAlex, OpenReview, Semantic Scholar, and Crossref. \emph{Scoop-Check} performs claim-level prior-art collision checking by decomposing a proposed novelty into problem framing, core mechanism, key insight, and application domain, then comparing those axes against retrieved prior work. \emph{IdeaSpark} is the end-to-end skill: it composes literature grounding, bottleneck diagnosis, pattern-guided candidate construction, collision retrieval, audit, and idea-card packaging into one research-problem-to-idea-card workflow. Paper-Search and Scoop-Check are therefore not hidden implementation details; they are standalone skills that also supply reusable search and review functions inside the larger ideation loop.

The empirical center of this report is IdeaSpark because it is the point where ResearchStudio-Idea's search, generation, and review functions must work together. We study it as \emph{outcome-grounded skill induction} for LLM research ideation. The object we induce is an inference-time skill, not an acceptance predictor, novelty score, or paper-writing agent. The skill consists of ideation pattern cards, workflow prompts, schemas, retrieval hooks, and validators that an LLM loads while generating a structured idea. We ground it in public ML conference outcomes because those outcomes contain traces of research practice. Oral papers show cases that program committees elevated. High-citation papers show what the community reused. Rejected submissions show how similar attempts fail. The goal is to turn those traces into operational guidance, while keeping acceptance and execution outside the claim.

This report asks two questions. First, can conference outcomes be mined into a compact but usable map of ideation patterns? Second, can that map be packaged as a model-agnostic skill suite whose workflow generates and audits one research idea? We answer the first question with a 1{,}947-paper analysis of ICLR, ICML, and NeurIPS outcomes. We answer the second by releasing ResearchStudio-Idea's three skills and evaluating IdeaSpark-generated ideas with blind automated judges. The evaluation is scoped to the idea stage. It does not test implementation success, human peer review, or program-committee selection.

\subsection{Motivation}

Existing systems address important pieces of research ideation, but they leave a missing skill layer between outcome data and generation. End-to-end systems automate long research workflows \cite{lu2024aiscientist,agentlab2025,airesearcher2025,from_automation_2025}. Multi-agent and search systems explore larger proposal spaces \cite{deepideation2025,nova2024,wang2026flowpie,multiworkflow2026,alienscience2026}. Retrieval-augmented methods ground hypotheses in adjacent literature \cite{lewis2020rag,scimuse2024,kghyp2024,gao2026graphsofresearch}. Novelty tools, benchmarks, and pattern-induction methods check prior-art collision or name recurring research moves \cite{novbench2026,opennovelty2026,graphmind2025,litgrounded2025,schopf2026rinobench,wang2024scireasoning,moRI2026,motivgraph2025,navigating_ideation2026}. These components are complementary. What remains under-specified is how to turn them into interoperable skills, and how to turn conference outcomes into an executable ideation workflow: a reusable procedural object that helps an LLM organize a gap, instantiate a research strategy, distinguish it from neighboring work, and audit the proposal before implementation.

Research assistants are increasingly capable of retrieving literature, synthesizing evidence, generating candidate ideas, and iteratively refining proposals. The transition from retrieved evidence to actionable research opportunities, however, remains weakly structured. Researchers still need to decide which limitations constitute meaningful bottlenecks, which opportunities remain unresolved, and which directions are distinct enough from prior work to justify further investment.

Current evaluations of LLM-generated ideas make this gap concrete. A large blind study finds that LLM-generated NLP ideas can be judged more novel than expert ideas while being less feasible \cite{si2024llmideas}. A follow-up execution study reports that the gap becomes sharper once ideas are carried out \cite{si2026execution}. HindSight similarly finds that ideas rated as more novel by an LLM are less likely to match real future work \cite{hindsight2026}. These results do not imply that an idea-stage skill can solve execution. They do show that surface novelty, proposal fluency, and post-hoc novelty scoring are weak proxies for useful ideation.

A useful skill should instead expose how real papers connect a gap to a method. It should show how they assemble differentiation evidence, and how similar attempts fail before experiments begin. We call these recurring procedural objects \emph{ideation patterns}. A pattern is not merely a label such as ``decompose the problem'' or ``add supervision.'' It must state what kind of gap it applies to, how the method is constructed, what distinguishes it from nearby prior work, what evidence makes it credible, and where similar attempts fail.

The missing middle is therefore an outcome-grounded skill layer that connects research evidence, identified bottlenecks, reusable ideation patterns, and candidate research directions. Retrieved papers expose observations, limitations, reviewer concerns, and unresolved questions. Generated proposals require strategic moves that connect those observations to plausible directions. The missing layer is the mechanism that transforms evidence into opportunities and opportunities into auditable research proposals.

We need to know how these patterns recur across ML papers, how they compose inside contributions, and how they differ across accepted, high-citation, and rejected work. We also need to know whether they can be operationalized as reusable inference-time skills. ResearchStudio-Idea addresses this missing middle as a suite, and IdeaSpark tests the strongest version of the claim: whether search, pattern-guided generation, collision checking, and audit can be composed into one faithful ideation workflow rather than another generic brainstorming prompt or novelty scorer. Section~\ref{subsec:related-diff} gives the detailed comparison to current idea-generation methods.

\subsection{Approach overview}

\begin{figure}[!htbp]
\centering
\includegraphics[width=\textwidth]{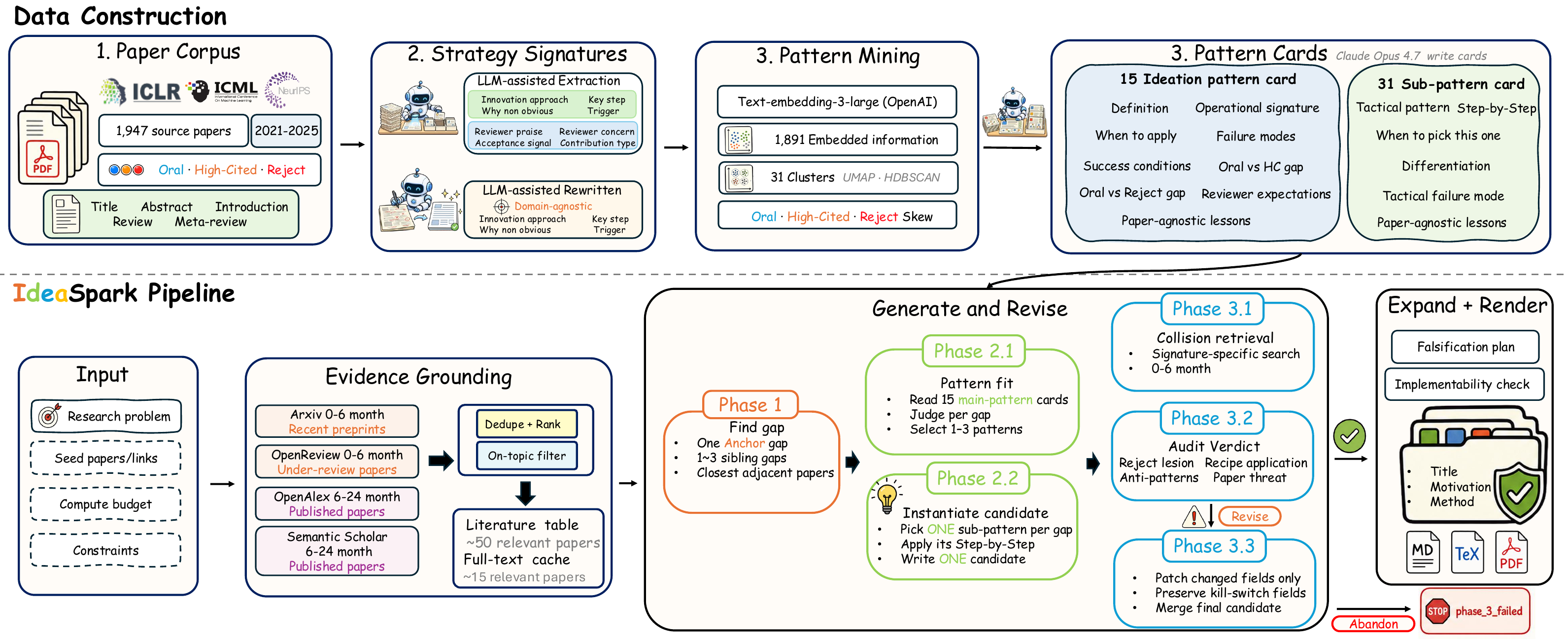}
\caption{IdeaSpark data-to-skill workflow. The upper band constructs reusable ideation assets from the 1{,}947-paper ICLR / ICML / NeurIPS corpus: papers are outcome-labeled, normalized into strategy signatures, mined into 31 sub-patterns, and induced into 15 operational pattern cards. The lower band shows how the skill uses those cards at inference time: evidence grounding and full-text retrieval feed a staged reasoning loop for bottleneck diagnosis, pattern-guided candidate generation, and collision/audit verdicts, followed by expansion, validation, and idea-card deliverables.}
\label{fig:pipeline}
\end{figure}

ResearchStudio-Idea packages early-stage ideation into three reusable skills. Paper-Search supplies the literature-grounding primitive. Scoop-Check supplies the claim-level prior-art review primitive. IdeaSpark induces ideation patterns from conference outcomes and packages them as the structured end-to-end ideation skill. Given a research problem and an evidence bundle, IdeaSpark performs evidence assessment, bottleneck diagnosis, innovation-pattern retrieval, candidate construction, differentiation analysis, collision retrieval, and outcome-informed auditing. The output is not only a candidate research direction, but also the supporting rationale, historical precedents, and diagnostic evidence needed to evaluate it. Figure~\ref{fig:pipeline} summarizes this data-to-skill flow from corpus construction to runtime use.

The empirical pipeline has four stages. First, we collect papers from ICLR, ICML, and NeurIPS using public OpenReview metadata and Semantic Scholar citation information \cite{openreview_api,semantic_scholar_api}. Each paper is labeled as Oral, High-Cited, Reject, or a combination when labels overlap. Second, we extract strategy-level innovation fields and rewrite them into domain-agnostic descriptions. This separates the methodological operation from topic vocabulary. As a result, clustering groups papers by research strategy rather than by application area. Third, we embed and cluster these strategy signatures, using a workflow inspired by scientific document representations and density-based topic discovery \cite{cohan2020specter,singh2022scirepeval,allenai_specter2_model_card,grootendorst2022bertopic,umap2018,hdbscan2017,openai_embeddings_docs}. Fourth, we induce higher-level ideation patterns from the fine-grained clusters and convert each pattern into an operational card.

The resulting cards are not taxonomy labels. Each card records the research situation the pattern addresses, the structure distilled from accepted examples, the differentiating mechanism relative to nearby prior work, the evidence expectations that make the claim inspectable, and the failure modes abstracted from rejected submissions. This contrast between success and failure offers insights beyond accepted-only inductions. In the runtime skill, the cards serve two roles. They guide structured idea generation, and they provide an audit reference for checking whether a candidate preserves the intended structure, avoids known failure modes, and names plausible prior-art threats. Generation and audit share the same corpus-derived object, while the paper's claim remains at the idea stage.

The IdeaSpark layer follows a model-agnostic, two-tier design. The runtime tier includes lean skill specification, phase prompts, schemas, retrieval hooks, and deterministic validators. The evidence tier consists of 15 ideation-pattern cards, 31 sub-pattern cards, a domain-by-pattern matrix, saturation records, and a corpus-derived failure-mode inventory. This separation follows recent skill-authoring guidance: keep the runtime lightweight, use progressive disclosure for detailed evidence, and validate phase boundaries with deterministic checks where possible \cite{anthropic_skill_bp_2025,schick2023toolformer,wei2022cot,yao2023react}. Consequently, the skill is faithful by construction rather than by instruction. Every significant claim either traces to a retrieved record or is explicitly marked as model-supplied. Retrieval gates and deterministic validators enforce this contract rather than relying on prompt compliance (\S\ref{subsec:faithfulness}).

\subsection{Main empirical findings}

This report supports five findings:

\textbf{First, the induced pattern space is compact but nontrivial.} From 31 fine-grained ideation sub-patterns, the pipeline induces 15 higher-level ideation patterns. The number 15 is not a canonical ontology of ML innovation. It is the operational granularity produced by the current extraction, abstraction, embedding, clustering, and induction pipeline. Because it comes from a single Opus~4.7 induction call, an inter-prompt, inter-seed, and inter-model stability study remains future work; we expect a comparable run to land at 12--18 categories with substantial overlap.

\textbf{Second, rejected and accepted papers often share the same high-level pattern space.} Reject-only re-clustering maps every rejected-paper cluster back onto the existing 15-pattern vocabulary, with no out-of-taxonomy bucket. This suggests that rejected submissions are most useful as contrastive evidence about weak instantiations, failure modes, and boundary cases, not as a separate negative strategy class.

\textbf{Third, multi-pattern composition is the norm.} A separate paper-level multi-label pass shows that papers commonly combine several ideation patterns rather than executing a single isolated move. In this corpus, $k=2$ is the modal composition size across Oral, HC, and Reject papers, with a $33.6\%$ tail at $k \geq 3$. This motivates the skill design choice to generate one to three pattern roles rather than selecting a single pattern as a recipe.

\textbf{Fourth, ideation patterns are broadly domain-covering but domain-conditional in effect.} The largest patterns appear across many induced research domains, which supports the use of domain-agnostic pattern cards. Acceptance and impact profiles, however, vary by domain-pattern cell. The skill therefore treats domain statistics as audit context, not as a deterministic generation prior.

\textbf{Fifth, endpoint idea quality is supported by an automated-judge study, but not by human review.} An automated-judge endpoint evaluation (Section~\ref{sec:evaluation}) indicates that IdeaSpark improves generated-idea quality over no-skill and generic-skill baselines. Acceptance-level claims still require a blind human study.

\subsection{Contributions}

This report makes three contributions.

\textbf{Outcome-grounded pattern map.} We provide a 1{,}947-paper corpus analysis spanning Oral, high-citation, and rejected papers from major ML conferences. The analysis induces 15 ideation patterns, 31 sub-patterns, and 28 research domains. It then measures pattern-level acceptance, citation, domain, conference, temporal, and multi-pattern composition profiles.

\textbf{Contrastive, failure-aware pattern cards.} We turn the accepted-versus-rejected contrast into card content. Each card pairs success conditions distilled from accepted work with failure modes distilled from rejected work. Accepted-only induction cannot supply this object. The empirical warrant is the reject-only mapping of \S\ref{sec:reject}: rejected papers do not occupy a separate strategy space, so their value is as contrastive evidence about weak instantiations and boundary cases.

\textbf{ResearchStudio-Idea skill suite and endpoint evaluation.} We package the work as three reusable skills: Paper-Search for literature grounding, Scoop-Check for prior-art collision checking, and IdeaSpark for end-to-end idea generation and audit. We also report a blind automated evaluation of IdeaSpark-generated ideas along quality and novelty axes.

\subsection{Report organization}

Section~\ref{sec:related} positions ResearchStudio-Idea and IdeaSpark against recent work on LLM-based scientific discovery, ideation, novelty assessment, benchmarks, and pattern induction. Sections~\ref{sec:data}--\ref{sec:methodology} describe the dataset, extraction procedure, clustering pipeline, and ideation-pattern induction. Sections~\ref{sec:analysis}--\ref{sec:trends} analyze acceptance profiles, domain structure, temporal trends, and conference-level variation. Section~\ref{sec:reject} validates the taxonomy against rejected papers. Section~\ref{sec:ablation} reports embedding and abstraction ablations. Section~\ref{sec:discussion} synthesizes the empirical takeaways. Section~\ref{sec:skill} specifies the IdeaSpark runtime and audit workflow within the ResearchStudio-Idea suite, and Section~\ref{sec:evaluation} reports the automated-judge evaluation of generated ideas. Section~\ref{sec:limitations} consolidates limitations and threats to validity across the corpus, skill, and evaluation. Section~\ref{sec:release} presents the released data, model, and skill-card artifacts.

\section{Related Work}
\label{sec:related}

The 2024--2026 literature on LLM research-idea generation has expanded rapidly. We group it into the families most relevant to IdeaSpark and survey each below. To keep the comparison coherent, we defer direct head-to-head differentiation to \S\ref{subsec:related-diff}.

\subsection{End-to-end ``AI scientist'' systems}
\label{subsec:related-scientist}

End-to-end systems automate the full research lifecycle. AI~Scientist~\cite{lu2024aiscientist} drafts, implements, evaluates, and writes up research ideas in a single loop, although its output was often judged incremental. AI-Researcher~\cite{airesearcher2025} extends this direction with structured architectures and decoupled training pipelines for autonomous innovation. Agent Laboratory~\cite{agentlab2025} starts from a human-supplied research idea and produces a research report and code, with variable human-in-the-loop control. Idea2Plan~\cite{idea2plan2025} bridges idea generation and research planning through ReAct-style scaffolding with arXiv tools. The Sakana AI Scientist v2~\cite{sakana_v2} adds evolutionary-search refinement, while Google AI co-scientist~\cite{ai_coscientist2025} uses a Gemini-based multi-agent architecture for hypothesis and proposal generation. Auto Research~\cite{autoresearch2025} spans literature review, ideation, innovation pattern, experimentation, paper writing, and rebuttal in a multi-agent framework.

A parallel critical literature documents why full-lifecycle autonomy remains fragile. Trehan et al.~\cite{trehan2026notscientists} distill six recurring failure modes from four autonomous ML-paper attempts: training-default bias, implementation drift under execution pressure, long-horizon context degradation, premature success declarations, thin domain knowledge, and weak experimental taste. Bisht et al.~\cite{bisht2026agentic} argue that current agentic scientists are not yet built for autonomy. They emphasize problem-selection bias, missing tacit laboratory knowledge, output homogenization from preference optimization, and the absence of experimental feedback loops. This line of critique motivates IdeaSpark's narrower scope: improve the idea before downstream execution begins.

\subsection{Multi-agent and search-based ideation}
\label{subsec:related-multiagent}

Multi-agent and search-based systems focus more directly on the ideation step. VirSci~\cite{virsci2024} and IRIS~\cite{iris2025} model scientific teamwork through role-separated agents, such as ideators, reviewers, and retrievers. Deep Ideation~\cite{deepideation2025} introduces an explore-expand-evolve workflow over a scientific concept network, with a critic engine trained on real reviewer feedback. Nova~\cite{nova2024} adds an iterative planning-and-search loop that retrieves external knowledge to enrich each candidate and improve novelty and diversity.

Other systems make search itself the core ideation mechanism. FlowPIE~\cite{wang2026flowpie} casts idea generation as a test-time evolution process driven by flow-guided Monte Carlo Tree Search over the literature. This design lets exploration and ideation co-adapt rather than run as separate stages. The multi-workflow benchmark~\cite{multiworkflow2026} compares reflection-based refinement, Sakana-style evolution, Google Co-Scientist multi-agent reasoning, GPT Deep Research recursive decomposition, and Gemini multimodal long-context workflows. It finds that decomposition-based and long-context workflows reach a mean novelty of 4.17/5. Alien Science~\cite{alienscience2026} formalizes a complementary creativity gap, \emph{cognitive availability}, and learns to sample coherent but cognitively unavailable directions from clustered ``idea atoms.''

\subsection{Pattern induction from conference outcomes}
\label{subsec:related-induction}

The closest line of work to ours induces innovation patterns from top-conference papers. Liu et al.~\cite{wang2024scireasoning} annotate Oral papers against a predefined 12-category taxonomy. MoRI~\cite{moRI2026} extracts \emph{motivation}\(\to\)\emph{innovation pattern} pairs from ICLR 2024--2025 accepted papers. It then trains a reasoning policy through supervised fine-tuning and controllable reinforcement learning on feasibility, novelty, and effectiveness. MotivGraph-SoIQ~\cite{motivgraph2025} builds a motivational knowledge graph and pairs it with Socratic dialogue for ideation. Navigating Ideation Space~\cite{navigating_ideation2026} decomposes scientific ideas into conceptual representations for positioning new ideas.

A parallel paradigm grounds idea generation in knowledge graphs mined from large literature corpora. SciMuse~\cite{scimuse2024} builds a knowledge graph from 58 million research papers. In a 100-research-group-leader evaluation, its cross-disciplinary graph paths help AI-generated ideas be perceived as novel and impactful. KG-grounded hypothesis generation~\cite{kghyp2024} retrieves relevant graph sub-structures to constrain LLM hypothesis output and reports improved novelty and plausibility on biomedical benchmarks. Graphs of Research~\cite{gao2026graphsofresearch} turns citation structure into a training signal. It extracts a per-seed-paper citation-evolution directed acyclic graph from the 2-hop reference neighborhood and fine-tunes an LLM to generate ideas that plausibly extend it.

\subsection{Novelty, evaluation, and benchmarks}
\label{subsec:related-novelty}

A separate family judges ideas rather than producing them. The first line focuses on novelty collision. NovBench~\cite{novbench2026} provides a large-scale benchmark for novelty evaluation, with 1{,}684 paper-review pairs and a four-dimensional framework covering relevance, correctness, coverage, and clarity. RINoBench~\cite{schopf2026rinobench} complements it with 1{,}381 expert-judged ideas focused on novelty judgment. It finds that LLM novelty verdicts can diverge substantially from expert gold, even when the model gives human-like rationales. We treat this as motivation for grounding novelty checks in retrieved evidence rather than model opinion. OpenNovelty~\cite{opennovelty2026} produces transparent evidence-based novelty reports for 500+ ICLR 2026 submissions. GraphMind~\cite{graphmind2025} ties paper key elements to retrieved literature for interactive novelty assessment. Literature-Grounded Novelty Assessment / Idea Novelty Checker~\cite{litgrounded2025} uses RAG with two-stage retrieve-then-rerank and literature-grounded rationales. Scideator~\cite{scideator2024} combines paper facets and novelty assessment modules for human-LLM ideation. Augmenting Research Ideation with Data~\cite{augment_ideation2026} integrates metadata and automated preliminary validation into ideation, reporting gains in feasibility and quality on social-science tasks.

The second line evaluates idea quality directly. Si et al.~\cite{si2024llmideas} ran a 100+-NLP-researcher blind study showing that LLM-generated ideas are rated \emph{more novel} but \emph{less feasible} than expert ideas. Their follow-up execution study~\cite{si2026execution} reports that AI ideas score significantly lower than human ideas after execution. The gap is therefore operational, not merely stylistic or novelty-related. AI Can Learn Scientific Taste~\cite{ai_taste2026} trains a taste model on accepted/rejected pairs. HindSight~\cite{hindsight2026} evaluates LLM-generated ideas via future-citation impact. Learning to Predict Future-Aligned Research Proposals~\cite{future_aligned2026} predicts whether a candidate proposal aligns with future research directions. MIRAI~\cite{li2026mirai} forecasts future paper impact from title and abstract, then uses that predictor to steer idea generation toward higher-impact directions.

The third line packages these judgments as standing benchmarks. IdeaBench~\cite{ideabench2024} evaluates research idea generation against referenced-work baselines. AI Idea Bench 2025~\cite{airesearchideas2025} curates 3{,}495 papers from ICLR top-2\% and CVPR highlights as ground truth. ResearchBench~\cite{researchbench2025} decomposes scientific discovery into inspiration retrieval, hypothesis composition, and hypothesis ranking. NewtonBench~\cite{newtonbench2025} evaluates scientific-law discovery on 324 physics tasks with counterfactual law shifts.

\subsection{Surveys and skill-engineering practice}
\label{subsec:related-substrate}

Two recent surveys map the broader landscape. ``From Automation to Autonomy''~\cite{from_automation_2025} organizes systems through a Tool / Analyst / Scientist taxonomy. ``Towards Scientific Intelligence''~\cite{sci_intel_survey2025} surveys architectures, benchmarks, and ethical questions. On the engineering side, Anthropic's skill-authoring practices~\cite{anthropic_skill_bp_2025} shape IdeaSpark's two-tier design: progressive disclosure, a lean top-level specification, deterministic routines for repetitive steps, and schema-validated phase boundaries. The embedding and clustering substrate is described where it is used, in \S\ref{sec:methodology} and \S\ref{sec:ablation}.

\subsection{How IdeaSpark differs from current idea-generation methods}
\label{subsec:related-diff}

Current idea-generation methods fall into three families. End-to-end ``AI scientist'' systems automate the full research lifecycle. Search-based systems generate many candidates and filter them with learned novelty, taste, or impact signals. Pattern- and knowledge-graph-induction methods mine accepted papers or large literature graphs for reusable structure. IdeaSpark draws ingredients from all three families: pattern induction, an accept/reject contrast, and retrieval-grounded generation. Its distinguishing move is to bind these ingredients onto one object.

That object is the non-parametric, data-induced pattern card. The same card that helps select a move for a bottleneck is also used to audit the generated candidate. This differs from searching a candidate space and ranking ideas by a learned signal. It also differs from fitting a parametric motivation\(\to\)method policy, where generation and checking can become separate procedures.

Three properties follow from this design. First, the cards are induced from a three-way Oral / HC / Reject contrast, so they carry an explicit \emph{failure} signal that accepted-only induction and concept-level knowledge graphs do not. Second, the skill stops at one reviewer-facing idea card rather than running the full research lifecycle. This choice reflects the premise that idea quality is an upstream bottleneck for many downstream failures. Third, the audit checks the candidate against the corpus-derived failure modes of the pattern that produced it, together with retrieved prior art. The knowledge that generates an idea and the knowledge that checks it are therefore one object, not two disconnected models. Execution-level gaps are surfaced rather than masked by surface novelty.

\section{Dataset Construction}
\label{sec:data}

\subsection{Scope and labeling}

We collect papers from ICLR, ICML, and NeurIPS for 2021--2025. Collection uses the OpenReview API~\cite{openreview_api} (v1 for ICLR 2021--2023; v2 for ICLR 2024+, ICML 2023+, and NeurIPS 2023+) and Semantic Scholar citation metadata~\cite{semantic_scholar_api}. ICLR 2020 is excluded from this release because its review-data schema differs from later years. For each paper, we retain the title, abstract, author list, OpenReview id, decision string, available review fields, available meta-review, and Semantic Scholar citation count. The review fields include rating, confidence, soundness, contribution, summary, strengths, weaknesses, and questions.

The three labels operationalize the contrast at the heart of this work. Oral marks what program committees most value. High-Cited marks what the community adopts. Reject supplies the failure signal that accepted-only induction cannot provide. Each paper receives one or more labels, and labels may overlap. In this corpus, 49 papers are simultaneously Oral and High-Cited.

\textbf{\textcolor{cblue}{Oral}} (1{,}014 papers): papers whose venue decision is Oral, marking program-committee preference.

\textbf{\textcolor{cgold}{High-Cited (HC)}} (260 papers, \(HC_{\mathrm{all}}\)): top-30 most-cited papers per venue-year by Semantic Scholar citation count, representing community adoption. For 2025, we use the top-10 most-cited papers because citation counts are still accumulating. The high-citation pool overlaps Oral: \(HC_{\mathrm{all}}=260\), Oral\(\cap\)HC=49, and \(HC_{\mathrm{disjoint}}=211\). We use \(HC_{\mathrm{all}}\) for citation-label coverage and \(HC_{\mathrm{disjoint}}\) when analyses need mutually exclusive Oral, HC, and Reject classes.

\textbf{\textcolor{cred}{Reject}} (722 papers): papers with an explicit Reject decision and a non-trivial review set, representing concrete failure signal.

Table~\ref{tab:data} shows the venue-year breakdown. The per-venue counts reflect two independent factors.

First, \textbf{venue openness}: ICLR is OpenReview-native and publishes both accept and reject decisions with full review threads. Its Reject pool is therefore the deepest. ICML and NeurIPS expose decisions on OpenReview only from 2023 onward. ICML 2023--2024 in particular released very few rejected submissions, which is why those rows show 0--3 in the Reject column.

Second, \textbf{review accessibility}: even when a decision is published, the meta-review and reviewer comments may be withheld or formatted such that automatic parsing fails. This is why metadata coverage (Section~\ref{subsec:meta-coverage}) shows only 340 of the 1{,}947 papers carrying a meta-review, even though all 1{,}947 have decisions.

Finally, NeurIPS 2025 contributes no HC papers. As the most recent venue (December 2025), its citation counts had not accumulated enough at collection time for even a top-10 cut to be meaningful. These asymmetries do not affect the clustering stage, but we treat them carefully in the quantitative analysis (Section~\ref{sec:analysis}).

\begin{table}[!htbp]
\centering
\caption{Dataset composition. Each column counts a paper if it carries that label; the same paper can appear in multiple columns. HC denotes \(HC_{\mathrm{all}}\). The label-inclusive Total is $1{,}996$, giving $1{,}947$ unique papers after the $49$ Oral$\cap$HC overlaps. Per-class disjoint pools used in Section~\ref{sec:analysis} are $1{,}014$ Oral, $211$ \(HC_{\mathrm{disjoint}}\), and $722$ Reject. ICLR 2020 is excluded.}
\label{tab:data}
\small
\begin{tabular}{llrrrr}
\toprule
\textbf{Venue} & \textbf{Year} & \textbf{\textcolor{cblue}{Oral}} & \textbf{\textcolor{cgold}{\(HC_{\mathrm{all}}\)}} & \textbf{Oral$\cap$HC} & \textbf{\textcolor{cred}{Reject}} \\
\midrule
ICLR    & 2021 &  52 & 30 & 5 &   0 \\
ICLR    & 2022 &  49 & 30 & 5 &   0 \\
ICLR    & 2023 &  85 & 30 & 6 & 143 \\
ICLR    & 2024 &  63 & 30 & 1 & 138 \\
ICLR    & 2025 & 186 & 10 & 4 & 128 \\
ICML    & 2023 & 147 & 30 & 9 &   3 \\
ICML    & 2024 & 135 & 30 & 9 &   0 \\
ICML    & 2025 &  99 & 10 & 3 &  65 \\
NeurIPS & 2023 &  69 & 30 & 4 &  94 \\
NeurIPS & 2024 &  60 & 30 & 3 & 101 \\
NeurIPS & 2025 &  69 &  0 & 0 &  50 \\
\midrule
\textbf{Total} &  & \textbf{1{,}014} & \textbf{260} & \textbf{49} & \textbf{722} \\
\bottomrule
\end{tabular}
\end{table}

\subsection{Metadata coverage}
\label{subsec:meta-coverage}

After merging and deduplicating by OpenReview submission id, 1{,}947 unique papers remain. All 1{,}947 carry the eight base fields. The four abstracted fields (Section~\ref{sec:extract}) are populated with non-empty content for 1{,}891 papers. The remaining 56 papers are dropped from the embedding pool (\S\ref{sec:clustering}).

Coverage of the initial merged metadata varies by field. In total, 1{,}074 papers carry the full abstract, 716 carry parsed reviews, 682 carry an extracted introduction, and 340 carry a meta-review. For card generation, we run a separate review-recovery pass over public OpenReview threads. The resulting card-level review-evidence coverage is reported in \S\ref{sec:discussion}. Papers without recoverable review text are still included in signature extraction. Their review-derived fields, however, are treated as low-evidence auxiliary fields rather than evidence grounded in reviewer text.

\subsection{Convention for in-paper paper references}
\label{subsec:paper-id-convention}

Throughout this paper, we reference corpus papers by an internal identifier of the form \texttt{<VENUE>\_<YEAR>\_<NNNN>}. Examples include \texttt{ICLR\_2022\_0094} and \texttt{NeurIPS\_2023\_0753}. Each identifier resolves uniquely to an OpenReview submission via the \texttt{openreview\_id} field in \texttt{data/dataset/all\_records\_full.json}. After consulting the dataset, a paper can be looked up at \texttt{https://openreview.net/forum?id=<openreview\_id>}.

Of the 1{,}947 corpus papers, approximately 1{,}088 are referenced at least once in this report. They usually appear as evidence under a methodology card's success-condition or failure-mode list. We do not include all corpus papers as bibliography entries. A complete corpus bibliography would obscure the second-order related-work citations that constitute Section~\ref{sec:related}. The prepared dataset is therefore the authoritative bibliographic resolution layer for any \texttt{paper\_id} appearing in this report.

When a specific corpus paper is discussed substantively in body text, we provide a proper bibliography entry. Examples include the diffusion-ELBO paper~\cite{kingma2023diffusion} = \texttt{NeurIPS\_2023\_0753}, LP-FT~\cite{kumar2022lpft} = \texttt{ICLR\_2022\_0094}, the DP continual-observation paper~\cite{henzinger2023dpcontinual} = \texttt{ICML\_2023\_0551}, EigenGame~\cite{gemp2021eigengame} = \texttt{ICLR\_2021\_0018}, and SAM-as-Bayes~\cite{mollenhoff2023sambayes} = \texttt{ICLR\_2023\_0331}.

\section{Two-Stage Innovation-Signature Extraction}
\label{sec:extract}

To convert each paper into a form suitable for strategy-level clustering, we extract a twelve-field \emph{innovation signature}. Eight base fields capture the paper in its own terms. Four describe the innovation mechanism, and four record review and outcome signals. A second stage rewrites the four mechanism fields into domain-agnostic form. Only these four rewritten fields are embedded for clustering. The eight base fields are retained for extraction and later card evidence.

\subsection{Stage 1: Eight base fields}

For Stage~1, we feed Claude Sonnet~4.6~\cite{anthropic_models_docs} the paper's title, available abstract, available introduction, available reviews, and available meta-review. Introductions are extracted from the PDF when recoverable. The model then returns the following structured fields:

\begin{enumerate}[nosep, leftmargin=1.5em]
\item \texttt{innovation\_approach}: one sentence describing \emph{how} this paper innovates as a reasoning strategy.
\item \texttt{key\_step}: the single most critical reasoning step from diagnosis to solution.
\item \texttt{why\_non\_obvious}: the cognitive barrier that blocked others from this approach.
\item \texttt{trigger\_condition}: domain-agnostic situation where this strategy applies, phrased as ``When X, apply Y to achieve Z.''
\item \texttt{reviewer\_praise}: three specific points the reviewers praised.
\item \texttt{reviewer\_concern}: three specific reviewer concerns.
\item \texttt{acceptance\_signal}: one sentence explaining what made the paper Oral / cited / rejected.
\item \texttt{contribution\_type}: one of \{theoretical, methodological, empirical, benchmark, system\}.
\end{enumerate}

The first four fields are the most strategy-bearing, so Stage~2 rewrites only those. The last four supply review and outcome evidence used later in card construction. Extraction runs under a JSON schema that enforces non-empty lists of length 3 for the two review-derived fields. All 1{,}947 papers pass Stage~1 extraction.

\subsection{Stage 2: Four domain-agnostic rewrites}

A direct embedding of paper descriptions tends to latch onto domain nouns, such as ``Transformer'', ``diffusion'', ``image'', or ``molecule''. This produces topic clusters rather than strategy clusters. We address this failure mode with a second Claude Sonnet~4.6 pass. It rewrites the first four base fields into strategy-only equivalents. The prompt instructs the model to replace domain nouns with generic placeholders, adopt imperative phrasing (``Audit \ldots'', ``Transfer \ldots'', ``Design \ldots''), preserve the mechanism, and drop application detail. The resulting four fields are:

\begin{enumerate}[nosep, leftmargin=1.5em]
\item \texttt{abstract\_strategy} $\leftarrow$ \texttt{innovation\_approach}
\item \texttt{abstract\_key\_step} $\leftarrow$ \texttt{key\_step}
\item \texttt{abstract\_why\_non\_obvious} $\leftarrow$ \texttt{why\_non\_obvious}
\item \texttt{abstract\_trigger\_condition} $\leftarrow$ \texttt{trigger\_condition}
\end{enumerate}

A representative contrast illustrates the shift. For the paper \emph{High Fidelity Speech Synthesis with Adversarial Networks}, \texttt{innovation\_approach} reads ``\textit{They solved a domain transfer problem by borrowing the PatchGAN multi-scale discriminator strategy from image GANs and re-instantiating it at multiple temporal windows in audio, adding conditioning signals to close the fidelity gap}''. The corresponding \texttt{abstract\_strategy} reads ``\textit{Transfer a multi-scale discrimination strategy from one data modality to another by re-instantiating it at the appropriate domain-specific granularity, then add conditioning signals to close the remaining fidelity gap.}'' The generic version is what we feed into the embedding model.

The contribution of this abstraction step to strategy-level clustering is quantified in the ablation of \S\ref{sec:ablation}. All four abstract fields are returned for the full corpus. Of the 1{,}947 papers, $1{,}891$ carry non-empty content in all four fields. The remaining $56$ have at least one empty field and are dropped from the embedding pool in \S\ref{sec:clustering}.

\section{Unsupervised Pattern Discovery}
\label{sec:clustering}

\subsection{Embedding}

For each paper, we concatenate the four abstracted strategy fields: \texttt{abstract\_strategy}, \texttt{abstract\_key\_step}, \texttt{abstract\_why\_non\_obvious}, and \texttt{abstract\_trigger\_condition}. Each field is prefixed as \texttt{field: content}. The average concatenated length is 1{,}185 characters. After the Stage~2 rewrite of \S\ref{sec:extract}, 1{,}891 of 1{,}947 papers carry all four fields. We embed these records with OpenAI \texttt{text-embedding-3-large} (3{,}072 dimensions) and L2-normalize the vectors.

\subsection{Clustering}

UMAP~\cite{umap2018} reduces the 3{,}072-dimensional cosine-normalized space to 10 dimensions. We use $n\_\text{neighbors}{=}15$, $\min\_\text{dist}{=}0$, and seed 42. HDBSCAN~\cite{hdbscan2017} then clusters the reduced space with $\min\_\text{samples}{=}\lceil \min\_\text{cluster\_size}/3 \rceil$ and \texttt{cluster\_selection\_method}=\texttt{eom}. We sweep $\min\_\text{cluster\_size}\in\{10,15,20,25,30,40\}$; Table~\ref{tab:sweep} reports the full sweep.

\begin{table}[!htbp]
\centering
\caption{HDBSCAN sweep on the 1{,}891-paper embedding. Selected row bolded.}
\label{tab:sweep}
\small
\begin{tabular}{cccccc}
\toprule
$\min\_\text{cluster\_size}$ & $k$ & $n_\text{unclustered}$ & \% unclustered & silhouette \\
\midrule
\textbf{10} & \textbf{31} & \textbf{902} & \textbf{47.7} & \textbf{0.584} \\
15 & 19 & 751 & 39.7 & 0.527 \\
20 & 16 & 879 & 46.5 & 0.566 \\
25 & 12 & 923 & 48.8 & 0.561 \\
30 &  8 & 853 & 45.1 & 0.505 \\
40 &  6 & 849 & 44.9 & 0.474 \\
\bottomrule
\end{tabular}
\end{table}

We select $\min\_\text{cluster\_size}{=}10$. This setting identifies 31 ideation sub-patterns and gives the strongest silhouette in the sweep ($0.584$, computed over clustered points). The tradeoff is that $902 / 1{,}891 = 47.7\%$ of papers remain unclustered. Coarsening $\min\_\text{cluster\_size}$ does not recover this fraction; it stays in the $40$--$49\%$ range across the sweep. The unclustered points are therefore far from all density peaks, not fine-granularity artifacts.

\paragraph{What ``unclustered'' actually means (and why we don't call it ``noise'').} HDBSCAN normally labels points outside any density peak as ``noise''. We instead use ``unclustered'' because the label is geometric, not substantive. It marks where a paper sits in the embedding; it does not mean the strategy is weak, idiosyncratic, or outside the framework.

The flagged $47.7\%$ are \emph{not} strategy-empty. All $1{,}891$ embedded papers carry full content in the four \texttt{abstract\_*} fields, and the unclustered and clustered groups have comparable median lengths. The label reflects embedding geometry at the chosen \textit{mcs} threshold. A paper lands ``unclustered'' when its strategy embedding falls between two or three modal clusters, often because it executes several ideation patterns in roughly equal measure. In short, ``unclustered'' means \emph{between modal clusters}, not \emph{outside the strategy space}. The 15-pattern framework works at the paper level through independent multi-label tagging (\S\ref{subsec:multilabel}), so it covers unclustered and clustered papers equally.

\subsection{Cluster inventory}

Cluster sizes (excluding unclustered) range from 13 (the smallest valid size) to 86, with mean 31.9 and median 22. The five largest clusters are:

\begin{itemize}[nosep, leftmargin=1.5em]
\item \textbf{C02} (\(n{=}86\)): \emph{Controlled Diagnostic Construction with One-Variable-Isolated Items}, under \emph{Design a Confound-Isolating Diagnostic}.
\item \textbf{C18} (\(n{=}82\)): \emph{Substrate Adaptation Across Modalities for a Mature Recipe}, under \emph{Unify Heterogeneous Inputs into One Space}.
\item \textbf{C06} (\(n{=}59\)): \emph{Probabilistic-Objective Unification via Equivalence Proof}, under \emph{Prove Equivalence to Unify}.
\item \textbf{C14} (\(n{=}50\)): \emph{All-Pairs Attention Operator Replacement}, under \emph{Substitute the Operator or Representation}.
\item \textbf{C05} (\(n{=}50\)): \emph{Identifiability-Theorem Pivot on the Limiting Assumption}, under \mbox{\emph{Audit and Pivot an Assumption}}.
\end{itemize}

Each cluster is automatically labeled by an Opus~4.7 pass. The pass reads the cluster's eight centroid-nearest representative papers and produces a 2--6 word label plus a one-sentence description. A second Opus~4.7 pass then produces a level-2 \emph{disambiguation card} for each cluster. The resulting tactic-level patterns are used in \S\ref{subsec:multistrat}. Each disambiguation card has six panels:

\begin{enumerate}[nosep, leftmargin=1.5em]
\item \emph{tactical\_pattern}: the cluster's specific reasoning move at the granularity below the level-1 ideation pattern.
\item \emph{Step-by-Step}: a 5-step abstract structural-move recipe distilled from the cluster's accepted examples, with reject-derived boundaries embedded.
\item \emph{differentiation\_within\_parent}: what distinguishes this cluster from its siblings under the same parent.
\item \emph{when\_to\_pick\_this\_one}: the situational trigger conditions under which this sub-pattern is the right choice.
\item \emph{tactical\_failure\_mode}: the cluster-specific failure shape that recurs in its Reject papers.
\item \emph{Examples}: a set of \emph{paper-agnostic} Oral and Reject lessons. No per-paper citations are used. The lessons are abstract distillations of the cluster's success and failure shapes, so that Phase~2.2 generation applies the abstract pattern to a new gap rather than mimicking specific papers.
\end{enumerate}

We use the 31-cluster inventory as raw material for ideation pattern induction in the next section.

\subsection{Per-cluster acceptance composition}
\label{subsec:clusterskew}

\begin{figure}[!htbp]
\centering
\includegraphics[width=\textwidth]{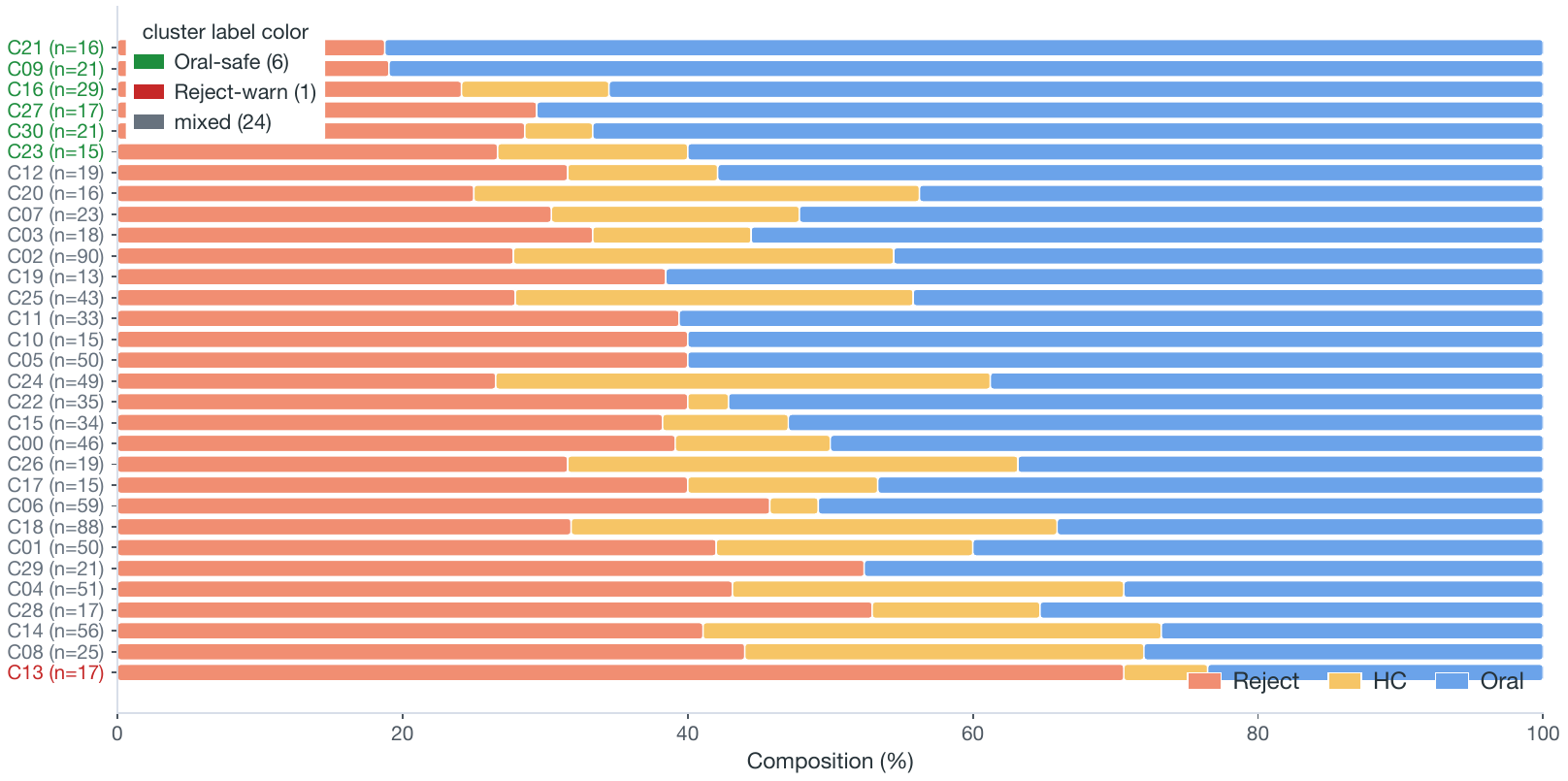}
\caption{Acceptance composition of the 31 clusters, sorted by Oral rate among $O{+}R$. Six clusters clear the 65\% Oral threshold; one clears the 65\% Reject threshold. The remaining 24 are mixed. Cluster labels are colored by risk flag (green = Oral-safe, red = Reject-warn, gray = mixed; deliberately off the bar palette); the threshold uses $p_O$ among $O{+}R$, so it is shown via label color rather than an $x$-axis line, because the bars are $O/HC/R$ shares that include HC.}
\label{fig:cluster_skew}
\end{figure}

\begin{figure}[!htbp]
\centering
\begin{subfigure}[b]{0.52\textwidth}
\includegraphics[width=\textwidth]{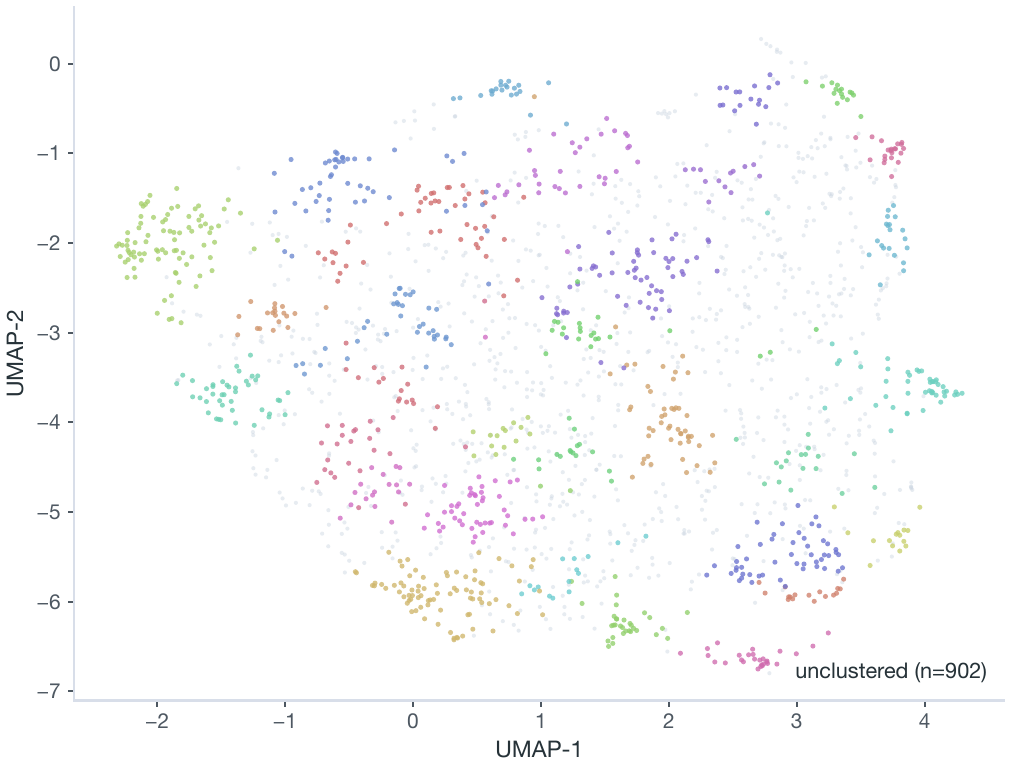}
\caption{31 clusters, colored; unclustered in gray.}
\end{subfigure}
\hfill
\begin{subfigure}[b]{0.44\textwidth}
\includegraphics[width=\textwidth]{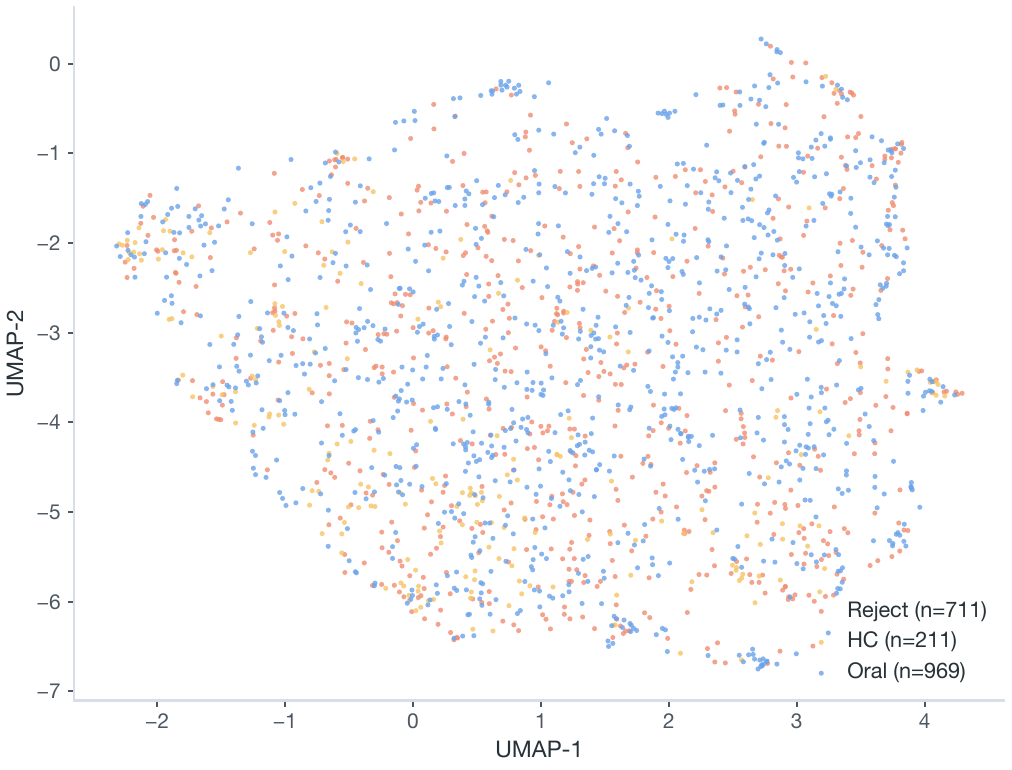}
\caption{Colored by label (Oral / HC / Reject).}
\end{subfigure}
\caption{2D UMAP projection of the 1{,}891-paper embedding. Oral, HC, and Reject papers are interleaved across essentially every cluster: the embedding captures strategy, not acceptance.}
\label{fig:umap}
\end{figure}

Although the embedding interleaves Oral, HC, and Reject papers across most clusters (Fig.~\ref{fig:umap}b), the cluster-level proportions are non-uniform. Figure~\ref{fig:cluster_skew} shows each cluster's Oral / HC / Reject composition, sorted by Oral rate. Six clusters clear the Oral-safe threshold ($p_O{\geq}65\%$ among $O{+}R$, $n_O{\geq}5$). One cluster crosses the Reject-warn threshold ($p_O{\leq}35\%$ among $O{+}R$, $n_R{\geq}5$). Each cluster below is written as Oral/HC/Reject counts and Oral share among $O{+}R$. The flagging is informational. It is surfaced to human artifact readers and the audit layer, but is never used to pre-filter retrieval or generation:

\begin{itemize}[leftmargin=1.4em, topsep=2pt, itemsep=2pt]
\item \textbf{Oral-safe}: \textbf{6 clusters}. These tactics, when properly executed, land at the top of the acceptance distribution. The five sharpest by Oral share among $O{+}R$ are C21 \emph{Sequential-Control Re-Casting} under \emph{Reframe as a Solvable Object} (13/0/3, 81\%), C09 \emph{Low-Sensitivity Operator Substitution} under \emph{Substitute the Operator or Representation} (17/0/4, 81\%), C16 \emph{Symmetry-Group Construction} under \emph{Encode Structure by Construction} (19/3/7, 73\%), C27 \emph{Intractable-Problem Re-Casting} under \emph{Reframe as a Solvable Object} (12/0/5, 71\%), and C30 \emph{Curvature-Object Substitution} under \emph{Substitute the Operator or Representation} (14/1/6, 70\%).
\item \textbf{Reject-warn}: \textbf{1 cluster}. C13 \emph{Relational-Topology Encoded as Structure} under \emph{Encode Structure by Construction} (4/1/12, 25\%) is the only cluster whose reject share dominates by the $n_R{\geq}5$ threshold. Reviewers consistently flag this cluster's papers as proposing a topology-as-prior architecture without empirically demonstrating that the topology is doing identifiable work over standard alternatives.
\item \textbf{Mixed}: the remaining 24 clusters. These tactics produce both Oral and Reject papers in comparable shares. They are the bulk of the working space, and acceptance hinges on the per-paper success conditions and failure modes characterized in the methodology cards (Section~\ref{subsec:cards}).
\end{itemize}

The per-cluster labels and full counts are in \texttt{clustering/extra/cluster\_composition.csv}.

\paragraph{These flags are risk signals, not recommendation labels.} \emph{Oral-safe} does not mean ``preferred''; \emph{Reject-warn} does not mean ``forbidden''. They are descriptive risk statistics with limited sample sizes. They are also sensitive to year, conference, and topic saturation. A Reject-warn cluster's rejected papers may have failed on execution rather than tactic choice. We therefore expose these flags to human artifact readers and to the audit layer as informational risk signals. We do not gate retrieval, generation, or recommendation on them. A user whose problem genuinely matches a Reject-warn cluster's tactic can still proceed, but the sub-pattern card informs them of the documented failure modes.

\section{Ideation-Pattern Induction}
\label{sec:methodology}

Fine-grained ideation sub-patterns answer ``what specific move did this paper make?'' They do not answer the higher-level question: ``what kind of move is that?'' To induce that level, we ask Claude Opus~4.7 to build a Level-1 ideation-pattern taxonomy from the 31 clusters. The prompt imposes four constraints. The taxonomy must contain 6--18 entries. Each entry must describe a reusable reasoning strategy rather than a domain or venue. Each entry must include a definition, an operational signature, and a when-to-apply clause. Finally, every cluster must map to one primary ideation pattern, with an optional secondary pattern.

Opus returns a 15-ideation pattern taxonomy (Table~\ref{tab:tax}) and a 31-cluster mapping in a single structured call. We make no edits to either. The 15 ideation patterns are:

\begin{table}[!htbp]
\centering
\caption{The 15 induced ideation patterns. \(n_\text{cl}\)=fine-grained clusters mapped as primary; Oral / HC / Reject counts are paper-level primary-mapping counts by cluster-level inheritance ($989$ of the $1{,}891$ embedded papers; the rest are HDBSCAN-unclustered and receive their pattern via the multi-label pass of \S\ref{subsec:multilabel}). $n$ is the unique paper count under cluster-level primary; because $49$ papers are both Oral and High-Cited, a row's Oral${+}$HC${+}$Reject can exceed $n$. Figure~\ref{fig:hierarchy_donut} sizes each wedge by the Oral${+}$HC${+}$Reject sum.}
\label{tab:tax}
\small
\begin{tabular}{p{7cm} c rrrr}
\toprule
\textbf{Ideation pattern (id)} & \textbf{\(n_\text{cl}\)} & \textbf{Oral} & \textbf{HC} & \textbf{Reject} & \textbf{$n$} \\
\midrule
Audit and Pivot an Assumption \texttt{(assumption\_audit\_and\_pivot)}                & 6 & 94 & 11 & 79 & 181 \\
Substitute the Operator or Representation \texttt{(architectural\_operator\_substitution)} & 4 & 57 & 21 & 39 & 109 \\
Liberate a Fixed Generative Component \texttt{(generative\_process\_redesign)}        & 3 & 46 & 22 & 30 & 94 \\
Design a Confound-Isolating Diagnostic \texttt{(controlled\_diagnostic\_design)}      & 1 & 41 & 24 & 25 & 86 \\
Unify Heterogeneous Inputs into One Space \texttt{(unify\_into\_shared\_representation)} & 1 & 30 & 30 & 28 & 82 \\
Reframe as a Solvable Object \texttt{(reframe\_as\_solvable\_object)}                 & 3 & 48 &  5 & 26 & 79 \\
Manufacture the Supervisory Signal \texttt{(self\_supervised\_signal\_engineering)}   & 3 & 26 & 17 & 24 & 66 \\
Encode Structure by Construction \texttt{(structural\_prior\_encoding)}               & 3 & 30 &  9 & 23 & 61 \\
Prove Equivalence to Unify \texttt{(algebraic\_equivalence\_unification)}             & 1 & 30 &  2 & 27 & 59 \\
Decompose for Differentiated Treatment \texttt{(heterogeneous\_decomposition)}        & 1 & 15 & 14 & 22 & 47 \\
Decompose and Delegate to Solvers \texttt{(decompose\_and\_delegate)}                 & 1 & 19 & 12 & 12 & 42 \\
Relax Discrete Search to Continuous \texttt{(relax\_discrete\_search\_to\_continuous)} & 1 & 20 &  1 & 14 & 35 \\
Adapt by Conditioning, Not Retraining \texttt{(adapt\_via\_conditioning)}             & 1 & 10 &  2 &  6 & 18 \\
Characterize a Limit, Then Surpass It \texttt{(characterize\_limit\_then\_surpass)}   & 1 &  9 &  0 &  6 & 15 \\
Design a Property-Targeting Pretext Objective \texttt{(targeted\_self\_supervised\_objective)} & 1 &  7 &  2 &  6 & 15 \\
\midrule
\textit{No cluster primary (HDBSCAN-unclustered)} &  & 532 & 88 & 355 & 902 \\
\bottomrule
\end{tabular}
\end{table}

Each ideation pattern is specified at the level needed for use in the skill, with its full definition, operational signature, and when-to-apply clause given in the per-pattern card.

\paragraph{Non-mutually-exclusive operators.} The 15 ideation patterns are \emph{not} mutually exclusive categories. Several pairs naturally overlap. \emph{Audit and Pivot an Assumption} can overlap with \emph{Prove Equivalence to Unify}, because an equivalence proof often relaxes a load-bearing assumption. \emph{Substitute the Operator or Representation} can overlap with \emph{Encode Structure by Construction}, because a structured operator both substitutes and encodes. \emph{Manufacture the Supervisory Signal} can overlap with \emph{Design a Property-Targeting Pretext Objective}, because the latter is a structured form of the former.

The cluster-level mapping forces one primary pattern per cluster for organization. The papers themselves often execute several operators in composition, and the paper-level multi-label tagging in \S\ref{subsec:multilabel} confirms this empirically. The right reading is therefore compositional: the 15 patterns are ideation operators, and a paper is represented as a composition of operators rather than assigned to one exclusive class.

\paragraph{A note on small-sample ideation patterns.} Three ideation patterns have small per-paper-primary populations under cluster-level inheritance: \emph{Adapt by Conditioning, Not Retraining} ($n{=}18$), \emph{Characterize a Limit, Then Surpass It} ($n{=}15$), and \emph{Design a Property-Targeting Pretext Objective} ($n{=}15$). These are valid operators because the underlying clusters are statistically distinct in embedding space. They are also \textbf{low-support operators}: their per-pattern acceptance statistics rest on thin evidence and should not be used as strong recommendation priors. The corresponding analytical cards therefore carry an internal \texttt{confidence: low} flag.

The runtime prompt cards do not expose raw counts or confidence flags to the model, because doing so induced distribution-bias homogenization toward high-$n$ patterns in earlier designs. Low-support patterns remain available when the user's problem genuinely fits them. They are neither privileged nor down-weighted by the skill's selection step. The corpus-side flag is retained here for analytical transparency.

\paragraph{Plain-language aliases.} The ideation pattern names use compact technical phrasing for analytical precision: for example, \emph{Audit and Pivot an Assumption} and \emph{Substitute the Operator or Representation}. Each pattern also has a \emph{plain-language alias} for user-facing skill output. This keeps the framework legible to non-specialists. Examples include \emph{Audit and Pivot an Assumption} $\to$ ``Audit the load-bearing assumption and pivot'', \emph{Reframe as a Solvable Object} $\to$ ``Reformulate the unsolved as a solvable object'', and \emph{Substitute the Operator or Representation} $\to$ ``Substitute the operator or representation''.

\subsection{From definition to operational card}
\label{subsec:cards}

A definition and operational signature tell a researcher \emph{what} an ideation pattern is. They do not explain \emph{when it lands and when it fails}. To capture that distinction, we run a second Opus~4.7 pass for each ideation pattern. The input includes all Oral, HC, and Reject papers assigned to the pattern, their four abstracted fields, and the available review and meta-review evidence. Review evidence is recovered for 83\% of the corpus overall; details are in \S\ref{sec:discussion}. The pass returns a structured card with seven panels:

\begin{enumerate}[nosep, leftmargin=1.5em]
\item \emph{success\_conditions}: cross-Oral patterns of what makes the pattern land.
\item \emph{failure\_modes}: cross-Reject patterns of how it fails.
\item \emph{oral\_reject\_gap}: the prose contrast between the two pools.
\item \emph{oral\_hc\_gap}: the prose contrast between PC-favored and community-favored execution.
\item \emph{reviewer\_expectations}: bullets tagged by provenance (\texttt{[oral\_reviews]}, \texttt{[reject\_reviews]}, \texttt{[both]}).
\item \emph{cognitive\_barriers}: the field-level blind spots that make the pattern non-obvious.
\item \emph{representative\_examples}: 6--8 paper-anchored exemplars (Oral and Reject), each with a one-line lesson.
\end{enumerate}

Evidence-bearing fields must cite paper IDs. This rule applies to \emph{success\_conditions}, \emph{failure\_modes}, \emph{oral\_reject\_gap}, \emph{oral\_hc\_gap}, \emph{reviewer\_expectations}, and \emph{representative\_examples}. The model is also instructed to set \texttt{confidence=low} when sample sizes or review coverage fall below thresholds, and to refuse the Oral-vs-HC contrast when \(n_{HC}<5\).

The 15 resulting cards are the operational substrate of the skill. Definitions, operational signatures, and when-to-apply clauses support Phase~2.1 pattern fit. Sub-pattern tactical recipes support Phase~2.2 instantiation. Failure modes, Oral-vs-Reject gaps, representative examples, and reviewer expectations support Phase~3 audit and Phase~4 explanation. They are not used as acceptance-rate priors or ranking signals.

The cards' \texttt{Examples} panel surfaces \emph{paper-agnostic Oral and Reject lessons}: bullets distilled from the cluster's acceptance pool, without per-paper citations. This design keeps Phase~2.2 focused on applying the abstract pattern to a new gap rather than mimicking any single corpus paper. App.~\ref{app:example-cards} reproduces the \emph{Audit and Pivot an Assumption} card---paired with one of its sub-pattern cards---as a high-coverage corpus-to-card example.

\subsection{Coverage and granularity}

The 15 ideation patterns have two coverage figures, and we report both because they answer different questions:
\begin{itemize}[leftmargin=1.4em, topsep=2pt, itemsep=2pt]
\item \textbf{Cluster-level coverage}: 989 of 989 clustered papers ($100\%$ of clustered; $52\%$ of the 1{,}891-paper embedding pool) receive a primary ideation pattern by inheritance from their ideation sub-pattern. Every cluster maps to exactly one of the 15 primary patterns; there is no out-of-taxonomy residual.
\item \textbf{Paper-level multi-label coverage}: the independent multi-label tagging pass (\S\ref{subsec:multilabel}) assigns one or more ideation pattern labels to \emph{all} papers including the $902$ HDBSCAN-unclustered ones. The 15-ideation pattern framework therefore covers the corpus completely at the paper level, even though only $52\%$ of papers are assignable via cluster-level inheritance.
\end{itemize}
The distribution has a long tail. By cluster-level primary count, the top three ideation patterns account for 384 papers: \emph{Audit and Pivot an Assumption} ($n{=}181$), \emph{Substitute the Operator or Representation} ($n{=}109$), and \emph{Liberate a Fixed Generative Component} ($n{=}94$). The bottom three account for 48 papers: \emph{Adapt by Conditioning} ($n{=}18$), \emph{Characterize a Limit} ($n{=}15$), and \emph{Design a Property-Targeting Pretext Objective} ($n{=}15$). This is the long-tail signature of an inductively derived taxonomy. Many innovations use a small number of deeply worked strategies; a smaller tail uses rare but still coherent operators.

\begin{figure}[!htbp]
\centering
\includegraphics[width=\textwidth]{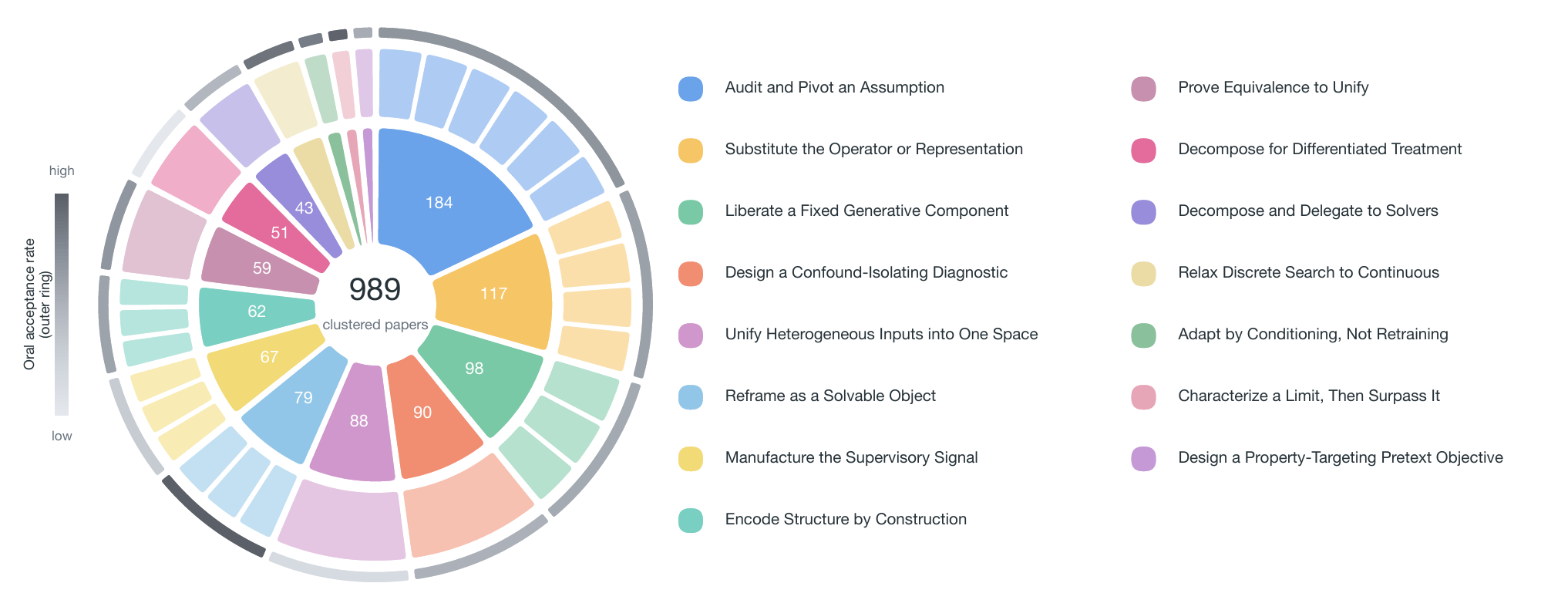}
\caption{\textbf{Ideation-pattern hierarchy across 989 clustered papers.} Three concentric rings share one angular layout, ordered by size clockwise from 12 o'clock; categorical colors identify the 15 patterns and are reused across all rings, with white gaps separating segments. \emph{Inner ring}: the 15 induced Level-1 ideation patterns, each wedge sized by its clustered-paper count (the count is printed inside the larger wedges; full pattern names are in the right-hand legend). \emph{Middle ring}: each pattern's arc is subdivided into its constituent sub-clusters (31 in total), shaded as a lighter tint of the parent pattern; the \emph{number} of segments within a wedge shows how finely that pattern fragments, while the segment \emph{angle} is the equal share within the pattern, not the per-sub-cluster paper count. \emph{Outer ring}: a thin heat band encoding each pattern's Oral acceptance rate $p_O = n_O / (n_O + n_{HC} + n_R)$ on the light-to-dark gray scale shown at left (light = low, dark = high), so reviewer outcome can be read against methodology at a glance. Totals: 15 patterns, 31 sub-clusters, 989 clustered papers (of 1{,}891 embedded).}
\label{fig:hierarchy_donut}
\end{figure}

Figure~\ref{fig:hierarchy_donut} visualizes the hierarchy with three concentric rings. The inner ring shows the 15 ideation patterns, sized by paper count. The middle ring subdivides each pattern into its constituent sub-clusters, 31 in total, tinted with the parent pattern's color. The outer band encodes each pattern's Oral acceptance rate on a light$\rightarrow$dark gray scale. Two empirical observations follow.

\textbf{(i) Methodology fragmentation is uneven across patterns, but the taxonomy is not over-split.} \emph{Audit and Pivot an Assumption} decomposes into 6 sub-clusters. \emph{Substitute the Operator or Representation} decomposes into 4. \emph{Reframe as a Solvable Object}, \emph{Manufacture the Supervisory Signal}, \emph{Encode Structure by Construction}, and \emph{Liberate a Fixed Generative Component} decompose into 3 each. The remaining 9 patterns are single-cluster patterns, which indicates tactically coherent moves that the abstraction does not split further. The 6-cluster decomposition under \emph{Audit and Pivot} reflects substantively different audit-pivot moves, such as identifiability-theorem pivots, distributional-assumption pivots, and sufficiency-bound pivots.

\textbf{(ii) Cluster sizes are tight and well-supported.} Cluster sizes range from 13 to 86 (mean 31.9, median 22). Of the 31 clusters, only 2 are small ($n<15$); the corresponding tactical cards are auto-flagged as \texttt{confidence: low}. The tight size distribution corresponds to a high silhouette ($0.584$).

\subsection{Multi-strategy assignments}
\label{subsec:multistrat}

This subsection reports the cluster-level view of multi-strategy composition. This view is deliberately limited: the cluster-level mapping caps each paper at \emph{primary~+~secondary}. The deeper paper-level multi-label view in \S\ref{subsec:multilabel} recovers the true modal composition, including the 902 HDBSCAN-unclustered papers excluded here. It should be read as the authoritative composition analysis.

\textbf{Paper-level coverage of the secondary slot.} The cluster-level \texttt{(primary, secondary)} assignment propagates to papers. Of the 989 clustered papers, $729$ ($73.7\%$) carry both a primary and a non-trivial secondary, while $260$ ($26.3\%$) carry only a primary. The $902$ HDBSCAN-unclustered papers are excluded from these cluster-level statistics and assigned later by multi-label tagging in \S\ref{subsec:multilabel}. Multi-strategy assignment is therefore the norm among coherently clustered papers.

Two cluster-level signals are worth stating because both fall outside the Oral-enriched paper-level ranking in \S\ref{subsec:multilabel} (Table~\ref{tab:multilabel_combos}). First, the clearest reject-enriched combination is \emph{architectural operator substitution} $+$ \emph{heterogeneous decomposition} ($-17.9$\,pp from the $58.4\%$ baseline), followed by \emph{algebraic equivalence unification} $+$ \emph{operator substitution} ($-8.4$\,pp). The likely failure mode is stacked unresolved tradeoffs: a more expressive operator is asked to absorb multi-population heterogeneity, so reviewers read the operator as the architecture rather than as a controlled inductive bias. Second, composition itself tracks acceptance. The $729$ paired papers reach $p_O = 58.2\%$ ($366/263$), essentially the $58.4\%$ baseline. The $260$ single-pattern papers reach only $52.7\%$ ($116/104$, $-5.7$\,pp). A single-primary cluster tends to be a lone tactical move, whose acceptance signal is weaker than a deliberately composed pair.

\subsection{Ideation-pattern adjacency in embedding space}
\label{subsec:adj}

\begin{figure}[!htbp]
\centering
\includegraphics[width=0.92\textwidth]{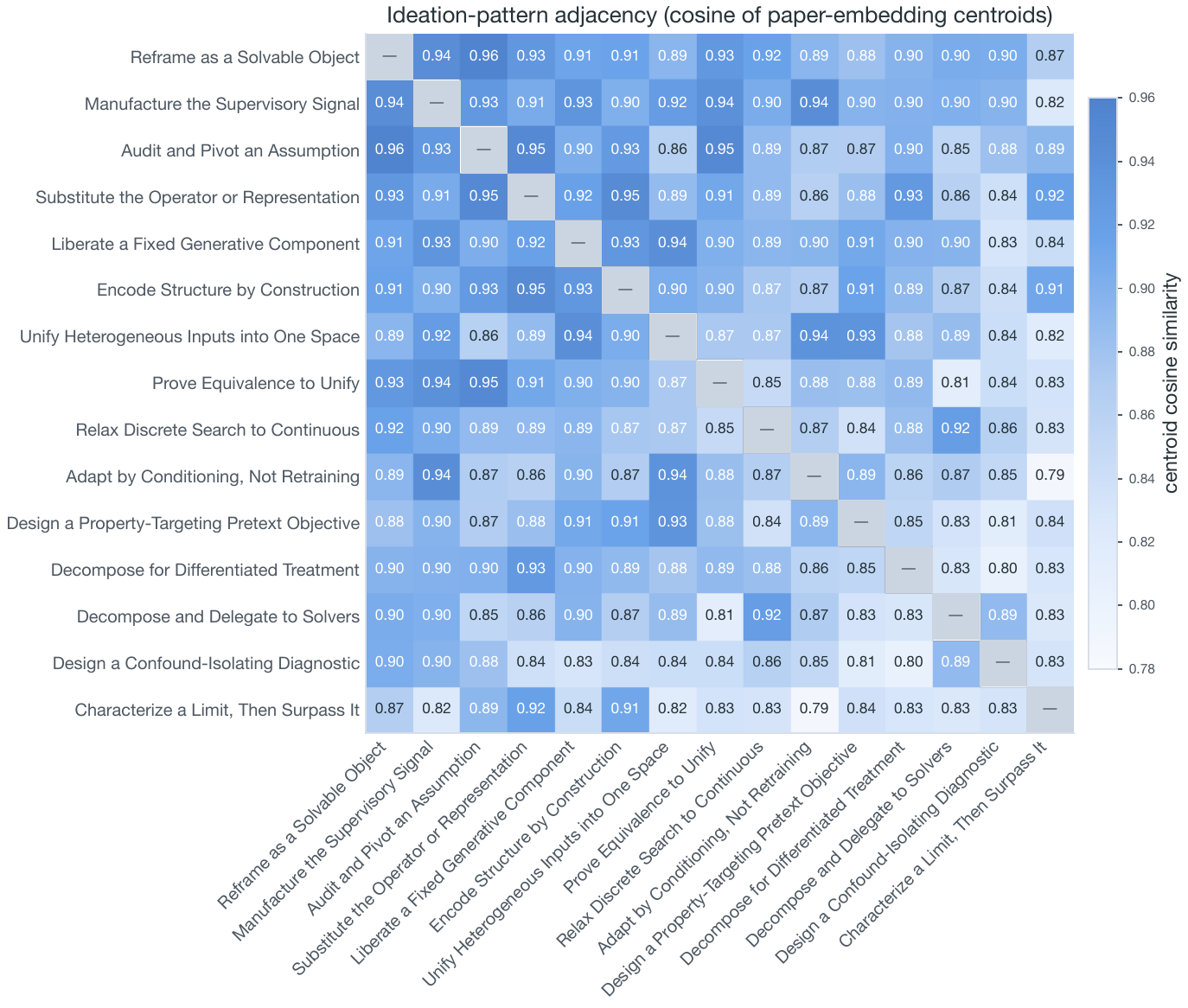}
\caption{Centroid-cosine similarity between the 15 ideation patterns. Rows/columns ordered by mean similarity (most ``central'' ideation patterns first). All off-diagonal entries fall in $[0.79, 0.96]$: ideation patterns are not orthogonal in this embedding; they form a dense neighborhood on a low-dimensional manifold.}
\label{fig:adjacency}
\end{figure}

Why do certain ideation pattern pairs combine often? The OpenAI \texttt{text-embedding-3-large} embedding gives a direct geometric measure. Each ideation pattern has a centroid, computed as the mean of its papers' normalized embeddings. Pairwise centroid cosine then quantifies how close two patterns are in strategy space. Figure~\ref{fig:adjacency} shows the full $15 \times 15$ matrix, and Table~\ref{tab:nn} gives each pattern's nearest neighbor.

\begin{table}[!htbp]
\centering
\caption{Nearest-neighbor ideation pattern and centroid cosine. Sorted by descending cosine. The two most isolated patterns are \emph{Design a Confound-Isolating Diagnostic} ($0.904$) and \emph{Characterize a Limit, Then Surpass It} ($0.917$); they do not share a neighborhood and form their own pockets in the embedding.}
\label{tab:nn}
\small
\begin{tabular}{p{6.0cm} p{6.0cm} r r}
\toprule
\textbf{Methodology} & \textbf{Nearest Neighbor} & \textbf{cosine} & \textbf{n} \\
\midrule
Reframe as a Solvable Object & Audit and Pivot an Assumption & 0.956 & 79 \\
Audit and Pivot an Assumption & Reframe as a Solvable Object & 0.956 & 181 \\
Substitute the Operator or Representation & Audit and Pivot an Assumption & 0.949 & 109 \\
Prove Equivalence to Unify & Audit and Pivot an Assumption & 0.949 & 59 \\
Encode Structure by Construction & Substitute the Operator or Representation & 0.946 & 61 \\
Manufacture the Supervisory Signal & Reframe as a Solvable Object & 0.943 & 66 \\
Adapt by Conditioning, Not Retraining & Manufacture the Supervisory Signal & 0.942 & 18 \\
Unify Heterogeneous Inputs into One Space & Liberate a Fixed Generative Component & 0.942 & 82 \\
Liberate a Fixed Generative Component & Unify Heterogeneous Inputs into One Space & 0.942 & 94 \\
Design a Property-Targeting Pretext Objective & Unify Heterogeneous Inputs into One Space & 0.935 & 15 \\
Decompose for Differentiated Treatment & Substitute the Operator or Representation & 0.927 & 47 \\
Decompose and Delegate to Solvers & Relax Discrete Search to Continuous & 0.923 & 42 \\
Relax Discrete Search to Continuous & Decompose and Delegate to Solvers & 0.923 & 35 \\
Characterize a Limit, Then Surpass It & Substitute the Operator or Representation & 0.917 & 15 \\
Design a Confound-Isolating Diagnostic & Reframe as a Solvable Object & 0.904 & 86 \\
\bottomrule
\end{tabular}
\end{table}

Three observations:

\textbf{(i) Adjacency predicts the Oral-enriched combinations.} The top three cluster-level Oral-enriched pairs are all high-cosine neighbors: \emph{Audit and Pivot} $+$ \emph{Structural Prior Encoding} ($+6.1$\,pp), \emph{Operator Substitution} $+$ \emph{Audit and Pivot} ($+5.5$\,pp), and \emph{Audit and Pivot} $+$ \emph{Controlled Diagnostic Design} ($+3.7$\,pp). \emph{Audit and Pivot an Assumption} sits at cosine $0.949$ from \emph{Operator Substitution}, and connects to \emph{Structural Prior Encoding} at $0.946$. Geometric proximity is therefore useful evidence for which secondary pattern is plausible to compose with a given primary.

\textbf{(ii) The two most isolated patterns are not simply the smallest.} \emph{Design a Confound-Isolating Diagnostic} ($n{=}86$) has the lowest nearest-neighbor cosine ($0.904$). \emph{Characterize a Limit, Then Surpass It} ($n{=}15$) has the second-lowest ($0.917$). Isolation therefore does not track support alone. The \emph{controlled diagnostic} pattern is geometrically distinct because its operational signature is to build a confound-isolating measurement instrument, which does not compose with the other 14 patterns in a standard way. This makes it a high-novelty leg candidate.

\textbf{(iii) Three patterns share hub status.} \emph{Audit and Pivot an Assumption}, \emph{Reframe as a Solvable Object}, and \emph{Substitute the Operator or Representation} each appear as the nearest neighbor for 3 other patterns. The hub set is distributed across diagnosis, framing, and substitution moves. Phase~2.1 therefore has three balanced default secondaries to consider when the bottleneck does not already imply a specific composition.

\subsection{Paper-level multi-label ideation pattern assignment}
\label{subsec:multilabel}

The cluster-level mapping in \S\ref{sec:methodology} caps each cluster, and therefore each clustered paper, at \emph{primary~+~secondary}. To test whether 2 is the right cap, and to surface 3+ pattern combinations directly, we run a separate Sonnet~4.6 pass on every paper independently. Given the 15 ideation-pattern definitions, the paper title, the abstract, and the four \texttt{abstract}$_*$ fields, the model emits a JSON list of patterns the paper actually \emph{executes}, not merely mentions. Each label includes a quoted evidence span and a \texttt{primary}/\texttt{supporting} role. There is no upper cap on \(k = |\text{applied\_patterns}|\); the minimum is~1.

All 1{,}891 papers with the four abstracted fields received a non-empty multi-label assignment. Rather than report a Bayesian acceptance rate (which requires constructing a denominator that mixes the OpenReview Oral/Reject decision with the citation-based HC class), we report the \emph{class-conditional distribution} \(P(k \mid \text{decision})\) directly. Table~\ref{tab:k_dist} shows, within each acceptance class, what fraction of that class's papers execute \(k\) ideation patterns.

\begin{table}[!htbp]
\centering
\caption{Class-conditional \(k\) distribution. Each column sums to 100\% within its class. \(\bar{k}\) is the per-class mean.}
\label{tab:k_dist}
\small
\begin{tabular}{c r r r r}
\toprule
\(k\) & all (n=1891) & Oral & HC & Reject \\
\midrule
1 &  137 (7.2\%)  &  ~~  &  ~~  &  ~~  \\
2 & \textbf{1119 (59.2\%)} & \textbf{mode} & \textbf{mode} & \textbf{mode} \\
3 &  611 (32.3\%) &  ~~  &  ~~  &  ~~  \\
4 &   24 (1.3\%)  &  ~~  &  ~~  &  ~~  \\
\midrule
\(\bar{k}\) & 2.28 & 2.28 & \textbf{2.32} & 2.26 \\
\bottomrule
\end{tabular}
\end{table}

Three observations:

\textbf{(i) \(k{=}2\) is the empirical mode in every class.} Overall, 59.2\% of papers execute exactly two ideation patterns. The mode is shared across Oral, HC, and Reject. Only $7.2\%$ of papers execute a single pattern, while a $33.6\%$ tail executes three or more. The cluster-based view in \S\ref{subsec:multistrat}, capped at \(k \leq 2\) by construction, therefore under-counts the 3-way tail. Paper-level multi-label tagging is the right granularity for capturing that tail.

\textbf{(ii) Class means cluster tightly around \(\bar{k}=2.3\).} The class means $\bar{k}_R = 2.26 \leq \bar{k}_O = 2.28 \leq \bar{k}_{HC} = 2.32$ are statistically indistinguishable at this sample size. Composition \emph{size} does not separate acceptance classes; composition \emph{content} (which patterns) does (\S\ref{sec:analysis}).

\textbf{(iii) Cluster-pair assignment under-represents true ideation-pattern coverage.} Cross-tabulating against the cluster-derived \texttt{(primary, secondary)} pair shows that multi-label tagging picks up signal beyond the dominant cluster. Most papers have at least one supporting pattern outside their cluster pair. The cluster pair captures the dominant axis, but not the routine 1--2 supporting moves that multi-label tagging surfaces.

\begin{figure}[!htbp]
\centering
\includegraphics[width=\textwidth]{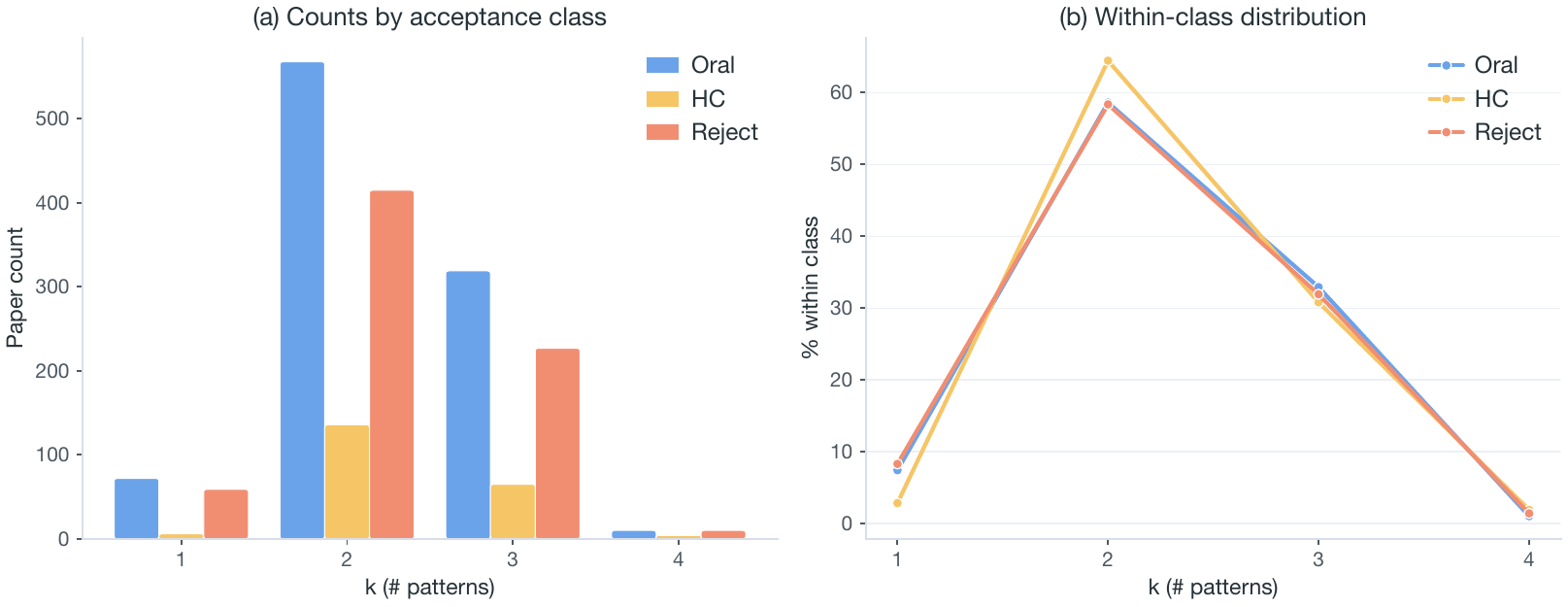}
\caption{Multi-label ideation pattern assignment. (a)~\(k\) distribution as grouped counts by acceptance class. (b)~Class-conditional distribution \(P(k \mid \text{decision})\): each curve is a probability mass function summing to 100\%. \(k{=}2\) is the mode for all three classes.}
\label{fig:multilabel_k}
\end{figure}

\paragraph{Top combinations under multi-label tagging.}
With paper-level multi-label assignments, we can rank 2-way and 3-way combinations directly (Table~\ref{tab:multilabel_combos}). We use the same definition of $p_O = n_O / (n_O + n_R)$ as in the cluster-level table, so the cross-table comparison is clean; the dataset baseline is $p_O = 58.4\%$. Three combinations stand out: \emph{Algebraic Equivalence Unification} + \emph{Generative Process Redesign} ($n_O + n_R = 20$, $p_O = 85.0\%$, $+26.6$\,pp), \emph{Algebraic Equivalence Unification} + \emph{Heterogeneous Decomposition} ($n_O + n_R = 22$, $p_O = 81.8\%$, $+23.4$\,pp), and \emph{Audit and Pivot} + \emph{Characterize a Limit, Then Surpass It} ($n_O + n_R = 60$, $p_O = 80.0\%$, $+21.6$\,pp). The third is the highest-volume strongly Oral-enriched combination in the dataset. Its empirical signature is: identify the load-bearing assumption, then prove a strict improvement over the limit it imposes. The Oral-enriched side of the 2-way table is consistent: pair a diagnostic or unification move with either a constructive substrate or a sharp analytical commitment.

\begin{table}[!htbp]
\centering
\caption{Top ideation pattern combinations under multi-label tagging, ranked by $p_O = n_O/(n_O + n_R)$ for direct comparability with the cluster-level combinations of \S\ref{subsec:multistrat}. 2-way threshold $n_O + n_R \geq 20$; 3-way threshold $n_O + n_R \geq 10$. Baseline $p_O = 58.4\%$ (1014 Oral, 722 Reject).}
\label{tab:multilabel_combos}
\small
\begin{tabular}{p{8.0cm} r r r r r}
\toprule
\textbf{Combination} & \(n_O\) & \(n_{HC}\) & \(n_R\) & \(p_O\) & \(\Delta p_O\) \\
\midrule
\multicolumn{6}{l}{\textit{Top 2-way ($n_O + n_R \geq 20$)}} \\
\midrule
\textbf{\textit{algebraic equivalence unification} + \textit{generative process redesign}}    & 17 & 0 &  3 & \textbf{85.0\%} & $+26.6$ \\
\textbf{\textit{algebraic equivalence unification} + \textit{heterogeneous decomposition}}    & 18 & 2 &  4 & \textbf{81.8\%} & $+23.4$ \\
\textbf{\textit{assumption audit \& pivot} + \textit{characterize limit then surpass}}        & \textbf{48} & 2 & 12 & \textbf{80.0\%} & $+21.6$ \\
\textit{generative process redesign} + \textit{reframe as solvable object}                    & 22 & 1 &  7 & 75.9\% & $+17.5$ \\
\textit{algebraic equivalence unification} + \textit{structural prior encoding}                & 15 & 0 &  7 & 68.2\% & $+9.8$  \\
\textit{generative process redesign} + \textit{self-supervised signal engineering}             & 19 & 7 &  9 & 67.9\% & $+9.5$  \\
\textit{decompose and delegate} + \textit{reframe as solvable object}                          & 19 & 1 &  9 & 67.9\% & $+9.5$  \\
\textit{controlled diagnostic design} + \textit{heterogeneous decomposition}                    & 20 & 8 & 10 & 66.7\% & $+8.3$  \\
\midrule
\multicolumn{6}{l}{\textit{Top 3-way ($n_O + n_R \geq 10$)}} \\
\midrule
\textbf{\textit{audit \& pivot} + \textit{decompose \& delegate} + \textit{reframe as solvable object}} & 8 & 0 & 2 & \textbf{80.0\%} & $+21.6$ \\
\textit{audit \& pivot} + \textit{characterize limit then surpass} + \textit{reframe as solvable object} & 11 & 0 & 3 & 78.6\% & $+20.2$ \\
\textit{audit \& pivot} + \textit{controlled diagnostic design} + \textit{heterogeneous decomposition}    & 8 & 2 & 3 & 72.7\% & $+14.3$ \\
\textit{audit \& pivot} + \textit{generative process redesign} + \textit{heterogeneous decomposition}    & 8 & 3 & 5 & 61.5\% & $+3.1$  \\
\textit{audit \& pivot} + \textit{reframe as solvable object} + \textit{structural prior encoding}        & 12 & 0 & 9 & 57.1\% & $-1.3$  \\
\bottomrule
\end{tabular}
\end{table}

\begin{figure}[!htbp]
\centering
\includegraphics[width=\textwidth]{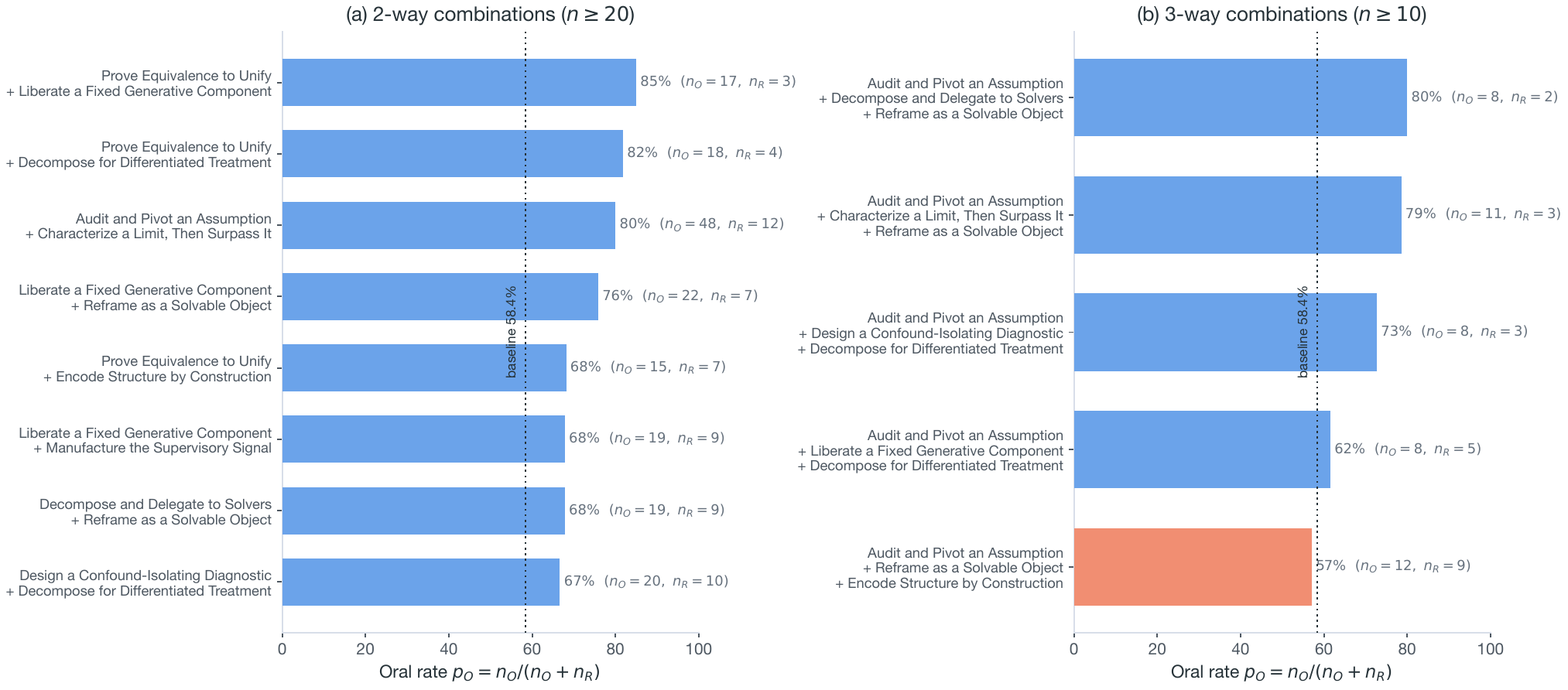}
\caption{Top combinations under multi-label tagging. (a) 2-way combinations ($n \geq 20$); (b) 3-way combinations ($n \geq 10$). Dotted line: dataset-wide baseline $p_O = 58.4\%$.}
\label{fig:multilabel_combos}
\end{figure}

\paragraph{What multi-label tagging implies for the skill.} The skill treats ideation-pattern composition as 1--3 entries per generated idea, with \(k = 2\) as the empirically supported default. For each composed pattern, it draws success and failure conditions from the ideation-pattern cards in \S\ref{subsec:cards}. The combination statistics in Table~\ref{tab:multilabel_combos} are surfaced only as context through the candidate's \texttt{innovation\_pattern\_landscape} block. That block shows the area's pattern usage alongside the candidate's chosen composition, so the user can see where the idea sits in the empirical distribution.

Pattern selection itself happens upstream by structural fit to the bottleneck in Phase~2.1, not by composition-rate priors. Loading priors at generation time was tested in an earlier design and removed because it induced distribution-bias homogenization across runs. The Oral-enriched combinations in Table~\ref{tab:multilabel_combos} are therefore empirical observations, not skill target templates.

\section{Acceptance and Impact Analysis}
\label{sec:analysis}

We define acceptance bias as $\Delta_{\text{OR}}(m) = p_\text{oral}(m) - p_\text{reject}(m)$. Here, $p_\text{oral}(m)$ is the fraction of \emph{cluster-level Oral primaries} whose primary pattern is $m$, and $p_\text{reject}(m)$ is the analogous fraction for Reject papers. Positive $\Delta_{\text{OR}}$ means the pattern appears more often in accepted work than in rejected work, relative to class totals. Negative $\Delta_{\text{OR}}$ means the opposite.

\begin{figure}[!htbp]
\centering
\includegraphics[width=0.95\textwidth]{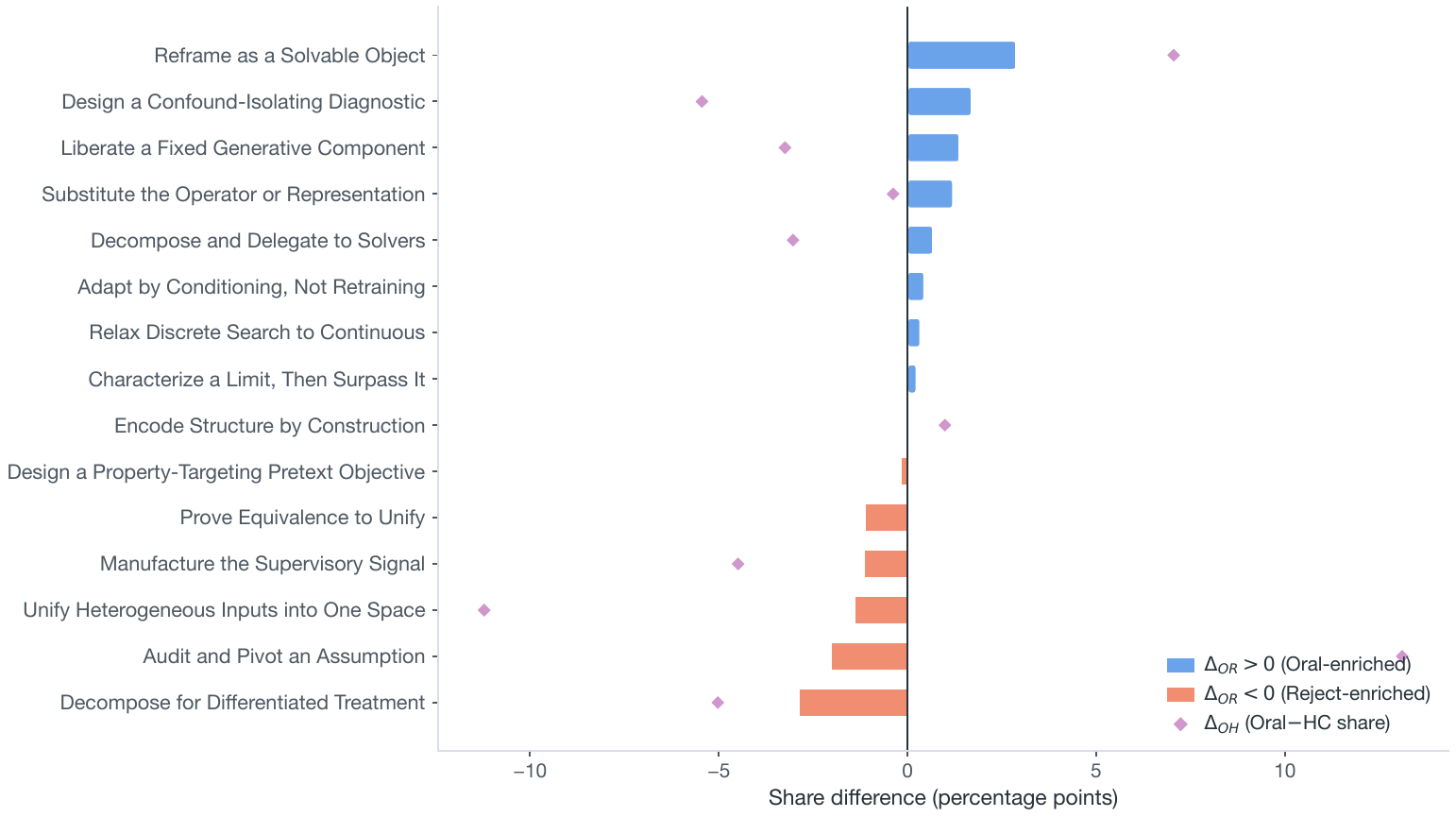}
\caption{Per-ideation-pattern acceptance bias. Bars show \(\Delta_{\mathrm{OR}}\); positive bars are Oral-enriched and negative bars are Reject-enriched. Diamonds show \(\Delta_{\mathrm{OH}}\) where \(n_{HC}\geq 5\).}
\label{fig:bias}
\end{figure}

This analysis is cluster-level and therefore partial. A cluster-level primary is available for $482$ of $1{,}014$ Oral papers and $367$ of $722$ Reject papers. The denominators of $p_X(m)$ are the cluster-level totals ($N_O = 482$, $N_{HC} = 172$, $N_R = 367$), not the full class populations. The remaining HDBSCAN-unclustered papers are covered by the multi-label tagging pass in \S\ref{subsec:multilabel}. Figure~\ref{fig:bias} plots the bias for all 15 patterns; Table~\ref{tab:bias} reports all values.

\begin{table}[!htbp]
\centering
\caption{Per-ideation pattern acceptance bias (cluster-level). $p_X(m)$ is the fraction of the cluster-level-primary papers in class $X$ whose primary is pattern $m$. Denominators are the cluster-level-primary totals: $N_O = 482$ Oral, $N_{HC} = 172$ HC, $N_R = 367$ Reject (these are the subsets of the full $1{,}014 / 260 / 722$ class populations that received a cluster-level primary; the remaining HDBSCAN-unclustered papers are covered by the multi-label tagging of \S\ref{subsec:multilabel}). $\Delta_{\text{OR}}{=}p_\text{oral}-p_\text{reject}$ in percentage points. $\Delta_{\text{OH}}{=}p_\text{oral}-p_\text{hc}$ (positive = PC-favored over community); reported only when $n_\text{HC} \geq 5$.}
\label{tab:bias}
\small
\begin{tabular}{p{6.6cm} rrrrr}
\toprule
\textbf{Ideation pattern} & \(p_\text{oral}\) & \(p_\text{hc}\) & \(p_\text{reject}\) & \(\Delta_{\text{OR}}\) & \(\Delta_{\text{OH}}\) \\
\midrule
Reframe as a Solvable Object & 10.0 & 2.9 & 7.1 & \textbf{+2.9} & +7.1 \\
Design a Confound-Isolating Diagnostic & 8.5 & 14.0 & 6.8 & +1.7 & $-$5.4 \\
Liberate a Fixed Generative Component & 9.5 & 12.8 & 8.2 & +1.4 & $-$3.2 \\
Substitute the Operator or Representation & 11.8 & 12.2 & 10.6 & +1.2 & $-$0.4 \\
Decompose and Delegate to Solvers & 3.9 & 7.0 & 3.3 & +0.7 & $-$3.0 \\
Adapt by Conditioning, Not Retraining & 2.1 & 1.2 & 1.6 & +0.4 & n/a \\
Relax Discrete Search to Continuous & 4.1 & 0.6 & 3.8 & +0.3 & n/a \\
Characterize a Limit, Then Surpass It & 1.9 & 0.0 & 1.6 & +0.2 & n/a \\
Encode Structure by Construction & 6.2 & 5.2 & 6.3 & 0.0 & +1.0 \\
Design a Property-Targeting Pretext Objective & 1.5 & 1.2 & 1.6 & $-$0.2 & n/a \\
Prove Equivalence to Unify & 6.2 & 1.2 & 7.4 & $-$1.1 & n/a \\
Manufacture the Supervisory Signal & 5.4 & 9.9 & 6.5 & $-$1.1 & $-$4.5 \\
Unify Heterogeneous Inputs into One Space & 6.2 & 17.4 & 7.6 & $-$1.4 & \textbf{$-$11.2} \\
Audit and Pivot an Assumption & 19.5 & 6.4 & 21.5 & \textbf{$-$2.0} & \textbf{+13.1} \\
Decompose for Differentiated Treatment & 3.1 & 8.1 & 6.0 & \textbf{$-$2.9} & $-$5.0 \\
\bottomrule
\end{tabular}
\end{table}

Two axes structure the table. The Oral-vs-Reject axis $\Delta_{\text{OR}}$ is tight, with a spread of only $\pm 2.9$\,pp. At the main-pattern level, strategy choice contributes little to acceptance. The orthogonal PC-vs-community axis $\Delta_{\text{OH}}$ is much wider, reaching roughly $\pm 13$\,pp, and is the stronger discriminating signal. Four observations follow.

\textbf{(O1) The $\Delta_{\text{OR}}$ axis is tight.} The range is $\pm 2.9$\,pp: \emph{Reframe as a Solvable Object} is at $+2.9$, while \emph{Decompose for Differentiated Treatment} is at $-2.9$. This narrow spread suggests that the main pattern alone explains little acceptance variance. Execution at the sub-pattern level, meaning the cluster rather than the Level-1 pattern, is what discriminates.

\textbf{(O2) \emph{Audit and Pivot} is high-volume, high-variance, slightly Reject-leaning, and heavily PC-favored.} It carries $19.5\%$ of Oral papers, the largest cluster-level Oral share, and $21.5\%$ of Reject papers, also the largest. This gives $\Delta_{\text{OR}}{=}{-}2.0$\,pp. Its $\Delta_{\text{OH}}{=}{+}13.1$\,pp is the largest positive PC bias in the taxonomy: PCs disproportionately Oral audit/pivot papers, while the community cites them less than expected. Execution quality, not strategy choice, separates this pattern's $94$ Oral from $79$ Reject papers.

\textbf{(O3) \emph{Unify Heterogeneous Inputs into One Space} is the dataset's clearest community-favored pattern.} Its $\Delta_{\text{OH}}{=}{-}11.2$\,pp is driven by a $17.4\%$ HC share against a $6.2\%$ Oral share, proportionally almost three times more HC than Oral. Cross-modal-alignment papers are reproduced and cited heavily, but PCs do not elevate them to Oral at the same rate. This is the inverse of \emph{Audit and Pivot}.

\textbf{(O4) \emph{Reframe as a Solvable Object} is the cleanest Oral signal.} Both $\Delta_{\text{OR}}{=}{+}2.9$\,pp and $\Delta_{\text{OH}}{=}{+}7.1$\,pp are positive. PCs reward the pattern, and the community does not over-cite it relative to Oral frequency. Combined with its high cosine to \emph{Audit and Pivot} ($0.956$) in the adjacency analysis, this supports a concrete Oral recipe: recast an intractable native formulation as a solvable object, then audit and pivot the assumption that the recast makes load-bearing.

\paragraph{The PC-vs-community axis ($\Delta_{\text{OH}}$).} The \(\Delta_{\text{OH}}{=}p_\text{oral}-p_\text{hc}\) column isolates a second axis: whether program committees and the citing community agree on what counts. Observations O2--O4 show that PCs tend to reward structural-insight moves, while the community tends to reward usable-infrastructure moves. The two rankings are nearly orthogonal. The audit layer therefore surfaces $\Delta_{\text{OH}}$ as a distinct signal, separate from the acceptance-prediction axis.

\paragraph{Use in the skill.} IdeaSpark does not use the acceptance or PC-vs-community statistics as selection priors at any phase. Pattern selection is driven by structural fit to the bottleneck. The statistics enter only as context for audit.

\section{Domain Distribution and Ideation-Pattern Breadth}
\label{sec:domains}

The ideation-pattern cards (\S\ref{subsec:cards}) and the per-pattern bias profile (\S\ref{sec:analysis}) treat the 1{,}947 papers as one population. ML, however, is not a monolithic field. A strategy that lands an Oral in \emph{learning theory} may not land in \emph{vision-language pretraining}. IdeaSpark therefore needs to distinguish broadly transferable patterns from domain-conditional ones. This section quantifies that distinction.

\subsection{Domain induction}
\label{subsec:domain-induction}

We do not preset a domain taxonomy. Instead, we induce one from the data using the same architecture that produced the 15 ideation patterns:

\begin{enumerate}[nosep, leftmargin=1.5em]
\item Sonnet~4.6 extracts 1--3 normalized research-subarea tags per paper from its title, abstract, and (when available, $77.9\%$ of the dataset) author-supplied OpenReview keywords.
\item The 3{,}909 unique free-form tags are embedded with OpenAI \texttt{text-embedding-3-large} and clustered with UMAP+HDBSCAN ($k{=}148$ tag-clusters at $\textit{mcs}{=}7$, unclustered $25.5\%$).
\item Sonnet~4.6 labels each tag-cluster.
\item Opus~4.7 induces a Level-1 domain taxonomy of $N \in [8, 30]$ entries and maps each tag-cluster to a domain label.
\end{enumerate}

The model returns $N{=}28$ domains. We make no edits.

The per-paper assignment is the union of domains over each paper's tags. Coverage is high: $\mathbf{98.5\%}$ of papers (1{,}918 / 1{,}947) receive at least one domain. The average paper receives $1.81$ domains. The 28 domains range from $n{=}256$ (\textit{Transfer, Continual \& Meta-Learning}) to $n{=}16$ (\textit{Online \& Bandit Learning}). The full ordered list and machine-readable taxonomy are in \texttt{clustering/domains/domain\_taxonomy.json}.

\subsection{Domain $\times$ ideation pattern heatmap}
\label{subsec:dm-heatmap}

\begin{figure}[!htbp]
\centering
\includegraphics[width=\textwidth]{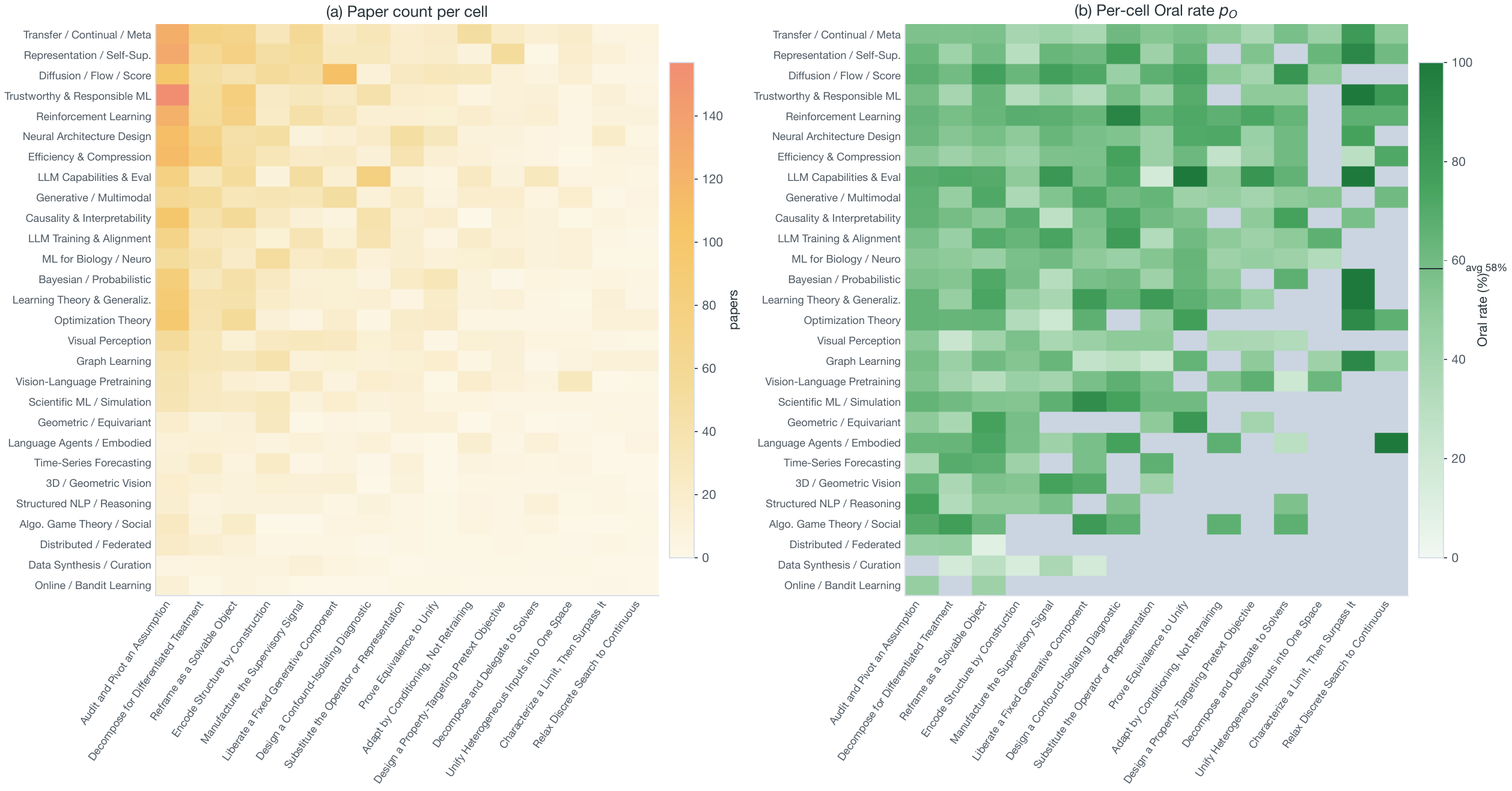}
\caption{(a) Paper counts per (domain, ideation pattern) cell, with domains and ideation patterns sorted by total. (b) Per-cell Oral rate $p_O$ (deeper green is higher; gray cells have $n_O + n_R < 5$, too few papers to estimate a rate; the colorbar marks the $58.4\%$ dataset average). The colored signal in (b) is heterogeneous: the same ideation pattern lands at very different Oral rates across domains.}
\label{fig:domain_method}
\end{figure}

We cross the 28 induced domains with the 15 ideation patterns using paper-level multi-label tagging (\S\ref{subsec:multilabel}). Each paper contributes one count to every (domain, pattern) cell it belongs to. The grid has $28 \times 15 = 420$ cells. Of these, $392$ ($93\%$) contain at least one paper, and $292$ ($70\%$) have $n_O + n_R \geq 5$. Figure~\ref{fig:domain_method} shows two views. Panel (a) reports raw paper count per cell. Panel (b) reports the cell-level Oral rate $p_O = n_O / (n_O + n_R)$, masked where $n_O + n_R < 5$.

The cell-level view adds domain specificity to \S\ref{sec:analysis}. The same ideation pattern can land at very different Oral rates across domains. One pattern is the main exception: \emph{Characterize a Limit, Then Surpass It} gives the cleanest cross-domain Oral signal in the corpus, occupying four of the five strongest cells. The five strongest cells, each with $p_O \geq 90\%$ on $n_O + n_R \geq 8$, are:

\begin{itemize}[leftmargin=1.4em, topsep=2pt, itemsep=2pt]
\item \textit{Trustworthy \& Responsible ML} $\times$ \textit{Characterize a Limit, Then Surpass It}: $\mathbf{100\%}$ Oral ($n_O{=}12$, $n_R{=}0$), the dataset's cleanest acceptance cell.
\item \textit{Learning Theory \& Generalization} $\times$ \textit{Characterize a Limit, Then Surpass It}: $100\%$ ($11/0$).
\item \textit{Reinforcement Learning} $\times$ \textit{Design a Confound-Isolating Diagnostic}: $93\%$ ($13/1$).
\item \textit{Representation \& Self-Supervised Learning} $\times$ \textit{Characterize a Limit, Then Surpass It}: $91\%$ ($10/1$).
\item \textit{Graph Learning} $\times$ \textit{Characterize a Limit, Then Surpass It}: $91\%$ ($10/1$).
\end{itemize}

The five weakest dense cells ($p_O \leq 25\%$) are:

\begin{itemize}[leftmargin=1.4em, topsep=2pt, itemsep=2pt]
\item \textit{Distributed \& Federated Learning} $\times$ \textit{Reframe as a Solvable Object}: $9\%$.
\item \textit{Graph Learning} $\times$ \textit{Substitute the Operator or Representation}: $21\%$.
\item \textit{Visual Perception \& Recognition} $\times$ \textit{Decompose for Differentiated Treatment}: $22\%$.
\item \textit{Model Efficiency \& Compression} $\times$ \textit{Adapt by Conditioning, Not Retraining}: $25\%$.
\item \textit{Graph Learning} $\times$ \textit{Liberate a Fixed Generative Component}: $25\%$.
\end{itemize}

The weakest dense cells show the other side of the same result. A pattern that wins in one domain can under-perform sharply in another. This supports the cards' guidance that success conditions do not transfer mechanically across domains.

The cross-domain consistency of \emph{Characterize a Limit, Then Surpass It} has a clear operational basis. When a paper identifies a formal limit, such as an information-theoretic, sample-complexity, expressivity, or generalization limit, and proves a strict improvement over it, the reviewer-defensible signature transfers across otherwise different domains.

\subsection{Ideation-pattern breadth: how domain-agnostic is each pattern?}
\label{subsec:breadth}

\begin{figure}[!htbp]
\centering
\includegraphics[width=0.95\textwidth]{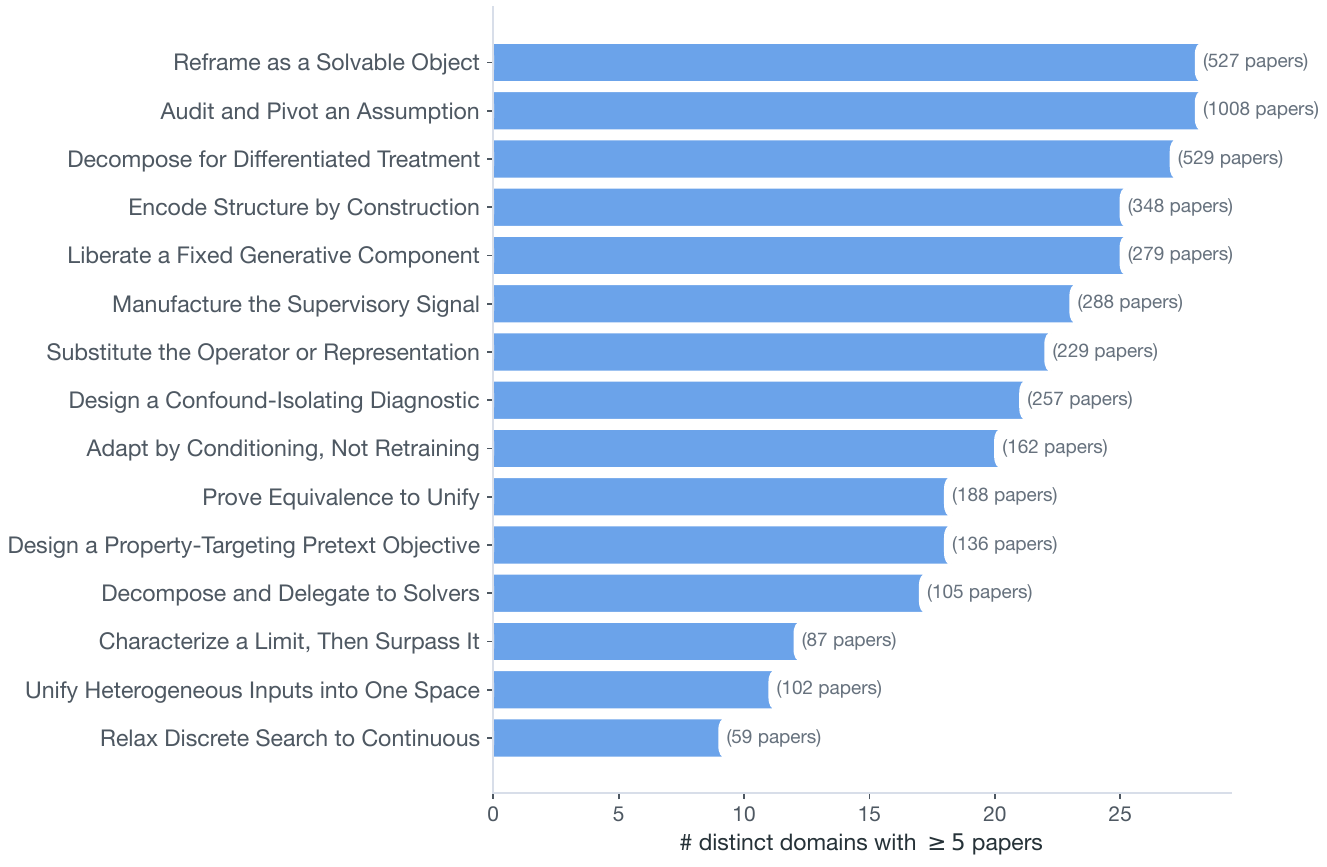}
\caption{Ideation-pattern breadth: number of distinct domains each ideation pattern has $\geq 5$ papers in (out of 28). The annotation gives each pattern's multi-label paper count: a paper is counted under every pattern it carries, so these counts overlap across patterns and sum to more than the $1{,}891$-paper corpus. Most patterns touch $\geq 18$ domains, supporting the skill's domain-agnostic positioning.}
\label{fig:breadth}
\end{figure}

For the skill to judge whether a pattern is plausible in a user's area, it needs each pattern's domain footprint. Figure~\ref{fig:breadth} plots, for each ideation pattern, the number of domains it touches and the share of its papers concentrated in its largest domain.

Three observations:

\textbf{(i) Diagnostic and framing patterns reach the entire field.} \emph{Audit and Pivot an Assumption} and \emph{Reframe as a Solvable Object} each have $\geq 5$ papers in \textbf{all 28 domains}. \emph{Decompose for Differentiated Treatment} reaches 27 domains. These three are also the highest-volume patterns in the multi-label distribution, with $1008$, $527$, and $529$ papers respectively under tagging. They also anchor the Oral-enriched multi-label combinations in Table~\ref{tab:multilabel_combos}. Breadth, volume, and acceptance signal therefore co-occur.

\textbf{(ii) Construction-level patterns are still broad, but more concentrated.} \emph{Encode Structure by Construction} reaches $25$ domains, \emph{Liberate a Fixed Generative Component} reaches $25$, \emph{Manufacture the Supervisory Signal} reaches $23$, and \emph{Substitute the Operator or Representation} reaches $22$. These patterns depend on a specific architectural choice or generative inductive bias. They appear across most of the field, but with clearer domain-modal anchors. Generative redesign concentrates in \textit{Diffusion, Flow \& Score Models}, and operator substitution concentrates in \textit{Sequence Modeling \& Architectures}.

\textbf{(iii) Domain-narrow patterns are not necessarily Reject-prone.} \emph{Relax Discrete Search to Continuous} touches $9$ domains, \emph{Unify Heterogeneous Inputs into One Space} touches $11$, and \emph{Characterize a Limit, Then Surpass It} touches $12$. The first two are narrow because their operational substrates exist only in some subfields: discrete-search-amenable problems and cross-modal alignment scaffolds. \emph{Characterize a Limit, Then Surpass It} is also narrow, but it is the cleanest cross-domain Oral signal we observe (\S\ref{subsec:dm-heatmap}). Its narrowness reflects a prerequisite. The field must already have produced a formal limit characterization, which is not true everywhere.

We do not report a per-domain temporal slice. Conditioning on (domain $\times$ year $\times$ pattern) leaves only 1--3 papers per cell for smaller domains, which is too thin for a quantitative claim. The skill also does not rely on such aggregate slices. Its Phase~2.1 saturation read is computed live from each run's Phase~0 retrieval (\S\ref{sec:trends}), so domain-specific saturation is read from the actual retrieved literature rather than from coarse dataset bins.

\section{Temporal and Conference Structure}
\label{sec:trends}

\subsection{Temporal trends}

\begin{figure}[!htbp]
\centering
\includegraphics[width=\textwidth]{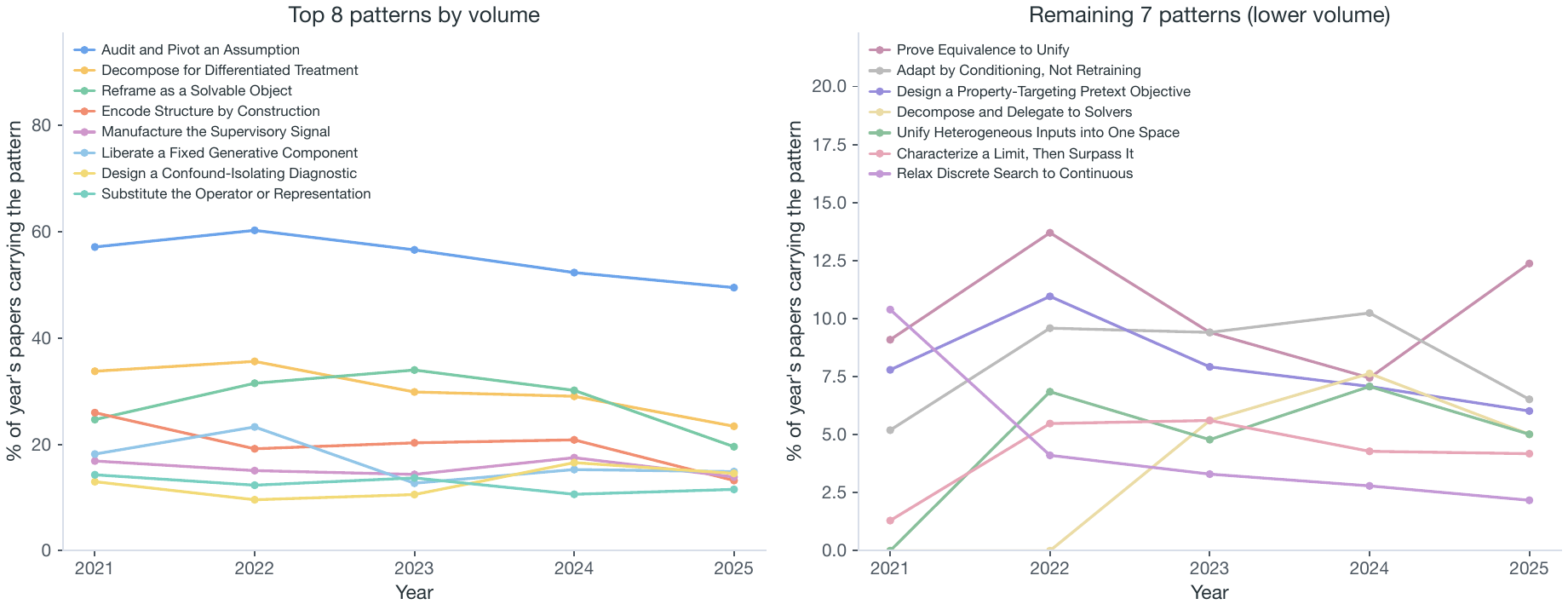}
\caption{Per-pattern share of papers by year under multi-label tagging (\% of that year's papers carrying the pattern in any role; rows sum to $\bar{k} \approx 230\%$). Left: top-8 by total volume. Right: remaining seven. Lines that rise through the window are the patterns currently ascending; those that fall are losing share.}
\label{fig:temporal}
\end{figure}

We summarize temporal change across the 2021--2025 window by comparing an early period, defined as the average over 2021--2022, with a late period, defined as the average over 2024--2025. The clearest movers are:

\begin{itemize}[nosep, leftmargin=1.5em]
\item \textbf{Risers.} \emph{Decompose and Delegate to Solvers} ($0.0\%$ early $\to$ $6.3\%$ late, $\Delta = +6.3$\,pp) is the dataset's fastest-growing pattern. It tracks the emergence of LLM-driven multi-agent and tool-orchestrated solver pipelines from 2023 onward. \emph{Design a Confound-Isolating Diagnostic} ($11.3\% \to 15.6\%$, $\Delta = +4.3$\,pp) reflects the maturation of controlled-perturbation evaluation as a first-class research move. \emph{Unify Heterogeneous Inputs into One Space} ($\Delta = +2.6$\,pp) is the third clearly rising pattern.
\item \textbf{Fallers.} \emph{Decompose for Differentiated Treatment} ($34.7\% \to 26.2\%$, $\Delta = -8.5$\,pp) loses the most ground. The ``split into sub-populations and treat differently'' move appears to have been absorbed into more specialized patterns, including operator substitution and structural priors. \emph{Audit and Pivot an Assumption} ($58.7\% \to 50.9\%$, $\Delta = -7.8$\,pp) remains the highest-volume pattern, but gradually gives back share. \emph{Liberate a Fixed Generative Component} ($\Delta = -5.7$\,pp) is the third faller, consistent with saturation of the generative-redesign wave that dominated 2021--2023.
\end{itemize}

\paragraph{Sample-size caveat for early years.} The corpus is unbalanced across years. The 2021 slice has 77 multi-labeled papers and 2022 has 73, while each year from 2023 to 2025 has $537$--$606$. The early-period share is therefore aggregated over a $\sim$150-paper base. Changes at the lower end of the $\Delta$ table ($|\Delta| < 2$\,pp) are not statistically robust, so we do not call them out. The risers and fallers above all carry $\Delta \geq 2.5$\,pp, corresponding to absolute paper-count moves of more than $\sim 15$ papers per direction.

\paragraph{How the skill uses these trend signals.} These corpus-wide temporal verdicts are descriptive context only. The skill never loads them as generation priors. Instead, Phase~1 reads pattern saturation live from each run's own retrieved literature (\S\ref{sec:skill}).

\subsection{Conference preferences}

\begin{figure}[!htbp]
\centering
\includegraphics[width=\textwidth]{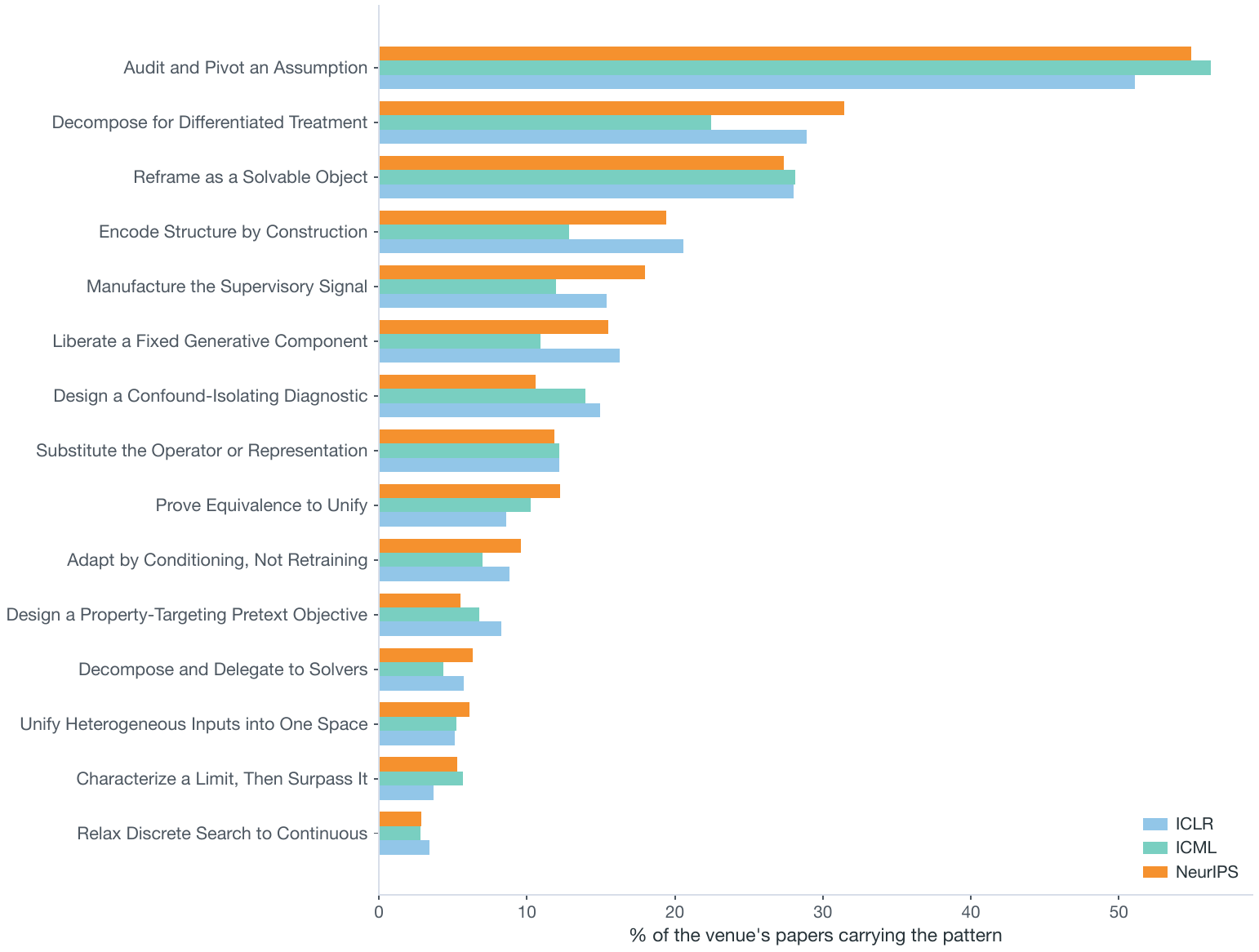}
\caption{Pattern usage by venue under multi-label tagging. Each bar reports the share of papers in a venue that carry the pattern in any role. Patterns are sorted by total volume.}
\label{fig:venue}
\end{figure}

Figure~\ref{fig:venue} reports each pattern's share within each venue, measured as the percentage of that venue's papers carrying the pattern under multi-label tagging. Three venue signatures emerge.

\textbf{ICML over-indexes on assumption-audit and limit-characterization.} \emph{Audit and Pivot an Assumption} reaches $56.2\%$ share at ICML versus $52.4\%$ elsewhere, a $+3.8$\,pp lift. \emph{Characterize a Limit, Then Surpass It} reaches $5.7\%$ versus $4.3\%$, a $+1.4$\,pp lift. Both patterns are diagnostic or theoretical, consistent with ICML's preference for sharply bounded analytical contributions.

\textbf{ICLR over-indexes on construction-side moves.} \emph{Encode Structure by Construction} reaches $20.6\%$ versus $16.2\%$ elsewhere ($+4.4$\,pp). \emph{Liberate a Fixed Generative Component} reaches $16.2\%$ versus $13.3\%$ ($+3.0$\,pp). ICLR also under-indexes on \emph{Prove Equivalence to Unify} ($-2.7$\,pp) and \emph{Audit and Pivot an Assumption} ($-4.5$\,pp). The venue signature favors architectural and representational construction over algebraic unification.

\textbf{NeurIPS over-indexes on heterogeneity-driven decomposition and supervisory-signal engineering.} \emph{Decompose for Differentiated Treatment} reaches $31.4\%$ versus $26.8\%$ elsewhere ($+4.7$\,pp). \emph{Manufacture the Supervisory Signal} reaches $18.0\%$ versus $14.3\%$ ($+3.7$\,pp). NeurIPS under-indexes on \emph{Design a Confound-Isolating Diagnostic} ($-4.0$\,pp). The venue signature is broadly systems-level composition plus invented supervision, consistent with its empirical bias toward large-experiment papers.

These venue signatures are quantified per pattern. The skill surfaces them only as audit context for the Phase~4 idea card, so the user can see the venue prior alongside the candidate's chosen pattern composition.

\section{Do Rejected Papers Inhabit a Different Strategy Space?}
\label{sec:reject}

A natural concern is that clustering Oral, HC, and Reject papers in one space may force Reject papers into clusters defined by accepted work. If rejected papers used \emph{different} strategies, the induced ideation patterns would under-represent them. We test this directly. The result is negative: rejected work clusters within the same 15-pattern taxonomy rather than in a strategy space of its own.

\subsection{Reject-only clustering and mapping}

We re-run the embedding+clustering pipeline on the 711 Reject papers that carry full \texttt{abstract\_*} fields. Of the 722 Reject papers, 11 lack one of the four fields. We keep the UMAP hyperparameters fixed ($n\_\text{components}{=}5$, $n\_\text{neighbors}{=}10$, cosine metric, seed 42). We also keep HDBSCAN fixed at $\min\_\text{cluster\_size}{=}10$ with \texttt{cluster\_selection\_method=leaf}. This configuration produces 13 Reject-only clusters, with an unclustered fraction of $40.5\%$ ($288 / 711$).

We then map each Reject-only cluster to its nearest main-pattern centroid in the original $3{,}072$-dimensional \texttt{text-embedding-3-large} space (\S\ref{sec:methodology}). All 13 clusters land within $0.897$--$0.986$ cosine of their assigned primary. In total, 11 of the 15 main patterns receive at least one Reject-only cluster. No Reject-only cluster requires an out-of-taxonomy label.

Four main patterns receive \emph{no} Reject-only mapping: \emph{Adapt by Conditioning, Not Retraining}, \emph{Characterize a Limit, Then Surpass It}, \emph{Design a Property-Targeting Pretext Objective}, and \emph{Encode Structure by Construction}. Three are small-sample patterns ($n \leq 18$ each), and one is a constructively encoded pattern. The empirical reading is that Reject papers do not form a separate strategic space. They map into the existing 15-pattern taxonomy at high cosine.

Table~\ref{tab:rejmap} compares the resulting Reject distribution against the main-taxonomy Reject distribution.

\begin{table}[!htbp]
\centering
\caption{Reject distribution. The \textbf{Reject-only} column counts papers in each main pattern via Reject-only-clustering $\to$ nearest-main-centroid mapping; the \textbf{Reject-in-full} column counts the same papers via their main-cluster primary inheritance. $\Delta$ is the divergence ($\geq 0$: Reject-only assigns more papers here than the main taxonomy does). $\sum$ Reject-only $= 423$ (rest unclustered); $\sum$ Reject-in-full $= 367$ (rest fall in non-mapped clusters). $\sum |\Delta| = 198$.}
\label{tab:rejmap}
\small
\begin{tabular}{p{7cm} rrr}
\toprule
\textbf{Ideation pattern} & \textbf{Reject-only} & \textbf{Reject-in-full} & \textbf{$\Delta$} \\
\midrule
Unify Heterogeneous Inputs into One Space     & 80 & 28 & $+52$ \\
Audit and Pivot an Assumption                 & 98 & 79 & $+19$ \\
Decompose and Delegate to Solvers             & 25 & 12 & $+13$ \\
Decompose for Differentiated Treatment        & 32 & 22 & $+10$ \\
Prove Equivalence to Unify                    & 36 & 27 & $+9$ \\
Substitute the Operator or Representation     & 47 & 39 & $+8$ \\
Design a Confound-Isolating Diagnostic        & 33 & 25 & $+8$ \\
Relax Discrete Search to Continuous           & 22 & 14 & $+8$ \\
Liberate a Fixed Generative Component         & 25 & 30 & $-5$ \\
Adapt by Conditioning, Not Retraining         &  0 &  6 & $-6$ \\
Characterize a Limit, Then Surpass It         &  0 &  6 & $-6$ \\
Design a Property-Targeting Pretext Objective &  0 &  6 & $-6$ \\
Manufacture the Supervisory Signal            & 14 & 24 & $-10$ \\
Reframe as a Solvable Object                  & 11 & 26 & $-15$ \\
Encode Structure by Construction              &  0 & 23 & $-23$ \\
\bottomrule
\end{tabular}
\end{table}

\subsection{Interpretation}

Ten of the 15 patterns yield comparable Reject counts across the two computations ($|\Delta| \leq 10$). The Reject pool's strategic shape is therefore stable whether we cluster it alongside accepted papers or separately.

The notable divergences are concentrated in two directions. Reject-only clustering assigns substantially more papers to \emph{Unify Heterogeneous Inputs into One Space} ($+52$) and \emph{Audit and Pivot an Assumption} ($+19$) than the in-full mapping does. Conversely, it assigns fewer papers to \emph{Encode Structure by Construction} ($-23$), \emph{Reframe as a Solvable Object} ($-15$), and \emph{Manufacture the Supervisory Signal} ($-10$).

The pattern is consistent: when Reject papers cluster among themselves, they move \emph{toward} the audit/unify lanes and \emph{away} from constructive lanes. The likely interpretation is that failed constructive moves can appear structurally like audit/unify moves once stripped of the construction that made them constructive.

\begin{figure}[!htbp]
\centering
\includegraphics[width=\textwidth]{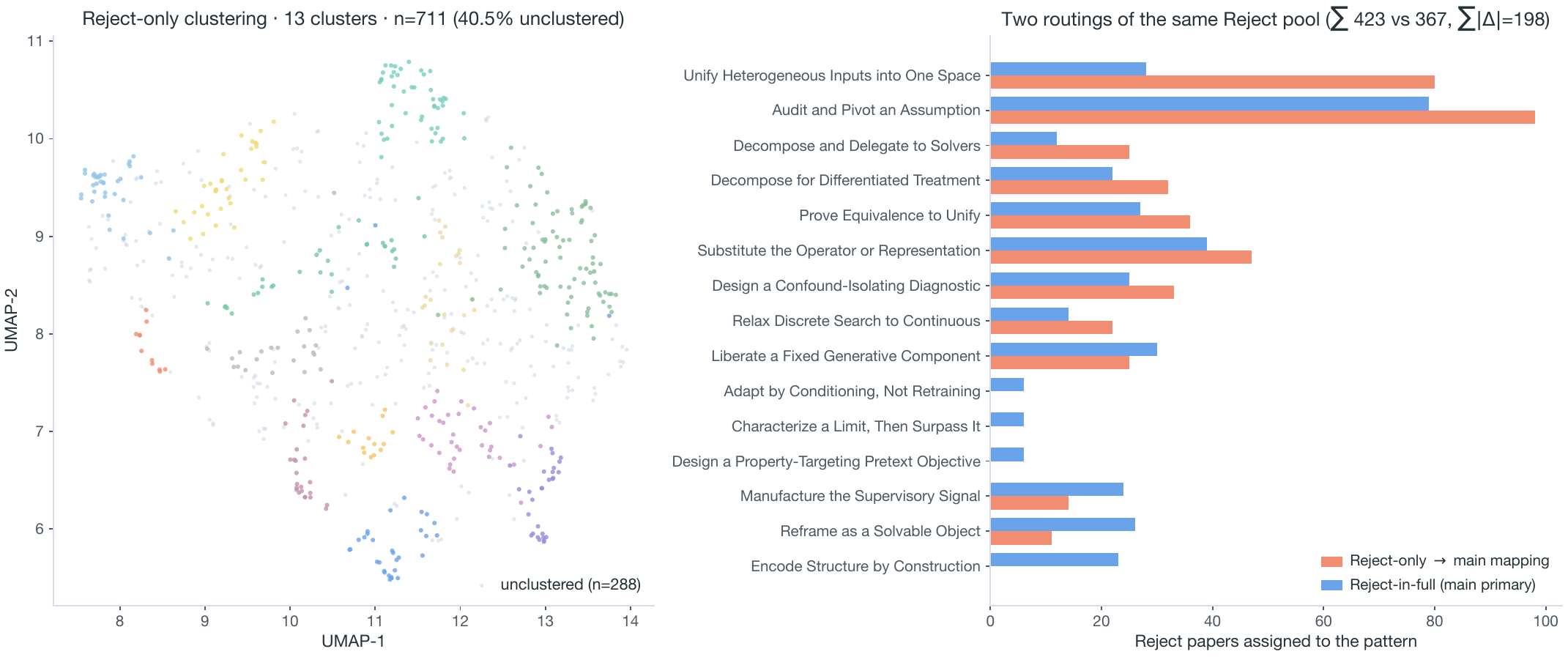}
\caption{Left: Reject-only clustering ($n{=}711$, 13 clusters, $40.5\%$ unclustered). Right: Reject-only-clustering $\to$ main mapping vs.\ main taxonomy primary distribution over the same Reject papers. The two columns largely agree; the divergences are concentrated on \emph{Unify Heterogeneous Inputs into One Space} and \emph{Audit and Pivot an Assumption} (over-attracted) vs.\ \emph{Encode Structure by Construction} (absent from the Reject-only clustering).}
\label{fig:reject_only}
\end{figure}

\textbf{Positive.} The 15-pattern taxonomy covers the Reject population without modification. Every Reject-only cluster maps into an existing pattern, with no out-of-taxonomy bucket. Rejection is therefore not explained by use of a different kind of strategy.

\textbf{Negative.} The Reject-only divergence ($\sum|\Delta| = 198$) shows that the fine-grained surface form of rejected work drifts toward more general audit/unify patterns and away from constructive patterns. This has a direct consequence for the skill. When Phase~2.1 selects a constructive primary pattern, such as \emph{Encode Structure by Construction}, \emph{Reframe as a Solvable Object}, or \emph{Manufacture the Supervisory Signal}, Phase~3.2 should check whether the candidate's executed surface form has collapsed toward the corresponding audit/unify lane.

\section{Ablation: Embedding Model and Abstraction Stage}
\label{sec:ablation}

\begin{figure}[!htbp]
\centering
\includegraphics[width=\textwidth]{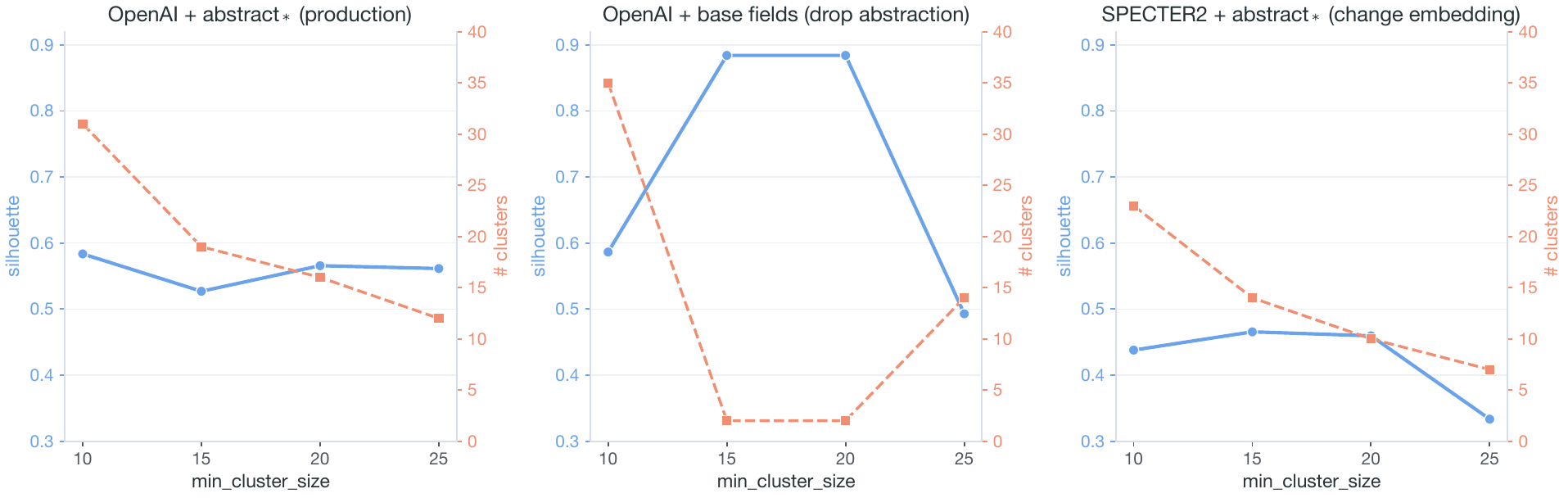}
\caption{Clustering quality across three embedding-input choices, as $\min\_\text{cluster\_size}$ varies. Solid line: silhouette. Dashed line: number of clusters $k$. Production (left) is stable. OpenAI + base fields (middle) collapses to $k=2$ degenerate mega-clusters at $\min\_\text{cluster\_size}=15$--$20$. SPECTER2 (right) stays at lower silhouette across the sweep.}
\label{fig:ablation}
\end{figure}

Two design choices are central to the induced taxonomy. First, the production pipeline uses a general-purpose paraphrase-strong encoder rather than SPECTER2. Second, it embeds the \emph{domain-agnostic} rewrites rather than the four original base fields. We ablate both choices on the same 1{,}891-paper subset used by the production pipeline. UMAP parameters are held fixed, and HDBSCAN is swept over $\min\_\text{cluster\_size} \in \{10, 15, 20, 25\}$. Figure~\ref{fig:ablation} and Table~\ref{tab:ablation} report the comparison.

\begin{table}[!htbp]
\centering
\caption{Clustering ablation. All three variants use UMAP (10-dim, $n\_\text{neighbors}{=}15$, $\min\_\text{dist}{=}0$, cosine) and an HDBSCAN sweep over $\min\_\text{cluster\_size}$. The two OpenAI rows differ only in which text is embedded. The production configuration is bolded.}
\label{tab:ablation}
\small
\begin{tabular}{l c c c c c}
\toprule
\textbf{Embedding} & \textbf{Text} & \textbf{mcs} & $k$ & \textbf{Unclustered \%} & \textbf{Silhouette} \\
\midrule
\textbf{OpenAI \texttt{text-embedding-3-large}} & \textbf{4 abstract fields} & \textbf{10} & \textbf{31} & \textbf{47.7} & \textbf{0.584} \\
OpenAI \texttt{text-embedding-3-large}          & 4 abstract fields          & 15 & 19 & 39.7 & 0.527 \\
OpenAI \texttt{text-embedding-3-large}          & 4 abstract fields          & 25 & 12 & 48.8 & 0.561 \\
\midrule
OpenAI \texttt{text-embedding-3-large}          & 4 base fields              & 10 & 35 & 35.7 & 0.586 \\
\textit{OpenAI + base (degenerate at mcs=15)}   & 4 base fields              & 15 & \textit{2} & \textit{0.0} & \textit{0.884} \\
\textit{OpenAI + base (degenerate at mcs=20)}   & 4 base fields              & 20 & \textit{2} & \textit{0.0} & \textit{0.884} \\
\midrule
SPECTER2 (768-dim)                              & 4 abstract fields          & 10 & 23 & 50.6 & 0.438 \\
SPECTER2 (768-dim)                              & 4 abstract fields          & 25 &  7 & 41.0 & 0.333 \\
\bottomrule
\end{tabular}
\end{table}

\subsection{Embedding model: OpenAI vs SPECTER2}

With identical input text and identical downstream processing, OpenAI \texttt{text-embedding-3-large} produces more clusters than SPECTER2 at $\min\_\text{cluster\_size}=10$ ($31$ vs.\ $23$). It also achieves a substantially higher silhouette ($0.584$ vs.\ $0.438$, a $+0.15$ gap). SPECTER2 degrades across the sweep, reaching $0.333$ at $\min\_\text{cluster\_size}=25$. Its unclustered fraction remains in the $41$--$54\%$ range.

The qualitative difference matches the training objectives. SPECTER2 is trained on citation-informed contrastive pairs and is strong at placing topically adjacent papers near each other. Our target, however, is strategy similarity rather than topic similarity. Two papers should cluster together if they both ``re-instantiate a proven mechanism at the appropriate domain-specific granularity'', even if they work on different modalities. The general-purpose paraphrase encoder better preserves this strategy-equivalence relation and produces a tighter strategy partition.

\subsection{Abstraction stage: silhouette is not enough}

The base-fields ablation gives a more subtle result. At $\min\_\text{cluster\_size}=10$, its silhouette is nearly identical to production ($0.586$ vs.\ $0.584$). Its unclustered fraction is also lower ($35.7\%$ vs.\ $47.7\%$). On the raw clustering metric, the base-fields variant therefore looks slightly better.

The same configuration is unstable across the sweep. At $\min\_\text{cluster\_size}=15$--$20$, it collapses to $k = 2$ mega-clusters with silhouette $0.884$ and $0\%$ unclustered. Inspection shows that these are not strategy clusters. One aggregates papers about engineering choices in deep learning. The other aggregates papers about formal or analytic ML. They are macro-topic clusters, not reasoning-move clusters.

This degenerate collapse indicates that topic cohesion has overpowered strategy cohesion. The base fields contain domain-specific vocabulary, including architecture names, dataset names, and field-specific mathematical objects. At coarser granularities, topic proximity then dominates the similarity structure. Two diffusion papers can cluster together because both are about diffusion, regardless of what they actually \emph{do} with diffusion.

The Stage-2 rewrite moves abstraction into the text before embedding. It removes topic anchors and leaves strategy-level signal. The production configuration remains well behaved across coarser settings: $k$ stays between 12 and 19, and silhouette stays in the $0.53$--$0.58$ range. It does not collapse into mega-clusters.

\paragraph{Ablation takeaways.}

We select the production configuration for construct validity and stability, not because it maximizes silhouette. The two ablations point in the same direction. The general-purpose paraphrase encoder beats SPECTER2 ($\Delta$ silhouette $= +0.15$) because SPECTER2's citation prior anchors on topic rather than on the strategy similarity needed here. The domain-agnostic rewrites tie the base fields on silhouette at the chosen setting, but avoid collapse at adjacent settings.

The base-field variant pays for cohesion in the wrong currency: it can achieve high silhouette by forming topic clusters. The Stage-2 rewrite removes those topic anchors and preserves robustness across the granularity sweep without giving up cluster quality. Production is therefore the only configuration whose clusters support the ideation-pattern taxonomy of \S\ref{sec:methodology} while remaining stable across the HDBSCAN sweep.

\section{Discussion}
\label{sec:discussion}

\subsection{Empirical takeaways}

The analyses above support four claims. A fifth claim is supported only at the idea-stage endpoint and remains open for human validation.

\textbf{(i) The strategy space is shared, not stratified.} Re-clustering the Reject pool in isolation (\S\ref{sec:reject}) reproduced the same 15-pattern vocabulary rather than a separate space. The signal separating accepted from rejected work therefore lies less in strategy choice than in execution. The per-pattern success conditions and failure modes carry that distinction. The one systematic divergence is that isolated Reject clusters drift toward audit/unify lanes and away from constructive lanes. This suggests that failed constructive moves can look like audit/unify moves once stripped of the construction that defines them.

\textbf{(ii) The Oral-vs-Reject axis is flat at the main-pattern level; PC-vs-community is the stronger discriminator.} Across granularities (\S\ref{sec:analysis}), ``which patterns'' contributes only a small main effect to acceptance. ``How well they compose'' contributes a larger one. This is the gap targeted by the skill's audit step. The nearly orthogonal $\Delta_{OH}$ axis, contrasting PC-favored and community-favored patterns, is wider and acts as a distinct second signal. The audit surfaces this signal independently of acceptance prediction. This is the softer of the two axes, however: HC papers come from a separately curated high-citation set rather than from OpenReview decisions, so we report $\Delta_{OH}$ only at $n_{HC} \geq 5$ and discount it elsewhere, and reconstructing HC from decision-aligned citation percentiles would let it carry the same weight as Oral-vs-Reject.

\textbf{(iii) Methodologies are broad in coverage but concentrated in effect.} Most patterns reach the majority of the 28 induced domains (\S\ref{subsec:breadth}). Yet the same pattern lands at very different Oral rates across domains (Fig.~\ref{fig:domain_method}b). The cleanest cross-domain Oral signal is \emph{Characterize a Limit, Then Surpass It}. For the skill, this means domain-cell rates should be surfaced as audit context, not loaded as generation priors.

\textbf{(iv) Composition is the norm, with $k{=}2$ as the empirical mode.} Paper-level multi-label tagging (\S\ref{subsec:multilabel}) shows that composition \emph{size} does not separate acceptance classes: $\bar{k}\approx 2.3$ in all three. Composition \emph{content} matters more. The skill therefore composes one to three patterns selected by bottleneck fit, defaulting to two, rather than ranking patterns by acceptance priors.

\textbf{(v) Partially closed: execution validity at the skill endpoint.} The construct-validity analysis in \S\ref{sec:ablation} shows that the cluster--pattern--card chain is internally coherent and stable across the HDBSCAN sweep. The reject-only analysis shows that every Reject-only cluster maps into the taxonomy. Section~\ref{sec:evaluation} adds an endpoint evaluation: when invoked at inference time and judged blind by two automated skills, IdeaSpark ideas receive the highest quality scores among four sources. What remains open is human-grounded validation against expert reviewers, implementation outcomes, and acceptance outcomes.

\paragraph{From taxonomy to system.} The taxonomy is the prerequisite for the skill system, but it is also a stand-alone deliverable. It has two immediate uses. A researcher can compare a candidate idea against the 15 ideation patterns and ask which success conditions the draft satisfies. A reviewer can also name a specific failure mode, for example ``combination indistinguishable from sum of parts.'' The skill system in \S\ref{sec:skill} operationalizes both uses inside an LLM scaffold.


\section{IdeaSpark Design}
\label{sec:skill}

We build \emph{IdeaSpark} to turn the empirical substrate in Sections~\ref{sec:methodology}--\ref{sec:reject} into a runnable research-ideation workflow. This section describes its runtime boundaries, evidence flow, phase contracts, audit logic, output surface, and mechanical validators. Endpoint idea quality is evaluated separately by automated judges in Section~\ref{sec:evaluation}. Human-rated quality and held-out rediscovery remain future work.

\subsection{Positioning and two-tier architecture}
\label{subsec:skill-positioning}

The skill converts an under-specified research direction into one reviewer-facing proposal. A successful run returns an artifact bundle centered on a compact idea card. The card contains the motivation, method flow, core claim, falsification prediction, and feasibility validation. One user input produces one of three outcomes: an idea-card bundle, a \texttt{do\_not\_generate} diagnosis when the direction cannot be grounded, or a \texttt{phase\_3\_failed} report when the audit abandons a candidate. The skill never asks the user mid-flow. Missing context is inferred when reasonable; unrecoverable ambiguity becomes an explicit stop condition.

The skill is not an ideation-pattern recommender, experiment planner, or paper writer. It does not propose datasets, baselines, ablation matrices, expected figures, or a calendar plan. Those choices depend on user context the skill cannot access. It produces an idea plus an honest feasibility judgment; empirical engineering remains the user's responsibility.

IdeaSpark is implemented as a two-tier skill. The runtime tier contains compact instructions, orchestration logic, retrieval hooks, phase prompts, rendering scripts, and deterministic validators. The evidence tier contains the 15 ideation-pattern cards, 31 sub-pattern cards, domain-by-pattern matrix, saturation records, and corpus-derived failure-mode inventory. Evidence is read progressively. Phase~2 reads only the pattern definitions and sub-pattern tactics needed to instantiate a candidate. Phase~3 reads failure modes, anti-pattern context, and collision evidence for audit. Phase~4 reads the audit findings to explain the final card. Raw counts and confidence flags remain analytical metadata for readers. They are not used as generation priors.

\subsection{Design principles}
\label{subsec:skill-principles}

The skill's core stance is that a publishable research idea is \emph{constructed}, not invented from scratch. The reasoning chain has three steps. First, read recent literature in the user's area and identify a structural gap no retrieved paper closes: the \emph{bottleneck}. Second, use the induced ideation-pattern vocabulary to ask what kind of research move could close that gap. Third, instantiate the chosen move as a concrete mechanism. The 15-pattern and 31-sub-pattern substrate supplies the vocabulary for the second step. It keeps generation at the level of pattern-applied-to-gap rather than unconstrained brainstorming.

\paragraph{Principle 1: Literature first, vocabulary second, idea third.} Bottleneck identification must be grounded in retrieved papers, not model recall. Pattern selection reads the shape of the unresolved gap and asks which operational move would close it. Idea instantiation then uses the selected sub-pattern as a tactical guide. It applies the abstract move to the specific gap rather than copying a paper-specific template.

\paragraph{Principle 2: Substantive contribution over methodological labels.} A candidate cannot defend itself as ``we use ideation pattern X while prior work used pattern Y.'' Method choice is process. The contribution is what the candidate \emph{derives, constructs, measures, or tightens} that prior work did not. This is enforced in the differentiation deltas and downstream audit checks. The 15 patterns are diagnostic vocabulary, never the contribution claim itself.

\paragraph{Principle 3: Generation freedom, audit anchoring.} Earlier iterations placed too many corpus-derived constraints inside generation. That biased candidates toward frequent corpus moves and away from the user's bottleneck. The current design keeps positive pattern vocabulary in Phase~2, but moves negative and frequency-based signals to audit and explanation. Phase~3 checks whether the candidate survives known failure modes. It does not reject a candidate merely because its pattern is common.

These choices follow from the empirical substrate. Rejected papers often use the same strategy operators as accepted papers, so the taxonomy serves as thinking and audit vocabulary rather than as a hard generator filter. Because $k=2$ is the modal paper-level composition, with a $33.6\%$ tail at $k \geq 3$, Phase~2 allows one to three gap-closure entries. Because some two-way combinations are reject-enriched unless the downstream artifact is substantively delivered, Phase~3 includes an anti-pattern substantive-verification check. Because acceptance is domain-conditional, saturation and domain-landscape information are surfaced as audit and explanation context rather than as recommendation priors.

\subsection{Workflow and phase contracts}
\label{subsec:skill-workflow}

The skill treats the 15 induced ideation patterns as diagnostic vocabulary, not as labels a candidate must classify into. The advance path combines model-reasoning phases with deterministic retrieval, rendering, and validation steps. The revise path adds one bounded repair step. That repair is constrained by audit targets and by byte-level preservation of the candidate's kill-switch commitments.

\subsubsection{Phase 0: Literature grounding}

Phase 0 builds the broad literature-grounding bundle. It is the only phase that retrieves \emph{area-level} context. The later Phase~3.1 retrieval is narrower: it checks for mechanism-level collision with the generated candidate.

The system turns the user direction into several search queries, checks which connectors are available, and queries four sources. Each source is used for the role it covers best. arXiv supplies recent preprints. OpenReview supplies forward-looking, in-review signal. OpenAlex supplies the broad published-paper graph. Semantic Scholar supplies CS-focused metadata and short summaries. arXiv and OpenReview cover the recent 0--6 month window. OpenAlex and Semantic Scholar cover the 6--24 month window.

This split is deliberate. No single source provides all four strengths: preprint recency, in-review signal, breadth, and clean cross-identifiers for deduplication. Each source is queried where it is strongest. The two time windows are kept non-overlapping. A paper that appears both as a recent preprint and as an in-review submission is counted once, avoiding inflated pattern-saturation counts in Phase~1.

The retrieved papers are then deduplicated across sources and lightly tagged with ideation patterns for later analysis. Immediately after the literature table is written, Phase~0+ fetches a small full-text cache for the most relevant references. Phase~1 hard-gates on this cache, so the pipeline cannot silently degrade to abstract-only bottleneck reasoning.

Two refinements harden the retrieval. \emph{For recall}, keyword matching is complemented by a meaning-based pass that finds conceptually close papers with little surface-word overlap. This pass is gated to the home field of the keyword results, so broader matching does not pull in off-field papers. Terminology drift would otherwise hide close competitors. \emph{For reading fidelity}, the system fetches open-access full text where available. Bottleneck and differentiation judgments can then rest on method sections, not abstracts alone.

The overall goal is that every downstream check (bottleneck, citation, collision) traces to a retrieved record rather than to model memory.

\subsubsection{Phase 1: Bottleneck identification}

Phase~1 first checks that Phase~0 produced a valid grounding bundle and that Phase~0+ produced the full-text cache. It then reads the user direction, retrieved records, abstracts, and cached full-text snippets. Its main output is one literature-grounded bottleneck: a concrete structural gap, not a topic label. It also names the closest adjacent papers, explains what each still leaves open, lists gaps Phase~0 did not close, and summarizes pattern usage across on-topic papers. These usage bands are descriptive, not temporal. A pattern is \emph{saturated} when it dominates the neighborhood, \emph{untested} when it appears at most once, and \emph{mid-frequency} otherwise. The summary is carried forward as audit and explanation context only.

To compare gaps on a common footing, Phase~1 arranges the neighborhood into a \emph{method-lineage tree}. Each retrieved method is a node linked to the earlier method it refines or replaces. The tree runs from foundational ancestors to current leaves. When the recent retrieval window misses an older ancestor, the model may add it as an \emph{awareness-only} node, explicitly marked non-citable. This division is deliberate. The window is recent because frontier grounding matters most there, and model memory is least reliable there. Older foundational methods are more stable. Because awareness-only nodes make no novelty claim, they serve as regression guards rather than evidence. An anti-fabrication rule keeps the tree shallow on genuinely young lines instead of padding it with invented ancestors.

The tree exposes three things a flat abstract list cannot. \emph{Additive} gaps sit at a leaf: an unmet need no current method achieves. \emph{Subtractive} gaps are read from a shared ancestor: the load-bearing assumption every leaf inherits. These gaps matter because strong papers often remove an assumed-necessary mechanism rather than add another component. Third, the tree blocks \emph{regression}. A proposed fix that an ancestor already implemented is treated as superseded, not novel. Throughout, the reading is evidence-gated. A subtractive opening may name only an assumption the retrieved methods actually show.

The route out of Phase~1 is binary: proceed, or stop with a diagnostic. Phase~1 never asks the user for missing information. It infers reasonable defaults, for example a single-researcher cloud compute budget when none is stated. It stops only when the direction is too broad, has no literature anchor, or the retrieved pool is too sparse to ground a bottleneck. The two structural routes cover an estimated 75--85\% of Oral types in the corpus (theory-plus-reframing, scaling-law, surgical-fix-plus-theory, evaluation-validity-audit, and most empirical-reveal Orals); benchmark-construction and pure system or infrastructure directions are not yet separate triggers and currently resolve through one of the two routes, with the subtype named only as diagnostic explanation. Adding dedicated triggers for them is a localized extension that would raise coverage without changing the architecture.

\subsubsection{Phase 2: Pattern-guided ideation}

Phase 2 separates pattern selection from candidate generation. This separation is important: the first step asks which corpus-derived research move can close the diagnosed gap, while the second step instantiates that move as a concrete candidate.

\emph{Pattern selection.} The system chooses one anchor gap from Phase~1 and keeps only the sibling gaps that can plausibly form a single paper with it. For each retained gap, it reads the 15 main ideation-pattern cards and selects the pattern whose operational signature structurally closes that gap. Pattern frequency and saturation are recorded as audit context, but they do not act as generation priors.

\emph{Candidate generation.} For each selected main pattern, the system chooses one sub-pattern and reads its tactical panels as a guide. The output is not just a title or theme. It contains the core mechanism, per-gap closure rationale, differentiation from the closest adjacent work, a compute budget, and a falsification prediction naming what should change if the mechanism is real.

Finally, the candidate passes through a deterministic sub-pattern citation gate. The gate verifies that every cited sub-pattern exists under the stated parent pattern before any expensive collision retrieval or audit is run.

\subsubsection{Phase 3: Quality gauntlet}

Phase 3 has three steps: a focused literature search, a corpus-anchored audit, and an optional revision.

First, the system re-queries the literature sources using terms drawn from the candidate's mechanism and claim. This search is separate from Phase~0 because it looks for \emph{recent mechanism overlap}, not broad area context.

Second, the audit reads the candidate, selected gaps, relevant sub-pattern cards, the Phase~0 evidence pool, the new collision hits, and the anti-pattern inventory. It runs four checks: gap-closure reject lessons, recipe application, anti-pattern substantive verification, and paper-pointed threat. It returns one of three verdicts, advance, revise, or abandon, plus concrete revision targets. The audit only judges; it does not edit the candidate.

Third, and only when revision is possible, a separate step applies the named edits or swaps in a same-parent sub-pattern. It does not re-judge the verdict, change the selected gaps, or touch the kill-switch commitments. Splitting judging from fixing keeps attack-finding and attack-resolution separate, and keeps every failure mode tied to corpus evidence.

\paragraph{Phase 3 audit details.} The audit is a single judgment step. It performs four corpus-anchored checks, applies a two-layer verdict, and emits revision targets when a bounded repair is possible. Each check requires concrete evidence from the retrieved literature, sub-pattern cards, or anti-pattern inventory. This makes the audit inspectable and resistant to rubber-stamping. The audit judges whether the candidate survives known failure modes. It is not a guarantee of novelty, acceptance, or endpoint idea quality.

\paragraph{Check 1: gap-closure-scoped reject scan.} \emph{What it catches}: candidates that look novel from the outside but fall into corpus-documented reject patterns specific to the chosen tactic. For each proposed gap closure, the audit reads the relevant sub-pattern's failure-mode panel and rejected-paper lessons. It evaluates whether the candidate clearly avoids the lesson, clearly matches it, or sits in a borderline state, and it must cite the candidate component that triggers the concern.

\paragraph{Check 2: recipe application.} \emph{What it catches}: candidates that cite a real sub-pattern but only execute its parent pattern's generic idea rather than the cluster's distinctive move, the leading cause of incremental output. Against the cited sub-pattern card's signature-move panel, the audit asks whether the core mechanism actually performs that move or merely gestures at the parent; an absent move marks the candidate \emph{bypassed}. This is the semantic backstop the deterministic citation gate (\S\ref{subsec:skill-workflow}) cannot provide: the gate proves only that the citation resolves to a real cluster under the right parent, never that the mechanism enacts the cluster's tactic.

\paragraph{Check 3: anti-pattern substantive verification.} \emph{What it catches}: keyword-stuffed mitigations, candidates whose core mechanism contains the right mitigation phrases without producing the artifact those phrases name. If the candidate uses one of the reject-enriched two-pattern compositions documented in Section~\ref{subsec:multistrat}, the audit verifies that the required mitigation is substantively delivered in the mechanism, rather than through wording alone in the candidate.

\paragraph{Check 4: paper-pointed threat.} \emph{What it catches}: candidates whose claim is already subsumed by a specific recent paper, whether retrieved by Phase 0's broad-domain queries or by Phase 3's mechanism-specific collision retrieval. If a paper that subsumes or competes with the candidate's claim is found, the audit emits the threat paper, a one-paragraph subsumption argument citing the threat's specific result, and a note on what would need to change. If no paper-pointed threat exists after honest search, the absence is a legitimate clearance signal; fabricating a generic threat to fill the audit is forbidden.

\paragraph{Why saturation is not gated here.} An earlier design added a fifth saturation-defense check. We removed it because it consulted no retrieved related work. It was an advisory prior, not a corpus-anchored fact, and therefore did not belong beside the four hard-floor checks above. Pattern saturation is still computed in Phase~1 and recorded in Phase~2. It now surfaces downstream as a Phase~4 reviewer concern rather than gating the audit. This keeps every Phase~3 check anchored on retrieved evidence.

\paragraph{Two-layer verdict.} The verdict has two layers. \emph{Layer 1} is a hard floor the model cannot override. It abandons the candidate when a reject lesson clearly matches, when a reject-enriched composition has no insertable mitigation, or when a recent paper has exact-mechanism overlap. \emph{Layer 2} is model judgment within the safe zone Layer~1 leaves. It chooses advance or revise based on whether borderline findings touch load-bearing structural properties. A recipe-application bypass always routes to revise: either swap to the sibling sub-pattern whose move the mechanism actually performs, or rework the mechanism to perform the cited move. If neither fits under the same parent, the run returns to Phase~2 regeneration rather than patching in place. The verdict must cite specific check findings; ``all checks pass'' without naming one is a process error.

\subsubsection{Phase 4: Expansion, implementability audit, rendering, and validation}

Phase~4 turns the candidate into the final idea-card artifact bundle. It expands the canonical candidate, runs a default-on implementability audit, renders side artifacts, and then runs deterministic validators. The expansion derives the motivation, method flow, feasibility, reviewer concerns, literature differentiation, and domain-landscape explanation from the candidate plus the Phase~3 audit report.

The implementability audit is a single bounded LLM call. For each method step, it states \emph{how} to build that step and flags anything left underspecified. It is written to a separate file. By construction, that file cannot carry the protected falsification and compute-budget commitments. The audit therefore adds buildability detail without changing the claim. If a run skips it, the implementability validators are not enabled.

Rendering is deterministic templating. The renderer builds the Markdown and LaTeX artifacts, and compiles a PDF when a local TeX engine is available. Because the templates are fixed, validators can compare protected commitments against the pre-render candidate and check that required sections are present before the user sees the card.

\paragraph{Phase 4 validation.} A suite of deterministic validators runs around the Phase~4 render path to check contracts asserted by earlier phases. They are mechanical. They detect violations through exact-text comparison, section presence, list length, schema coverage, readability checks, and citation consistency, not model judgment.

\paragraph{Kill-switch integrity (hard fail).} The candidate's falsification prediction and compute budget must be preserved from initial candidate generation through final rendering. The validation chain depends on the verdict path: direct expansion when the audit advances, or revised candidate followed by expansion when the audit requests a bounded fix. The Phase~3 revision step is forbidden from modifying these two commitments. The validator catches any drift mechanically rather than relying on prompt compliance.

\paragraph{Expansion completeness (hard fail).} Phase~4 expansion must contain the structural sections the idea-card rendering depends on: motivation with multiple prior-work stopping points, method-flow steps, feasibility validation with component-level and overall verdicts, non-empty proposal summary, core claim, and sub-claims. The check is section presence and list length. Absence is a hard fail because missing sections would silently render as blank content in the final artifacts.

\paragraph{Additional contract checks.} Three further validators harden the pipeline contracts. \texttt{subpattern\_citation\_consistency} runs immediately after Phase~2.2 and before any Phase~3 work. It checks that each cited sub-pattern resolves to a real cluster under the stated parent pattern. \texttt{implementability\_completeness} confirms that the audit covers every method step when Phase~4.1.5 ran. \texttt{implementability\_readability} flags known readability regressions in the standard-register buildability text.

\paragraph{What the validators do not check.} They do not enforce semantic correctness inside a section, they do not check whether the candidate's claim is actually novel, and they do not enforce cross-run determinism. The phase calls are LLM-reasoning steps with non-zero variance, so the same query can yield different but coherent candidates across runs; this is expected for an inference-time skill rather than a defect, and reporting distributions over repeated runs or seed-ensembled selection would make the variance explicit (\S\ref{sec:limitations}).

\subsection{Output surface}
\label{subsec:skill-schema}

The user-facing deliverable is deliberately slimmed to the three things a reader first judges: title, motivation, and method. An earlier, richer card printed the hook, claims, rebuttals, and provenance together. That version buried the idea under its own audit log. The current card is emitted in English and Chinese. Heavier material remains in the artifact bundle: the hook, problem-to-outcome summary, explicit claims, feasibility verdicts, literature-differentiation deltas, reviewer concerns and responses, and the literature-breakdown note. This lets an inspector trace why the candidate survived without turning the card into an audit dump.

\begin{tcolorbox}[breakable,colback=bgskill,colframe=cblue!70,fonttitle=\small,title={\textbf{A condensed generated card} \hfill \textit{held-out ICLR~2026 seed: RL for long-context reasoning}},before skip=6pt,after skip=6pt]
\small
\textbf{Title.} Phase-Stratified Credit: Verifier-Free Per-Phase Advantage for Long-Context Reasoning RL \;(\emph{method:} PS-GRPO).

\textbf{Motivation (the bottleneck).} GRPO collapses each long rollout into one group-normalized score and broadcasts it uniformly over every token. But only a few phases---where the model traces the key chain and grounds a fact---decide correctness, while most tokens scan distractor text. As context grows toward 128K, the decisive phases stay a near-constant budget while the filler grows, so the useful signal per decisive token shrinks like $1/\text{length}$: the method collapses exactly in the long-context regime it was built for.

\textbf{Method (step 1 of 4).} Build the strongest equal-compute GRPO baseline; record each rollout's group-normalized advantage $A_g = (R_g-\bar R_q)/\sigma_q$, broadcast uniformly to all $T_g$ tokens, and its accuracy out to 128K---the reference curve. \emph{[Three further steps harvest a verifier-free per-phase credit density from the in-group correct/incorrect ordering; full card in App.~\ref{app:example-card}.]}

\textbf{Falsification (kill-switch).} Load-bearing variable: $C$, the fraction of advantage magnitude on the decisive phases. Negative control: permute the phase labels so the \emph{same} $C$ lands on random spans---if 128K accuracy survives, the gain came from generic variance reduction, not from localizing credit to the steps that determine correctness. \emph{(Full prediction with positive control in App.~\ref{app:killswitch}.)}
\end{tcolorbox}

The falsification prediction is a single paragraph naming the minimal experiment and the metrics that should move. It gives qualitative direction but no pre-committed numerical thresholds. When the candidate makes a mechanism claim, the schema requires one \emph{load-bearing variable}: the quantity whose behavior carries the claim. It also requires a negative-control intervention whose predicted effect lands on the downstream task outcome, not on the variable's own definitional transform. This blocks a recurring self-confirmation failure: ``intervene on $X$ so that $X$ becomes zero.'' That tests a definition, not a mechanism. Tying the control to an independent downstream effect makes the claim falsifiable. The distinguisher is required for mechanism candidates and optional for pure ``does this work?'' candidates. The output excludes experiment matrices, ablation plans, baseline tables, expected figures, and calendar projections. These belong to the user's experimental engineering, not to the idea the skill produces.

\subsection{Faithfulness: resisting hallucination}
\label{subsec:faithfulness}

The standing objection to LLM ideation is fabrication: invented citations, papers, numbers, or method detail asserted with false confidence. IdeaSpark distributes its defenses across the phases in \S\ref{subsec:skill-workflow}. The enforced property is simple: every load-bearing claim either traces to a retrieved record or is explicitly marked as not doing so.

\paragraph{Grounding over memory.} Retrieval is not left to the model's discretion. Phase~0 follows a fixed orchestration path rather than an in-context tool the model could replace with recall. Phase~1 refuses to start until the grounding bundle and full-text cache are present, so a bottleneck is not written from abstracts or memory alone. The method-lineage tree extends the same discipline backward in time. Ancestors supplied from model memory enter only as non-citable, awareness-only nodes. They make no novelty claim. A subtractive gap may name only an assumption the retrieved methods actually exhibit.

\paragraph{Citation faithfulness.} A deterministic gate checks every sub-pattern citation when the candidate is written. It halts the run on a hallucinated parent, a mis-filed sub-pattern, or an invented name before any costly Phase~3 retrieval is spent. The pattern cards carry paper-agnostic lessons with no per-paper identifiers. This removes, rather than merely polices, the surface on which a \texttt{paper\_id} could be hallucinated. A cleared audit may legitimately report no threat found; fabricating a generic threat to fill the slot is forbidden.

\paragraph{Claim faithfulness.} Two guards bind the candidate's own assertions. Quantitative claims about predicted behavior, such as a rate, exponent, or scaling constant, must be cross-checked against same-setting papers in the full-text cache. The claim must then either align to that value or state the regime difference. This closes the recurring failure of pasting a number across incompatible regimes. The two kill-switch fields, falsification prediction and compute budget, are held byte-identical from generation through rendering by a hard validator. The model cannot silently soften the claim that makes the idea testable.

\paragraph{Honest abstention.} When the evidence does not support a proposal, the run stops with a \texttt{do\_not\_generate} or \texttt{phase\_3\_failed} report rather than producing one anyway. When the implementability audit reaches a load-bearing detail it cannot supply without guessing, it records that detail as an open author decision. An honest gap is treated as more valuable than fluent filler.

Together, these mechanisms turn ``do not hallucinate'' from an instruction the model may ignore into a property enforced by retrieval gates, deterministic validators, and provenance discipline.

\section{Evaluation: Generated-Idea Quality and Novelty}
\label{sec:evaluation}

This section reports an \emph{automated} endpoint evaluation. It compares IdeaSpark against baselines on two axes, \emph{quality} and \emph{novelty}, using two auxiliary judging skills rather than human reviewers. The headline result is twofold. First, at matched length and format, judged blind, IdeaSpark's idea cards receive the highest quality scores by a wide margin while remaining competitively novel. Second, novelty alone is misleading. The evaluation exposes a ``novel-but-empty'' failure mode, in which an idea appears highly novel because it is too vague to collide with prior art. We therefore treat the quality--novelty plane as the primary readout. As discussed in \S\ref{sec:limitations}, both judges are LLM-based skills, so this is an automated-judge study, not a human acceptance study.

\subsection{Systems compared}
\label{subsec:eval-systems}

We compare four idea sources on a shared set of problem directions (Table~\ref{tab:eval-systems}). The comparison points form a ladder of controls around IdeaSpark. \textbf{IdeaSpark} is the system of this report. \textbf{Opus-self-gen} is a generic ideation skill automatically authored by an Opus skill-creator, with no corpus grounding. Comparing it with IdeaSpark isolates the effect of \emph{our} corpus-grounded skill from the effect of merely having \emph{some} structured skill. \textbf{Opus-4.8 (bare)} is a single prompt with no skill and no retrieval. IdeaSpark, Opus-self-gen, and Opus-4.8 (bare) all run on the same Opus~4.8-high backbone. Together they form a same-backbone ladder: bare model, generic auto-built skill, and corpus-grounded skill. \textbf{GPT-5.5 (bare)} is a second bare baseline from a different model family, included to separate backbone-specific effects from skill effects.

\begin{table}[!htbp]
\centering
\caption{The four idea sources compared. ``Skill'' = produced by a structured multi-phase skill; ``Retrieval'' = the generator read live literature while producing the idea.}
\label{tab:eval-systems}
\small
\begin{tabular}{p{3.2cm} p{7.2cm} c c}
\toprule
\textbf{Source} & \textbf{What it is} & \textbf{Skill} & \textbf{Retrieval} \\
\midrule
IdeaSpark & This report's method; corpus-grounded multi-phase skill, on Opus~4.8 high & \checkmark & \checkmark \\
Opus-self-gen & Generic ideation skill auto-authored by an Opus skill-creator (no corpus grounding), run on Opus~4.8 high: live web search $\rightarrow$ cross-paper gap $\rightarrow$ idea & \checkmark & \checkmark \\
Opus-4.8 (bare) & Bare Opus~4.8 high: single prompt, no skill, no retrieval & --- & --- \\
GPT-5.5 (bare) & Bare GPT-5.5: single prompt, no skill, no retrieval (different model family) & --- & --- \\
\bottomrule
\end{tabular}
\end{table}

\subsection{Problem seeds: 100 method-agnostic directions from ICLR 2026}
\label{subsec:eval-seeds}

Each system is asked to produce one idea per \emph{problem seed}. We construct 100 seeds from ICLR~2026 Oral acceptances retrieved through the OpenReview API. ICLR~2026 is chosen deliberately: it post-dates the induction corpus (2021--2025) and the assumed training data of the generators and judges. The seeds therefore act as a forward, held-out test rather than a rediscovery exercise.

To avoid leaking each paper's own solution into the prompt, each seed is a \emph{method-agnostic rewrite} generated from the title alone. The rewrite keeps the domain object and task or capability, but strips method suffixes and any abstract-derived bottleneck diagnosis. For example, a paper title containing a specific mechanism for physical realism in video becomes the seed ``\emph{a novel ML research idea about physical realism in text-to-video models}.'' One seed is generated per paper and screened for leakage. The resulting seeds name \emph{what problem} to work on without hinting \emph{how}.

\subsection{Output normalization (controlling confounds)}
\label{subsec:eval-norm}

Quality and novelty judgments are sensitive to surface confounds, so all four sources are normalized to the same deliverable contract before judging. Every idea is converted to a three-section Markdown document: Title, Motivation, and Method. Motivation is held to $[259,330]$ words and Method to $[449,866]$ words, matching the length distribution of the source ICLR Oral papers. This removes the length and verbosity confound that otherwise rewards longer write-ups.

Every Method must also include formal equations with per-equation interpretation. This requirement matches the style of the source papers and puts vague prose and precise formalization on the same footing. The goal is to make judges score substance rather than surface technicality. We also strip protected ``author-decision'' annotations from IdeaSpark cards before scoring. These annotations are honest flags of unresolved design choices, but no other source emits them. Leaving them in would impose an asymmetric ``honesty tax''; a controlled comparison in \S\ref{subsec:eval-analysis} quantifies that effect.

\subsection{Two automated judges}
\label{subsec:eval-judges}

\paragraph{Quality: the \texttt{idea-quality} skill.} Quality is scored by a self-contained skill that judges an idea from first principles on three axes, each assessable without experiments. The axes are \emph{(A) problem position}, whether the attacked gap is real, important, and non-obvious rather than a soft target; \emph{(B) method quality}, decomposed into depth, soundness, and feasibility, with depth dominating; and \emph{(C) problem-fit}, whether the method actually closes the stated gap. The skill requires a quoted phrase of evidence for every axis score. It judges substance rather than length or fluency, and applies an A/C gate so that a strong method on a trivial problem cannot rank highly. For comparison, it runs in \emph{listwise} mode, ranking several ideas for the same seed against one another while blind to source identity. Relative judgments are less noisy than absolute scores.

\paragraph{Novelty: the \texttt{scoop-check} skill.} Novelty is scored by a prior-art collision skill. For each idea, the skill performs a \emph{live} literature search, deep-reads the closest candidates, and matches the proposed contribution against each prior work on four axes: problem framing, core mechanism, key insight, and application domain. The number of matching axes maps to a five-level scale, $\text{level} = 5 - (\text{axes matching})$. Thus \textbf{Level~5 = no overlap (most novel)} and \textbf{Level~1 = all four axes match (fully scooped)}. The verdict for an idea is the worst case, or minimum level, over its closest retrieved prior works. We take the worst case rather than the mean because one sufficiently close prior work can scoop an idea. Averaging would let unrelated retrieved papers dilute a decisive collision and inflate novelty. Higher is more novel, so novelty and quality point in the same direction.

\subsection{Protocol and metrics}
\label{subsec:eval-protocol}

Both axes use a blind, repeated protocol over all 100 seeds. For each seed, the four systems' ideas are copied to neutral filenames under a fresh random label permutation. The shuffle differs by seed and by round to cancel position bias. Source identity is withheld from the judge. \textbf{Quality} is a listwise rank: rank 1 scores $4$, rank 2 scores $3$, rank 3 scores $2$, and rank 4 scores $1$. \textbf{Novelty} is the Level~$\in\{1,\dots,5\}$ defined above.

Each seed is judged in \emph{three independent rounds} with fresh judge instances. For quality, we report each system's mean over the 100 seeds after averaging the three rounds per seed. For novelty, we report the mean level over all 300 judgments. Quality and novelty are run as separate passes.

\subsection{Results}
\label{subsec:eval-results}

\begin{figure}[!htbp]
\centering
\includegraphics[width=0.86\textwidth]{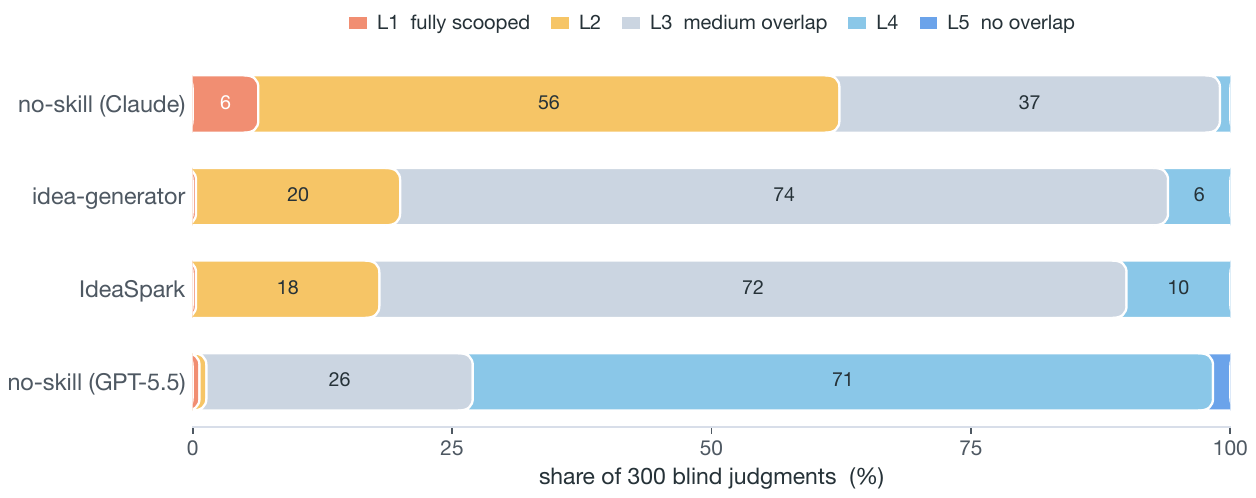}
\caption{Distribution of novelty levels per system (L1 = fully scooped $\rightarrow$ L5 = no overlap). The skill generators concentrate at L3 (medium overlap: shared framing/domain, distinct mechanism); the bare GPT-5.5 baseline piles up at L4, where its vagueness evades collision.}
\label{fig:novelty_dist}
\end{figure}

Tables~\ref{tab:eval-quality} and~\ref{tab:eval-novelty} report the two axes. Figure~\ref{fig:quality_novelty} places the four systems in the quality--novelty plane. Figure~\ref{fig:novelty_dist} shows the full distribution of novelty levels.

\begin{table}[!htbp]
\centering
\caption{\textbf{Quality} (\texttt{idea-quality} listwise rank score, $1$--$4$, higher is better). Each of the 100 seeds is judged in 3 blind rounds; mean and std are taken over the 100 seeds, each averaged across its 3 rounds. ``Wins'' is the number of seeds the system ranked first on its 3-round average (out of 100).}
\label{tab:eval-quality}
\small
\begin{tabular}{l c c c}
\toprule
\textbf{System} & \textbf{Quality} (mean) & \textbf{std} & \textbf{Wins / 100} \\
\midrule
IdeaSpark & \textbf{3.87} & 0.35 & \textbf{88} \\
Opus-self-gen & 2.57 & 0.55 & 6 \\
Opus-4.8 (bare) & 2.56 & 0.57 & 6 \\
GPT-5.5 (bare) & 1.00 & 0.00 & 0 \\
\bottomrule
\end{tabular}
\end{table}

\begin{table}[!htbp]
\centering
\caption{\textbf{Novelty} (\texttt{scoop-check} level, $1$--$5$, higher is more novel). Each of the 100 seeds is judged in 3 blind rounds; mean, std, and the level distribution are taken over all 300 judgments (100 seeds $\times$ 3 rounds). Right block: how the judgments distribute over levels L1 (fully scooped) to L5 (no overlap).}
\label{tab:eval-novelty}
\small
\begin{tabular}{l c c | c c c c c}
\toprule
\textbf{System} & \textbf{Novelty} (mean) & \textbf{std} & \textbf{L1} & \textbf{L2} & \textbf{L3} & \textbf{L4} & \textbf{L5} \\
\midrule
GPT-5.5 (bare) & \textbf{3.73} & 0.53 & 2 & 2 & 77 & 214 & 5 \\
IdeaSpark & 2.92 & 0.53 & 1 & 53 & 216 & 30 & 0 \\
Opus-self-gen & 2.86 & 0.50 & 1 & 59 & 222 & 18 & 0 \\
Opus-4.8 (bare) & 2.32 & 0.60 & 19 & 168 & 110 & 3 & 0 \\
\bottomrule
\end{tabular}
\end{table}

\subsection{Analysis}
\label{subsec:eval-analysis}

\paragraph{IdeaSpark wins on quality.} IdeaSpark attains the highest quality score ($3.87/4$) and ranks first on $88$ of the $100$ seeds by 3-round average. No other system exceeds $2.6$. The generic auto-authored skill, Opus-self-gen, scores $2.57$, essentially tied with the bare Opus-4.8 baseline at $2.56$. A structured skill alone therefore does not lift idea quality.

The same-backbone ladder isolates the effect of IdeaSpark's specific design. IdeaSpark, Opus-self-gen, and Opus-4.8 (bare) share one backbone (\S\ref{subsec:eval-systems}). Opus-self-gen also has live retrieval. The observed gain is therefore not explained by model family or by retrieval alone. The lift comes from IdeaSpark's corpus-grounded pattern cards together with its multi-phase, audit-anchored workflow.

\paragraph{Single-axis novelty is misleading.} On novelty alone, the \emph{worst} system wins. The bare GPT-5.5 baseline scores highest on novelty ($3.73$, with $214/300$ judgments at L4), but is last on quality ($1.00$, zero wins). The judges' rationales explain the mismatch. This baseline emits a near-identical topic-agnostic template, a generic ``diagnostic-heads + contrastive-pairs + uncertainty-routing'' scaffold that occasionally drifts off topic. Because the idea is vague, it has \emph{no precise prior art to collide with}, so it scores as ``novel.''

This is the ``novel-but-empty'' regime. It is invisible to either axis alone. Only the quality$\times$novelty plane (Figure~\ref{fig:quality_novelty}) separates genuine novelty, high on both axes, from vacuous novelty, high on novelty but low on quality. The two skill-based systems instead sit in the high-quality region while remaining at medium overlap (L3). Their ideas share problem framing and domain with the closest prior work but contribute a distinct mechanism, an honest and defensible novelty profile.

\paragraph{Calibration sub-study (honesty annotations).} On a $5$-seed pilot, leaving IdeaSpark's self-flagged author-decision annotations in the card cost roughly one rank position, dropping it behind the bare Opus-4.8 baseline. Removing them restored its lead. These annotations are author-facing meta-notes rather than part of the idea, and no other source emits them. This motivates the normalization in \S\ref{subsec:eval-norm}: strip the annotations before scoring.

\section{Limitations}
\label{sec:limitations}

Each limitation below is a bounded property of the current study paired with a concrete next step. None changes the claims we make, which remain scoped to the idea stage. They mark where additional evidence would extend those claims.

\paragraph{Coverage is bounded by three ML conferences, which biases the induced distribution.} The corpus is built entirely from ICLR, ICML, and NeurIPS outcomes and their public reviews, and the evaluation seeds are all ICLR~2026 Orals. The induced domains, the ideation patterns, and the implicit sense of what counts as a strong direction therefore track the distribution of mainstream ML-conference research rather than science in general. Fields and venues under-represented in these proceedings are correspondingly under-covered, so generated ideas can inherit this bias and need not transfer to other communities. We write the patterns to be domain-agnostic to limit topic leakage, and the novelty pass checks overlap with live literature, but we do not enforce as-of backdating of the retrieval window. Cross-venue and cross-discipline corpora, backdated retrieval for a leakage-free forward-prediction variant, and cross-backbone judges are the direct ways to test and widen external validity.

\paragraph{The endpoint evaluation is automated, with a human study as the next stage.} Both axes are scored by LLM-based skills, so the results measure agreement with structured automated reviewers rather than human program-committee judgment, and the judges may share blind spots with the generators. An automated study is the appropriate first stage at the idea level; a blind expert study is the planned next step before any acceptance-level claim.

\paragraph{The judge metrics carry two known biases.} First, \texttt{idea-quality} rewards a novel \emph{mechanism}, so benchmark, systems, and pure-measurement contributions score lower on its method-depth axis by construction; we therefore read quality as \emph{relative} among generation systems, not as an absolute measure of research value. Second, novelty is meaningful only jointly with quality, since a sufficiently generic idea can evade prior-art collision and score as novel (the GPT-5.5 case), which is why we never rank on novelty alone but report it as one axis of a quality--novelty plane. Rubrics for resource and systems contributions, and a specificity-aware novelty metric that penalizes under-committed ideas, would reduce both biases.

\section{Artifact Cards}
\label{sec:release}

\subsection{Data card}

The corpus (Section~\ref{sec:data}) uses public metadata and public review material. Author identities are retained as paper metadata but are not used as model features for pattern induction or recommendation.

\subsection{Model and backend card}

The current pipeline uses LLMs for extraction, taxonomy induction, card generation, and several skill phases, plus an embedding model for clustering. The principal models are Claude Sonnet~4.6 for signature extraction and paper-level tagging; Claude Opus~4.7 for taxonomy / domain induction and pattern-card generation; OpenAI \texttt{text-embedding-3-large} for strategy-signature clustering; and Claude Opus~4.8 high for the skill runtime and endpoint evaluation in \S\ref{sec:evaluation}. GPT-5.5 is used only as the cross-family baseline. These choices are backend-specific, and cross-backend quality and format-validity measurement remains future work.

\subsection{Skill card}
\label{subsec:skill-card}

IdeaSpark is appropriate for literature-grounded ML research directions where the user can state a domain, adjacent literature, constraints, and a desired contribution type. It is not appropriate for confidential literature without connector permission, broad agenda-setting prompts with no adjacent papers, pure benchmark construction, complete paper writing, automated submission production, or claims requiring private peer-review data. It returns a structured idea card, a \texttt{do\_not\_generate} diagnosis, or a \texttt{phase\_3\_failed} report, but it cannot guarantee novelty, acceptance, feasibility, or reviewer agreement. Users remain responsible for prior-art review, experimental design, feasibility assessment, ethical review, and downstream claims.

\section{Responsible Use}
\label{sec:responsible-use}

Research-ideation systems reduce the cost of producing plausible-looking proposals, which also increases misuse risks: low-quality proposal generation, reviewer burden, shallow arXiv submissions, and overconfident novelty claims. IdeaSpark is therefore scoped as an ideation scaffold, not a paper-writing agent. It outputs one idea card rather than a manuscript, includes failure paths, audits candidates against corpus-derived reject lessons and prior-art threats, and keeps falsification and feasibility checks visible.

Because the method draws on \emph{rejected} submissions and reviews, we treat that signal as sensitive. Rejected papers are used only as aggregate, pattern-level evidence about how strategic moves fail, never to rank, profile, or single out papers, authors, or reviewers. Author identities are retained only as metadata and are not model features (\S\ref{subsec:meta-coverage} / \S\ref{sec:release}). Public deployment should include rate limits, artifact provenance, citation grounding, warnings that the system does not predict acceptance, and release rules for review-text redistribution and removal requests.

\section{Conclusion}
\label{sec:conclusion}

We presented ResearchStudio-Idea, a three-skill suite for the first mile of literature-grounded ML research ideation. Paper-Search provides standalone multi-source literature grounding, Scoop-Check provides standalone prior-art collision checking, and IdeaSpark composes these functions with outcome-grounded pattern guidance to produce one auditable idea card. The empirical contribution is a 1{,}947-paper analysis of ICLR, ICML, and NeurIPS outcomes from 2021--2025, inducing 15 ideation patterns, 31 sub-patterns, and 28 research domains. The central system artifact is IdeaSpark: an operational card library and skill specification that make those patterns executable at inference time, with the same contrastive cards guiding bottleneck-to-move selection and candidate audit.

The main claim is deliberately bounded. ResearchStudio-Idea does not show that an LLM can generate Oral papers, predict acceptance, or replace experimental judgment. It shows that public conference outcomes contain reusable strategic operators, that rejected and accepted papers often share those operators, and that their success and failure conditions can be packaged into practical skills for search, generation, and review. The automated-judge evaluation in Section~\ref{sec:evaluation} provides initial endpoint evidence: at matched length and format, IdeaSpark improves judged idea quality over no-skill and generic-skill baselines while remaining competitively novel. The remaining gap is validation beyond the idea stage, including human reviewer preference, implementation success, and transfer outside the studied ML conference setting. Its value is to make pre-experimental ideation more structured, evidence-grounded, and auditable while leaving final prior-art review, experimental design, and scientific claims with the researcher.

\appendix
\section*{Appendix}
\addcontentsline{toc}{section}{Appendix}
One generated card and its JSON evidence trail, a reject-path run, and a pattern\,/\,sub-pattern card --- all real artifacts from \S\ref{sec:evaluation}, reproduced verbatim.

\section{End-to-end Generated Idea Card}
\label{app:example-card}
IdeaSpark's verbatim \texttt{std}-register output for a held-out ICLR~2026 seed (\emph{RL for long-context reasoning}), the system's rank-1 card in \S\ref{sec:evaluation}; its JSON evidence trail (core claim, falsification, literature deltas) is in App.~\ref{app:killswitch}.

\begin{center}
\fbox{\includegraphics[page=1,width=0.76\linewidth]{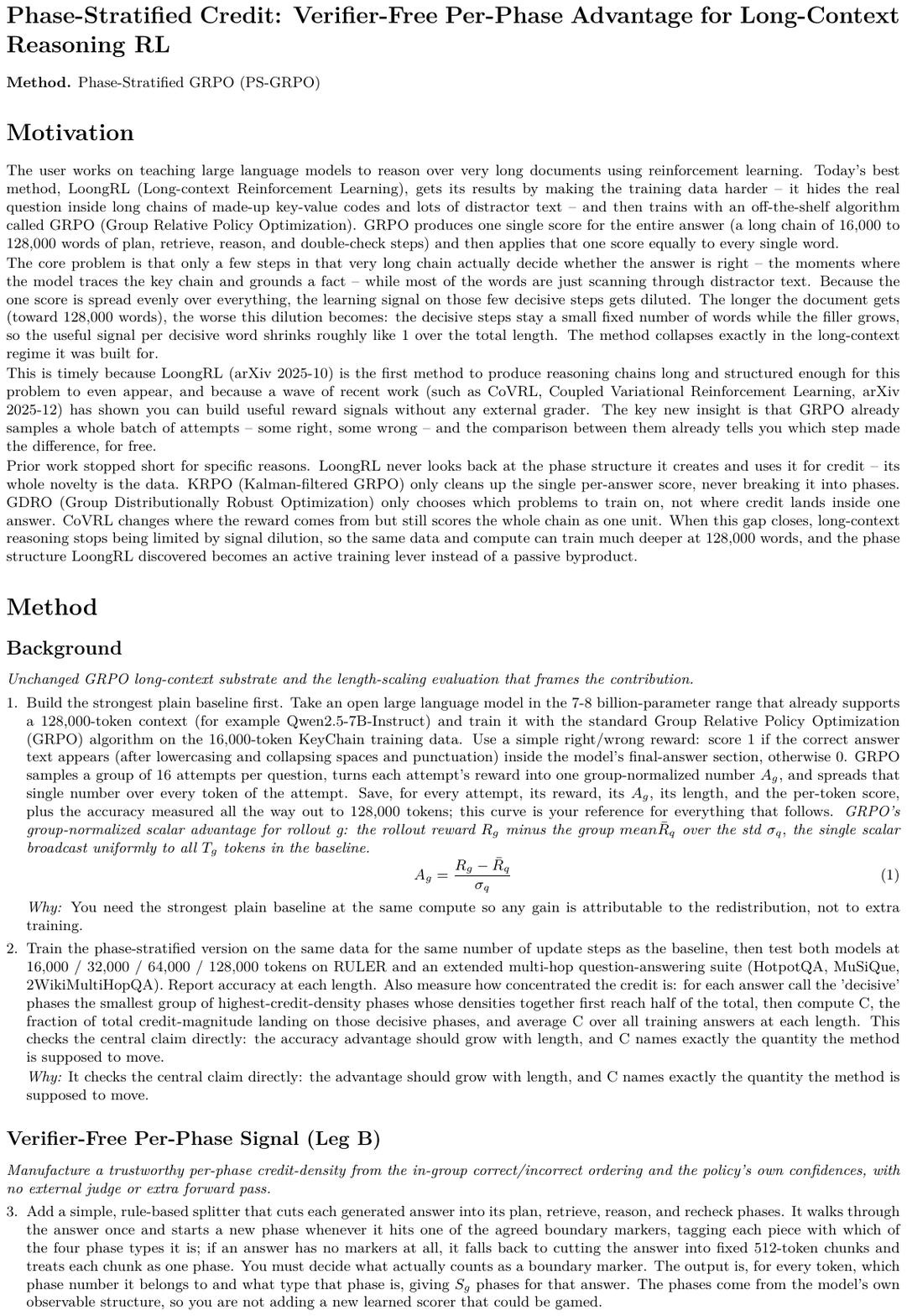}}\\[5pt]
\fbox{\includegraphics[page=2,width=0.76\linewidth]{figures/fig_example_card_std_en.pdf}}
\end{center}

\section{Kill-switch and Falsification Fields}
\label{app:killswitch}
These fields back the same card but are deliberately \emph{not} printed on it: they are the
reviewer-defensible machinery of \S\ref{sec:methodology} (mechanism-aware falsification with a named
load-bearing variable plus a non-tautological negative control, and byte-identical kill-switch
preservation across phases). Excerpted verbatim from \texttt{phase4\_expansion.json}.

\begin{lstlisting}[breaklines=true,basicstyle=\ttfamily\footnotesize,frame=single,framerule=0.4pt,xleftmargin=4pt,xrightmargin=4pt,aboveskip=4pt,belowskip=2pt]
{
  "core_claim": "Redistributing GRPO's single trajectory-level advantage onto a verifier-free, in-group-harvested per-phase credit-density -- with nothing added -- recovers the long-context learning signal that uniform broadcast dilutes as context grows to 128K.",
  "compute_budget": "30 A100-day (within the 90 A100-day default; comfortable). Two GRPO training runs on a single 7-8B policy over 16K-token KeyChain data plus 128K evaluation are the cost; leg B adds no extra forward passes (it reuses the log-probs already produced during sampling), and there is no separately trained reward model or critic to fund.",
  "falsification_prediction": "Minimal experiment: train two checkpoints from the same base LLM on identical 16K KeyChain data and equal compute -- (i) vanilla GRPO (LoongRL substrate) and (ii) phase-stratified GRPO -- and evaluate both at 16K/32K/64K/128K on RULER + extended multi-hop QA. If the candidate works, the accuracy gap (ii minus i) INCREASES with context length, i.e. stratified credit is roughly on par at 16K but its 128K answer accuracy is materially higher, because the pivotal phases stay a near-constant token budget while the distractor bulk grows. The single named load-bearing variable is the per-phase advantage concentration ratio C (fraction of total advantage magnitude placed on pivotal-phase tokens); the mechanism claim is that gains are caused by concentrating credit on the pivotal minority. Negative control: permute the phase labels so the density d lands on randomly chosen spans instead of the contrast/self-score-identified pivotal phases (C is held at the SAME numeric level, only its placement is randomized) -- the prediction is that downstream 128K answer accuracy collapses back to the vanilla-GRPO curve, NOT that C changes. This is the real Popper test: if accuracy survives random placement, the gain came from a generic variance-reduction or extra-compute effect rather than from localizing credit to the steps that determine correctness. Positive control: a stripped model that keeps ONLY the minimal-pair contrast term (drop self-score) should recover most of the 128K accuracy gain, identifying the in-group correct/incorrect ordering as the load-bearing source of the per-phase signal.",
  "differentiation_from_lit[0]": {
    "paper_id": "openalex:W4416246797",
    "delta": "LoongRL leaves the GRPO advantage A_g untouched and broadcasts it uniformly across the whole 16K-128K rollout, concentrating its novelty in KeyChain data construction; we construct a per-phase stratified advantage A_{g,s} that redistributes that same A_g onto the pivotal-phase minority, an artifact their unmodified-optimizer design never produces."
  }
}
\end{lstlisting}

\section{Reject-path Output (Phase~3 Abandon)}
\label{app:reject}
Not every direction yields a card. When the Phase~3 audit hits an unaddressable hard floor, the run
emits a \texttt{phase\_3\_failed.md} instead of an idea (one of the three one-shot outcomes of
\S\ref{sec:methodology}). The example below is a real run on a seed whose direction was, essentially,
the title of an existing 2025 paper; the corpus-anchored audit caught the exact-mechanism prior-art
collision and abandoned rather than ship a scooped idea.\footnote{In the evaluation harness this run
was subsequently re-routed onto a different Phase~2.1 gap to still produce a card for scoring; the
abandon artifact reproduced here is the genuine output of the abandon branch.}

\begingroup\small

\textbf{Run:} run\_prompt85\_tns
\textbf{Seed direction:} "A novel ML research idea about Non-Asymptotic analysis of (sticky) Track-and-Stop."
\textbf{Candidate generated (Phase 2.2):} "A Non-Asymptotic Sample-Complexity Certificate for Sticky Track-and-Stop"
\textbf{Verdict:} abandon
\textbf{Verdict layer:} hard\_floor

\subparagraph{\textbf{Triggering checks}}\

\paragraph{1. paper\_pointed\_threat --- exact-mechanism collision (hard floor, unaddressable)}
\begin{itemize}[leftmargin=*]
\item \textbf{Threat paper:} \texttt{semanticscholar:63d8201fbe32063e2ac336b0708f589e05cc3759} --- "Non-Asymptotic Analysis of (Sticky) Track-and-Stop" (Poiani, Bernasconi \& Celli, arXiv:2505.22475, 2025-05). Surfaced via Phase 3.1 collision retrieval.
\item \textbf{Subsumption:} The threat abstract states verbatim that Track-and-Stop and Sticky Track-and-Stop "are optimal in the asymptotic regime [but] their non-asymptotic guarantees remain unknown. In this work, we fill this gap and provide non-asymptotic guarantees for both algorithms." This is byte-for-byte the candidate's headline deliverable (a finite-confidence / all-delta guarantee for the genuine sticky-T\&S sampling-plus-stopping rule, algorithm held fixed).
\item \textbf{Candidate's sole differentiator fails:} the named additive burn-in functional B(mu), built from the cumulative empirical-vs-oracle allocation deficit D\_t, is an INTERNAL proof quantity. Any non-asymptotic bound on sticky-T\&S must already control exactly this tracking-error/burn-in term, so the prior paper necessarily bounds it (named or not). Not a distinct external claim.
\item \textbf{addressable\_via:} null (cannot be repaired by a field edit; the headline deliverable already exists in print).
\end{itemize}

\paragraph{2. gap\_closure\_reject\_check --- triggered}
\begin{itemize}[leftmargin=*]
\item \textbf{C19 lesson fires (candidate\_match=yes):} "A relaxation the field reads as a mechanical lift of a known argument will be judged incremental unless a genuinely novel obstacle is named and overcome." The obstacle (non-asymptotic guarantees for sticky T\&S) is already overcome in arXiv:2505.22475.
\item \textbf{C29 lesson fires (candidate\_match=yes):} "Re-deriving a rate that merely matches the baseline with near-identical algorithmic changes is read as an incremental extension." The candidate holds the sticky-T\&S algorithm fixed and re-derives the same object the prior paper already certifies.
\end{itemize}

Recipe application check = applied (both legs perform their cited cluster's signature move) and anti\_pattern check = no match --- i.e. the failure axis is prior-work overlap, NOT recipe misuse or anti-pattern composition.

\subparagraph{\textbf{Root cause}}\
Two independent hard-floor / triggered findings point at the same root cause: the user's seed direction is, essentially, the literal title of an existing 2025 paper. The novelty residual the pipeline could find (a named B(mu) burn-in term) is an internal lemma object, not an externally differentiating claim.

\subparagraph{\textbf{User-side options}}\
\begin{itemize}[leftmargin=*]
\item \textbf{Drop / replace the direction.} Non-asymptotic analysis of (sticky) Track-and-Stop as a headline is already occupied by the 2025 paper arXiv:2505.22475.
\item \textbf{Re-frame to an unoccupied residual.} Possible pivots NOT covered by the threat paper, e.g.: (a) the forced-exploration / "sticky" schedule's finite-confidence tradeoff (a deferred gap in Phase 2.1) --- quantify how the forcing rate, set only for asymptotic consistency, governs the finite-delta overhead; (b) instance-dependent LOWER bounds on the burn-in (an impossibility result), rather than an upper bound that the prior paper already supplies; (c) a structured / non-exponential-family or contextual setting the prior non-asymptotic analysis does not reach. Each would require a fresh Phase 0 retrieval to confirm it is unoccupied.
\item \textbf{Re-run Phase 2.1 + 2.2 on a different selected gap} from \texttt{phase1\_output.json} (4 gaps were found) with the threat paper now in scope, then re-critique.
\end{itemize}

\subparagraph{\textbf{Artifacts on disk}}\
\begin{itemize}[leftmargin=*]
\item phase0/ (lit\_table.md, lit\_results.json, fulltext\_cache.json)
\item phase1/phase1\_output.json (state=proceed)
\item phase2\_select/phase2\_select\_output.json
\item phase2\_generate/phase2\_generate\_output.json (passed citation gate)
\item phase3\_collision/collision\_hits.json (173 hits; critical: arXiv:2505.22475)
\item phase3\_critique/phase3\_critique\_output.json (verdict=abandon, hard\_floor)
\item phase\_3\_failed.md (this file)
\end{itemize}

No phase4/ produced. evaluation/opus4\_8/85 NOT created (abandon path).

\endgroup

\section{Exemplar Ideation Pattern and Sub-pattern Card}
\label{app:example-cards}
One of the 15 induced pattern cards (the high-coverage \emph{Audit and Pivot an Assumption} exemplar of
\S\ref{sec:methodology}), paired with one of its sub-pattern cards (the tactical recipe a Phase~2.2
candidate instantiates). Reproduced verbatim from the artifact deck; the full 15 patterns $\times$ 31
sub-patterns are in the GitHub artifact release.

\begin{tcolorbox}[colback=bgskill, colframe=cblue!70, title={\textbf{Audit and Pivot an Assumption} \textcolor{gray}{\small\texttt{assumption\_audit\_and\_pivot}} \hfill \textit{Audit the load-bearing assumption and pivot}}, breakable, before skip=4pt, after skip=4pt]
\textcolor{gray}{\small Cluster-level primary: Oral $n=94$ | HC $n=11$ | Reject $n=79$ | Total $n=181$}

\textbf{Definition.} Locate the load-bearing implicit assumption a result, guarantee, or defense rests on, then pivot on it: relax it to a weaker condition and re-prove (extending the guarantee), or violate it with a constructed counterexample/exploit (breaking the system or unlocking new behavior).

\textbf{Operational signature.} \textit{identify the implicit assumption a result or defense rests on $\to$ relax it (weaker condition) or violate it (counterexample/exploit) $\to$ re-derive the guarantee or demonstrate the new behavior}

\textbf{When to apply.} When a result's strength or a system's safety hinges on an assumption that real settings can weaken or that an adversary can violate.

\textbf{Step-by-Step:}
\begin{enumerate}[leftmargin=14pt,topsep=2pt,itemsep=1pt]
  \item Surface the single implicit assumption the target result, guarantee, or defense silently rests on --- the precondition the whole subfield has inherited as fixed (a quantity treated as immutable, a property treated as protective, a structure treated as required). Name it explicitly and show it is load-bearing: the existing result holds *because* of it. Do not pivot on an assumption whose removal leaves the achievable outcome unchanged --- relaxing a non-binding assumption yields only an incremental generalization, the cluster's most common rejection.
  \item Choose the pivot direction deliberately. Either relax the assumption to a strictly weaker condition and plan to re-derive the guarantee (extending it to settings previously thought impossible), or violate it with a constructed counterexample or exploit to break the system or unlock new behavior. The direction must change what is achievable --- reach an outcome the assumption was believed to forbid --- not merely restate the prior result under a cosmetic reframing.
  \item Establish that the pivoted condition is real, not a substitution of one strong assumption for another. For a relaxation, verify empirically --- in natural data or the actual deployment regime --- that the weaker condition genuinely holds, and confirm you have not smuggled in an equally restrictive or unverifiable replacement; the recurring failure is trading a known assumption for a distributional form or abstract regularity condition no real instance is shown to satisfy. For a violation, confirm the assumption is breakable by a realistic, resource-bounded actor rather than an idealized one.
  \item Re-derive rigorously and certify tightness. For a relaxation, prove the guarantee under the weaker condition and supply a matching lower bound, an exact equivalence, or an impossibility result, so the pivot is certified as the true barrier rather than an artifact of loose analysis. For a violation, demonstrate the exploit and trace its success to a structural mechanism of the pipeline, turning an isolated trick into a systematic, reusable template --- empirical success with no identified mechanism is read as anecdote.
  \item Prove the pivot was necessary and differentiate it. Show the prior approach fails precisely because of the audited assumption via an explicit counterexample or failure-case characterization, and sharply distinguish your move from any prior work that pivoted similarly. Avoid bundling off-the-shelf components and crediting the bundle, and never name a tool or framework as the contribution without exercising it in the actual derivation --- both are recurring rejection shapes in this cluster.
\end{enumerate}
\textbf{Success conditions} (from Oral):
\begin{itemize}[leftmargin=12pt,topsep=2pt,itemsep=2pt]
  \item \textbf{The relaxed or violated condition is shown to hold in the real setting, not merely assumed --- measured in natural data, or demonstrated reachable by a bounded adversary.} \\ A pivot is only valuable if the weaker precondition is actually satisfied where the method will be used; grounding the condition empirically converts a theoretical relaxation into a usable one and pre-empts the reviewer suspicion that one strong assumption was swapped for another.
  \item \textbf{A relaxed upper bound is paired with a matching lower bound, exact equivalence, or impossibility result that certifies the pivot as the genuine barrier.} \\ Tightness certificates separate a loose extension from a sharp one: they prove the audited assumption --- not the analysis technique --- was what limited prior results, which is the structural claim the contribution actually makes.
  \item \textbf{The audited assumption is replaced by a strictly weaker or structurally different one, often shifting the burden from the prior distribution to the geometry of the map or from pointwise to expectation-level conditions.} \\ Moving the constraint to a different and milder level of the problem is what unlocks new achievable results; a same-strength substitution re-imports the original barrier and leaves the frontier unchanged.
  \item \textbf{For exploits and violations, the success is traced to a structural property of the pipeline and packaged as a systematic design template rather than a one-off attack.} \\ A root-cause-level explanation generalizes across instances and gives the community a reusable lens, whereas an isolated demonstration is fragile and reads as engineering rather than insight.
  \item \textbf{Prior methods are shown to fail precisely because of the audited assumption, via explicit failure-case characterization presented before the positive result.} \\ Demonstrating that the assumption is the binding constraint --- not an incidental simplification --- is what makes the pivot necessary rather than optional, and reviewers consistently reward this clarity.
\end{itemize}
\textbf{Failure modes} (from Reject):
\begin{itemize}[leftmargin=12pt,topsep=2pt,itemsep=2pt]
  \item \textbf{Pivoting on an assumption whose relaxation produces no new achievable result, yielding a generalization that reviewers read as incremental.} \\ When the weaker condition leaves the attainable guarantee or rate unchanged, the contribution collapses to 'we also handle case X' with no unlocking; the absence of a qualitatively new capability is the most cited reason these attempts fall short.
  \item \textbf{Replacing the audited assumption with an equally strong or unverifiable substitute --- a restrictive distributional form, or an abstract regularity condition no known instance is shown to satisfy.} \\ The substitution re-imports the very barrier the paper claimed to remove, and reviewers flag that the new condition is untestable or unmotivated, so the claimed relaxation is not actually a relaxation.
  \item \textbf{Demonstrating an exploit or violation without validating it against an adaptive or informed defender.} \\ The central claim of a violation is that the property is hard to counter; showing only that an attack succeeds against a static target leaves that claim unproven, and the 'commonness implies safety' or 'stealth' arguments remain intuitive rather than substantiated.
  \item \textbf{Stacking several off-the-shelf components and attributing the gain to the bundle without isolating which pivot is load-bearing.} \\ Reviewers cannot identify a unifying principle and read the work as an engineering combination; without an ablation pinning the benefit to a specific structural move, the novelty above the constituent parts is unconvincing.
  \item \textbf{Naming a theoretical tool or framework as the contribution while never exercising it in the proofs, or shipping an error in the stated result.} \\ If the advertised machinery (a continuous-time analysis, a new perspective) is absent from the actual derivation, or a headline rate is wrong, confidence in the entire pivot evaporates regardless of the idea's appeal.
  \item \textbf{Failing to sharply differentiate from prior work that already performed a near-identical prior pivot.} \\ In crowded areas a correct but under-differentiated move reads as redundant; reviewers point to the closest precedent and demand a decisive reason to prefer the new approach over that pivot, which an unfocused positioning cannot supply.
\end{itemize}
\textbf{Oral vs Reject gap.} Accepted executions perform three observable moves that rejected ones skip. First, after naming the assumption they verify the pivoted condition concretely --- measuring that the weaker condition holds in real or natural data, or that the violated property is reachable by a bounded adversary --- whereas rejected papers relax an assumption and immediately substitute an equally strong or unverifiable one (a linear-Gaussian form, an abstract regularity condition with no exhibited satisfying function class). Second, accepted relaxations ship a tightness certificate --- a matching lower bound, an exact algebraic equivalence, or an impossibility result --- while rejected ones offer only a one-directional upper bound or a bare empirical demonstration, leaving open whether the pivot was the real barrier or just loose analysis. Third, accepted papers establish that prior methods fail precisely because of the audited assumption via an explicit counterexample or failure-case characterization, and for exploits they trace success to a structural pipeline mechanism that becomes a reusable template; rejected papers instead present the work as a generalization or an isolated trick, bundle off-the-shelf components without isolating the load-bearing pivot, or never exercise the theoretical tool they claim as their contribution.

\textbf{Oral vs HC gap.} With a moderate HC sample (\textasciitilde{}10 papers, several of them attacks, benchmarks, or scaling studies), the pattern is that HC papers pivot on an assumption and characterize the consequence at scale or package it as urgently needed reusable infrastructure --- a power-law fit for an attack, a measured safety/helpfulness trade-off, a large boundary-targeting benchmark, a broadly transferable defense --- but typically stop at empirical characterization or a single bound. Oral papers add a tightness certificate on top of the same pivot: a matching lower bound, an exact equivalence, an identifiability proof, or an impossibility/optimality result that converts 'this works and here is how well' into 'this is provably the right lever and here is the exact boundary.' Oral papers also more often formalize the failure modes of prior approaches before presenting the positive result, so the pivot reads as a principled diagnosis rather than a strong empirical finding. Given the sample size this is a tendency rather than a strict law.

\textbf{Reviewer expectations:}
\begin{itemize}[leftmargin=12pt,topsep=2pt,itemsep=1pt]
  \item \textit{[both]} When a result rests on a relaxed assumption, supply a matching lower bound or tightness certificate confirming the analysis is not loose; the absence of a matching bound is repeatedly cited as a reason to hold back.
  \item \textit{[reject\_reviews]} Characterize the new or weakened assumption --- show which function classes, distributions, or real settings actually satisfy it --- rather than introducing it as a convenient condition that merely lets the proof go through.
  \item \textit{[reject\_reviews]} For attacks and violations, validate against adaptive or informed defenders, not just a static target, before claiming the violated property is hard to counter.
  \item \textit{[both]} Sharply differentiate the pivot from prior work that made a similar move, giving a decisive reason to prefer it; overlap with a close precedent (a prior drift-correction, a prior attack surface) must be addressed head-on.
  \item \textit{[reject\_reviews]} Do not advertise a theoretical framework or tool as the contribution unless it is actually used in the derivation, and ensure headline rates and claims are error-free.
  \item \textit{[oral\_reviews]} Establish and cleanly characterize the failure cases of prior approaches before presenting the positive result, so the pivot is seen as a principled diagnosis of the binding constraint.
\end{itemize}
\textbf{Cognitive barriers:}
\begin{itemize}[leftmargin=12pt,topsep=2pt,itemsep=1pt]
  \item The audited assumption is invisible because the entire subfield inherited a framing that treats it as a fixed precondition rather than a controllable choice --- a 'fixed' local objective, a 'required' supervision signal, a trigger that 'must' live in the input. You cannot relax what you do not first see as a variable, so the hard part is denaturalizing a convention everyone reasons within.
  \item A stated safety or design property is mistaken for an inherent guarantee: practitioners read 'reduced probability mass,' 'norm-bounded imperceptibility,' or 'scale dilutes adversaries' as protection, which masks that the very property is the exploitable lever. Reframing a feature as a vulnerability requires inverting the intent the property was introduced to serve.
  \item The pivot often hinges on importing a tool from an adjacent domain and recognizing it transfers --- concentration tools from sequential testing, blocking from dependent-sequence analysis, scalarization from multi-objective optimization. Because the tool was built for a different purpose, its applicability to the audited assumption is not visible from within either field's standard practice.
  \item The relaxation frequently lives at a different level of the problem than the natural formulation suggests --- reparameterizing the latent basis instead of the observed variables, constraining the geometry of the generative map instead of the prior distribution, measuring increments or contrasts instead of raw values. The obvious formulation hides the lever, so the move feels unreachable until one deliberately changes the level of analysis.
\end{itemize}
\textbf{Examples} (paper-agnostic lessons):
\begin{itemize}[leftmargin=12pt,topsep=2pt,itemsep=1pt]
  \item \textbf{[Oral lesson]} When a guarantee is blocked by a quantity treated as fixed, ask whether that quantity is actually a controllable variable you can reshape cheaply each step --- turning a hard constraint into a degree of freedom often dissolves the barrier.
  \item \textbf{[Oral lesson]} Measure that the weaker condition you rely on actually holds in real data before claiming the relaxation is practical; a precondition verified in natural statistics is worth far more than one merely assumed.
  \item \textbf{[Oral lesson]} Pair every relaxed upper bound with a matching lower bound, exact equivalence, or impossibility result, so the pivot is certified as the true barrier rather than an artifact of loose analysis.
  \item \textbf{[Oral lesson]} Replace a distributional assumption with a structural or architectural one when the problem's geometry already encodes the constraint --- identifiability can rest on the shape of the generative map rather than on the prior over latents.
  \item \textbf{[Oral lesson]} When building an exploit, trace its success to a structural property of the pipeline and convert it into a reusable design template rather than presenting an isolated trick.
  \item \textbf{[Oral lesson]} Read every 'bounded,' 'small,' or 'imperceptible' guarantee as a budget an adversary can neutralize --- a stated protective property is frequently the exact lever to violate.
  \item \textbf{[Reject lesson]} Relaxing an assumption that leaves the achievable result unchanged reduces the contribution to a generalization reviewers read as incremental.
  \item \textbf{[Reject lesson]} Swapping the audited assumption for an equally strong or unverifiable one --- a restrictive distributional form, or a regularity condition no known instance is shown to satisfy --- re-imports the barrier you claimed to remove.
  \item \textbf{[Reject lesson]} Demonstrating an exploit only against a static defender leaves the central claim, that the violation is hard to counter, unproven.
  \item \textbf{[Reject lesson]} Bundling several off-the-shelf components and crediting the bundle, without isolating which pivot is load-bearing, invites a 'no unifying principle' rejection.
  \item \textbf{[Reject lesson]} Naming a theoretical framework as the contribution while never exercising it in the proofs --- or shipping an error in the stated rate --- collapses confidence in the whole result.
  \item \textbf{[Reject lesson]} Failing to sharply distinguish the work from a prior near-identical pivot makes even a correct result read as redundant in a crowded area.
\end{itemize}
\end{tcolorbox}

\noindent\textit{Representative sub-pattern under this pattern (the tactical recipe a Phase~2.2 candidate instantiates):}

\begin{tcolorbox}[colback=bgskill!50, colframe=cgreen!70, title={\textbf{C01} \textcolor{gray}{\small (O20/H9/R21)}}, breakable, before skip=4pt, after skip=4pt]
\textbf{Tactical pattern.} Across this cluster the audited assumption is always about the *representational locus* at which a security, detection, provenance, or safety property is presumed to be enforced --- and the tactical move is to relocate the attack or defense to a different locus the field treated as irrelevant or invariant. Concretely: triggers are moved from the input image to the supervision/label co-occurrence space; the comparison signal for malicious updates is moved from parameter/gradient divergence to scalar-score distribution geometry; the detection feature is moved from a scalar confidence to local curvature, positional-depth profile, or the full sampled output distribution; provenance is moved from raw parameter comparison to transformation-invariant kernel alignment over intermediate activations; defense interception is moved from the output boundary to internal state trajectories. For attacks, the papers show the overlooked locus is exploitable at far lower cost or access than the assumed locus implied; for defenses, they show the overlooked locus carries an attack-agnostic, transformation-invariant signal the assumed locus lacks. The unifying verbs are "relocate the signal," "re-anchor enforcement," and "demonstrate exploitability/robustness at the new locus."

\textbf{Step-by-Step:}
\begin{enumerate}[leftmargin=14pt,topsep=2pt,itemsep=1pt]
  \item Name the single representational locus that current attacks or defenses in the target security/detection setting all operate on (input perturbation, parameter/gradient divergence, scalar confidence score, deterministic mode output, externally-injected artifact), and state the tacit sufficiency-or-invariance claim bolted to it as a falsifiable proposition about where the load-bearing signal lives. The sharper the claim ('the trigger must be in the input', 'scale dilutes contamination', 'the scalar score is the signal'), the more leverage the pivot has.
  \item Identify an alternative locus --- supervision records, intermediate-activation geometry, score-distribution shape, the full stochastic output distribution, behavioral input-output traces, internal-state trajectories --- and argue structurally why the property of interest must manifest there too, ideally MORE robustly: invariant to the very transformations or access restrictions that defeat the assumed locus. If the alternative is only a cosmetic variant of the assumed locus, reviewers fold the contribution back into prior work, so the distinction must be load-bearing and explicit.
  \item Build the mechanism at the new locus: for an attack, the lowest-access, lowest-budget operation that exploits it; for a defense, a signal demonstrably agnostic to the attacker's configuration or insertion point. Where possible, couple detection and action through the same locus signal so the move is one unified primitive rather than two bolted-on stages.
  \item Validate against the strongest adversary that actually knows the new locus --- an adaptive attacker for a defense, or the realistic access/budget constraints for an attack --- not merely a fixed, non-adaptive baseline. Quietly demonstrating the move only against the bypassed locus, or against outdated baselines, is the single most common reason this tactic fails to convince.
  \item Supply a mechanistic account of WHY the new locus carries the signal --- an invariance proof, an impossibility-style bound, or a root-cause decomposition of the pipeline --- not merely evidence THAT it does. A relocation shown to work but never explained reads as a stacked combination of existing components or an anecdote, and forfeits the conceptual-lens credit that elevates these papers.
\end{enumerate}
\textbf{Differentiation within parent.} \textit{All siblings share the parent's audit-and-pivot signature, but they pivot mathematical objects inside a fixed locus, whereas this cluster pivots the locus itself in adversarial/empirical security and safety settings and validates with attack-success rates and detection performance rather than tighter analysis. Cluster 11 re-derives privacy-accounting bounds by reframing repeated selection as a resource-bounded mechanism --- the assumption is relaxed but the quantity (a cumulative bound) and its locus (the accounting calculus) are unchanged; this cluster instead relocates the enforcement point entirely (e.g., from parameter-divergence to score-distribution geometry) and shows the old locus was the wrong place to look. Likewise cluster 29 proves a new property of a projection operator on a constraint set and cluster 5 derives an exact algebraic equivalence between model families --- both are theorem-driven reformulations of analytic structure, while this cluster's deliverable is a demonstrated attack surface or a transformation-invariant detection signal. The contrast is empirical-locus-shift-under-adversary, not closed-form quantity tightening.}

\textbf{When to pick this one.} Pick this instantiation when the gap sits in a security, safety, detection, or provenance setting and you can observe that every existing attack or defense is anchored at one representational locus (raw input, parameters/gradients, a scalar score, a deterministic output, or an injected artifact). The trigger to instantiate here is a suspicion that the real load-bearing signal lives at an overlooked locus --- labels, activations, distribution shape, or behavior --- that the field has tacitly assumed irrelevant or invariant.

\textbf{Tactical failure mode.} The recurring failure is a successful relocation that is demonstrated but not adversarially closed: the paper shows its new-locus attack or defense beats a fixed, non-adaptive baseline, yet never stress-tests against an informed adversary who also operates at the new locus, so stealthiness, evasion, or robustness claims are asserted rather than proven. A second shape is a novelty-boundary collapse --- the "new" locus turns out to be one a prior paper already used (normalization statistics, activation-similarity alignment, pre-encoded semantic concepts), and the work fails to quantify what is genuinely different. A third is reintroducing the very assumption the relocation was meant to dissolve, by depending on a clean reference model, a known anomaly distribution, or other access the realistic threat model denies. A fourth is offering only a demonstration with no mechanistic account of why the locus carries the signal, which reads as a stacked pipeline of off-the-shelf parts. Self-check: can an adversary who knows your new locus defeat it as cheaply as the old one, is your locus provably distinct from published loci, and can you state structurally WHY the signal lives there?

\textbf{Examples} (paper-agnostic lessons):
\begin{itemize}[leftmargin=12pt,topsep=2pt,itemsep=1pt]
  \item \textbf{[Oral lesson]} Moving a trigger or detection signal from the input space to the supervision/label space opens an attack surface the entire defense literature was built around the wrong locus to catch.
  \item \textbf{[Oral lesson]} A detection or provenance signal anchored in intermediate-representation geometry survives exactly the transformations --- fine-tuning, pruning, unit reordering, rescaling --- that destroy parameter- and output-level signals.
  \item \textbf{[Oral lesson]} Treating the shape of a learned distribution (its local curvature, positional-depth profile, or full sampled spread) as the signal, rather than a scalar summary, surfaces structure that point estimates provably discard.
  \item \textbf{[Oral lesson]} Replacing a randomized, probabilistic guarantee with a deterministic combinatorial construction at the same structural locus can upgrade a soft bound into an exact certificate.
  \item \textbf{[Oral lesson]} Showing that a stated security property itself --- an imperceptibility norm bound, scale-induced dilution, or deterministic evaluation --- is the exploitable handle converts a design parameter into a provable vulnerability.
  \item \textbf{[Oral lesson]} Decomposing an empirically observed failure into structural root causes of the training pipeline turns ad hoc attack discovery into a systematic, predictive design process.
  \item \textbf{[Reject lesson]} Claiming a relocated signal is 'stealthier' or 'harder to remove' without defeating an adaptive defender who searches the same new locus leaves the central claim unverified.
  \item \textbf{[Reject lesson]} Relocating to a locus already exploited by published work --- normalization statistics, activation-similarity losses, semantic concept triggers --- collapses the novelty boundary unless the difference is made explicit and measured.
  \item \textbf{[Reject lesson]} A defense whose relocation silently depends on a clean reference model or a known anomaly distribution reintroduces precisely the strong assumption the move was supposed to eliminate.
  \item \textbf{[Reject lesson]} Demonstrating the new locus only on fixed, non-adaptive or outdated baselines while omitting any account of why it works reads as a stacked pipeline of existing components rather than a principled shift.
  \item \textbf{[Reject lesson]} Wiring off-the-shelf generators or tools into a new attack setting without a structural reason for the locus change gets judged as engineering integration, not insight.
\end{itemize}
\end{tcolorbox}

\bibliographystyle{plain}
\clearpage
\bibliography{references}

\end{document}